\documentclass[10pt,twocolumn]{article}

\usepackage{times}
\usepackage{epsfig}
\usepackage{amssymb}
\usepackage{amsmath}
\usepackage{lineno}
\usepackage{authblk}
\usepackage{hyperref}
\usepackage{multirow}
\usepackage{booktabs}
\usepackage{xcolor}
\usepackage{colortbl}
\usepackage{adjustbox}
\usepackage{graphicx}
\usepackage{subcaption}
\usepackage{pifont}
\usepackage[T1]{fontenc}
\usepackage{tabularx}

\hypersetup{
  colorlinks=true,
  linkcolor=blue,
  citecolor=blue,
  urlcolor=blue,
  pdfborder={0 0 0},
}

\hypersetup{
    pdftitle={},
    pdfauthor={}
}

\title{Inference Dynamics: Image Classifiers Keep Using Visual Cues After They Are Removed}

\author[1]{Ruiqi Wang}
\author[2]{Mingming Lu}

\affil[1]{Central South University, Changsha, Hunan, China}

\date{}

\begin{document}
\maketitle

\begin{abstract}
Image classifiers are typically evaluated by their final predictions,
treating inference as a static image-to-label mapping. Reliability,
however, also depends on whether a model can relinquish evidence after
that evidence disappears. We reveal a counterintuitive phenomenon:
removing a local visual cue from its original location may not remove
its influence on the prediction. Once captured by early layers, the cue
can become distributed across later representations, making ``no longer
visible'' different from ``no longer relied upon.''
We introduce \textbf{Inference Dynamics}, a process-level framework for
tracing how cue influence evolves and remains withdrawable across network
depth. Given an image and a matched cue-removed reference, we replace
cue-aligned states in the original branch with their reference
counterparts at successive layers and measure recovery toward the
cue-free prediction. The resulting trajectory separates evidence
locality from decision-level withdrawability.
Across datasets and architectures, we find locality declines before
withdrawability collapses: cue-induced information becomes less
concentrated at its original support while its effect remains active,
before entering an \textbf{evidence lock-in} stage. Consequently, fixed
late-stage correction may alter the final readout without removing
accumulated cue dependence. We therefore propose \textbf{Evidence Reset},
which selects an input-dependent layer before lock-in and reconstructs
cue-aligned states toward their cue-free counterparts. Our results
establish evidence withdrawability as a process-level dimension of
classifier reliability and suggest a new intervention principle:
correction depth should follow the dynamics of evidence rather than be
fixed in advance.
\end{abstract}

\section{Introduction}
\begin{figure}[t]
    \centering
    \includegraphics[width=\columnwidth]{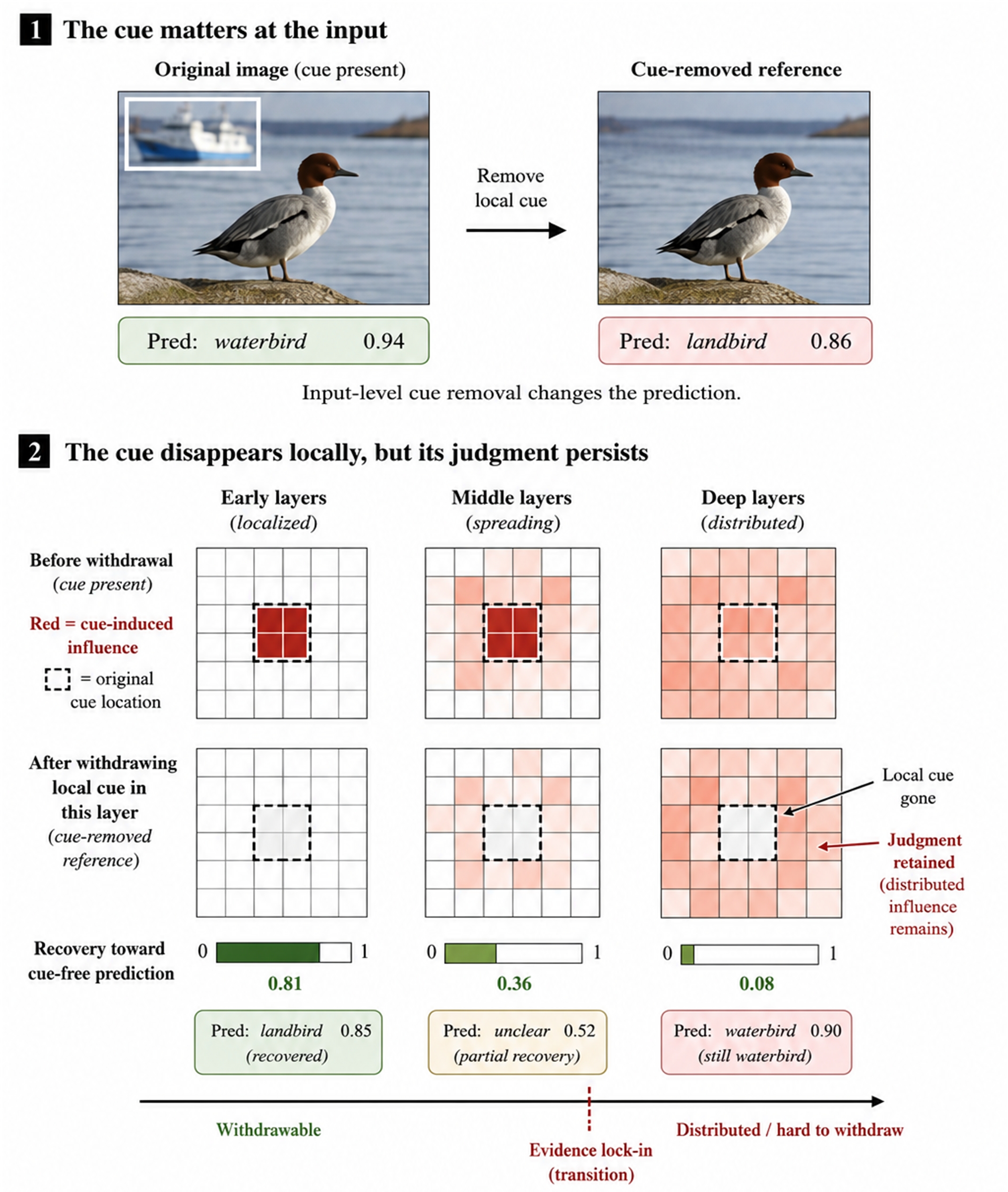}
    \caption{
    The cue disappears locally, its judgment persists.
    }
    \label{fig:teaser}
\end{figure}

Modern image classifiers can achieve high benchmark accuracy while
relying on spurious texture or background cues that fail under
distribution shift~\cite{c:1,r:1,c:2}. This has motivated a shift from
asking only whether a prediction is correct to asking what evidence
supports it~\cite{c:3}. We study a complementary question: once
supporting evidence disappears, can the model stop relying on it?
As illustrated in Fig.~\ref{fig:teaser}, removing a local cue from its
original location need not eliminate its predictive influence. The cue
may vanish locally while its induced judgment persists, making
``no longer visible'' different from ``no longer relied upon.''
Reliable inference should therefore require not only using valid
evidence, but also relinquishing the decisions it supports when that
evidence is removed.

Existing evidence-analysis methods either perturb image regions and
measure output changes~\cite{c:4,sobieski2025rethinking}, or localize
influential pixels, regions, neurons, and concepts through attribution
and representation analysis
\cite{c:5,c:6,pmlr-v80-kim18d,cai2025gefa}. Recent shortcut studies
further identify interpretable shortcut concepts and characterize their
layer-wise effects
\cite{kuhn2025efficientunsupervisedshortcutlearning,tsoy2025measuring},
while causal diagnostics distinguish representing a shortcut from
relying on it~\cite{grobner2026representation}. These methods reveal
what a classifier encodes, where it appears, and whether it affects the
output, but not whether a cue remains withdrawable after entering the
computation. In particular, they do not test at which layers replacing
the cue's local state can still recover the cue-free decision.

Shortcut mitigation methods use reweighting, robust training,
classifier retraining, or intermediate-representation interventions
\cite{sagawa2019distributionally,pmlr-v139-liu21f,kirichenko2022last,
pmlr-v202-tiwari23a,NEURIPS2024_dc4db2ff}; recent approaches also
suppress spurious internal activations or concept regions
\cite{NEURIPS2024_95c6ae3f,Kuhn_2025_ICCV,Elisha_2026_CVPR}.
However, intervention depth is typically fixed, and success is judged
mainly by final performance, leaving unclear whether accumulated cue
influence is removed or only the final readout is altered.

We introduce \textbf{Inference Dynamics}, which pairs an image with a
cue-removed reference, replaces cue-aligned states with their reference
counterparts at successive layers, and measures recovery toward the
cue-free prediction. The resulting trajectory reveals that evidence
locality can decline before decision-level withdrawability collapses,
exposing an \textbf{evidence lock-in} transition. We then propose
\textbf{Evidence Reset}, which uses each cue's dynamics to select an
intervention depth before lock-in and reconstruct cue-aligned states
from still-valid visual information.
Our contributions are:
\begin{itemize}
    \item We introduce \textbf{evidence withdrawability} as a
    process-level dimension of classifier reliability and develop
    \textbf{Inference Dynamics} to trace it across layers.

    \item We uncover a mismatch between \textbf{evidence locality} and
    withdrawability, revealing an \textbf{evidence lock-in} stage in
    which cue influence can no longer be withdrawn from its original
    support.

    \item We propose \textbf{Evidence Reset}, which selects an
    input-dependent correction depth and acts before lock-in rather
    than at a fixed layer.
\end{itemize}

\section{Inference Dynamics}
\begin{figure*}[t]
    \centering
    \includegraphics[width=\textwidth]{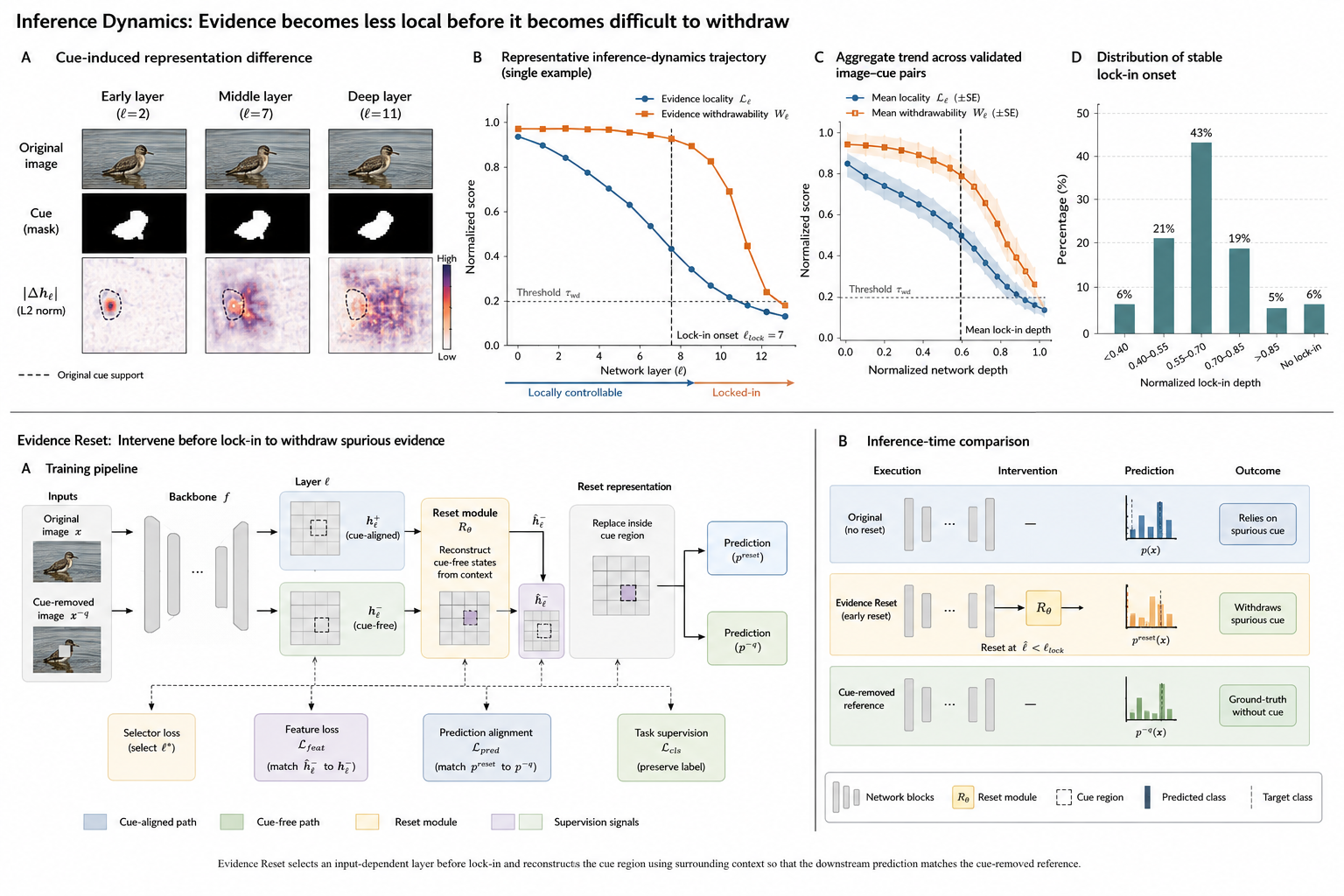}
    \caption{
    Overview of our framework.
    }
    \label{fig:framework_overview}
\end{figure*}
Figure~\ref{fig:framework_overview} provides an overview of our framework.
The upper part summarizes Inference Dynamics, which traces how a validated cue becomes less local and eventually difficult to withdraw.
The lower part summarizes Evidence Reset, which uses this trajectory to select an intervention layer before evidence lock-in and reconstruct cue-free states.
\subsection{Problem Formulation and Cue Validation}
\label{sec:cue_validation}
Our goal is to trace how the influence of a visual cue evolves as an image passes through a classifier. To study this process, we compare two matched executions of the same model: one on the original image and one on a counterpart in which the selected cue has been removed.
Let $f$ be an image classifier that maps an input image $x$ to class logits:
\begin{equation}
z(x) = f(x) \in \mathbb{R}^{C}.
\label{eq:classifier_output}
\end{equation}
Let $c = \operatorname*{arg,max}_{k} z_k(x)$ denote the class predicted from the original image. We consider a visual cue $q$, whose input-space support is specified by a mask $m_q$, and denote by $x^{\setminus q}$ a matched image in which $q$ is removed while the remaining scene content is preserved as much as possible. The cue may correspond to an object part, a background region, a co-occurring object, or another visual pattern used by the classifier.
We use the term \emph{evidence} in a model-relative sense: a cue is evidence for class $c$ if its removal reduces the model's support for that prediction. We measure this effect by
\begin{equation}
\Delta_q(x)
=
z_c(x) - z_c\left(x^{\setminus q}\right).
\label{eq:cue_effect}
\end{equation}
A positive $\Delta_q(x)$ indicates that the classifier relies on $q$ when predicting class $c$. The predicted label need not change, since a cue may substantially influence the decision without being sufficient to alter the final argmax. We retain only image--cue pairs for which the effect exceeds a relevance threshold:
\begin{equation}
\Delta_q(x) \geq \tau_{\mathrm{rel}}.
\label{eq:cue_relevance}
\end{equation}
We additionally exclude pairs in which cue removal fails, substantially changes non-target content, or introduces conspicuous editing artifacts. The cue-removal procedure and corresponding quality checks are described in the experimental setup.
Each retained pair $(x, x^{\setminus q})$ therefore provides two matched forward passes: one in which the selected evidence is present and one in which it is absent. In the next section, we use the internal states from these two executions to withdraw the cue-aligned representation at different network depths and determine how long its effect remains locally removable.
\subsection{Layer-wise Evidence Withdrawal}
\label{sec:evidence_withdrawal}
The previous section establishes that the selected cue $q$ affects the model's prediction. We now ask a different question: \textbf{at what depth can this effect still be removed from the location where the cue originally enters the network?}
Our intervention is deliberately local. It does not attempt to erase every downstream trace of $q$. Instead, at each layer $l$, we test whether replacing the representation aligned with the cue's original support is still sufficient to undo its effect on the final decision.
For each validated pair $(x, x^{\setminus q})$, we perform two forward passes through the same classifier. Let $f_{\leq l}$ denote the network up to layer $l$, and let $f_{>l}$ denote the remaining network. The paired representations at layer $l$ are
\begin{equation}
\begin{aligned}
h_l^{+} &= f_{\leq l}(x), \
h_l^{-} &= f_{\leq l}\left(x^{\setminus q}\right),
\end{aligned}
\label{eq:paired_representations}
\end{equation}
where $h_l^{+}$ is obtained from the original image and $h_l^{-}$ from its cue-removed counterpart.
Let $M_l$ denote the input cue mask $m_q$ mapped to the spatial units of layer $l$. For convolutional networks, $M_l$ selects the corresponding feature-map locations; for vision transformers, it selects the associated image tokens. Architecture-specific mapping details are provided in the experimental setup.
We withdraw the cue-aligned representation at layer $l$ by replacing the selected states in the original branch with their counterparts from the cue-removed branch:
\begin{equation}
\widetilde{h}_l^{,q}
=
\left(1-M_l\right)\odot h_l^{+}
+
M_l\odot h_l^{-}.
\label{eq:layerwise_withdrawal}
\end{equation}
The modified representation is then propagated through the remaining layers:
\begin{equation}
z_l^{\mathrm{wd}}
=
f_{>l}\left(\widetilde{h}_l^{,q}\right).
\label{eq:withdrawal_output}
\end{equation}
We refer to this intervention as \emph{evidence withdrawal at layer $l$}. Unlike zero ablation or noise injection, the replacement is taken from the same model processing a matched image without $q$. It therefore provides a context-matched representation of the selected locations in the cue's absence.
Each layer is intervened on independently, using the intact paired forward passes rather than the output of an earlier intervention. If replacing the cue-aligned states substantially reverses the cue-induced prediction change, then the cue's effect remains removable from its original support at that depth. If the replacement has little effect, we make the narrower conclusion that the original support is no longer sufficient to withdraw the cue's decision influence. Whether this loss of withdrawability coincides with a spatial redistribution of cue-induced information is examined separately in the next section.
Repeating the intervention across all layers yields the withdrawal outcomes ${z_l^{\mathrm{wd}}}_{l=0}^{L}$. The next section uses these outcomes to quantify how the spatial concentration and decision-level withdrawability of visual evidence evolve with network depth.
\subsection{Quantifying Evidence Locality and Withdrawability}
Layer-wise withdrawal reveals how the prediction responds when the cue-aligned representation is replaced at each depth. This response alone, however, does not show how the representation induced by the cue is distributed inside the network. We therefore measure two complementary properties: \emph{evidence locality}, which describes where cue-induced representation differences are concentrated, and \emph{evidence withdrawability}, which describes whether removing the cue-aligned states is still sufficient to reverse their effect on the prediction.

\paragraph{Evidence locality.}
For each layer $l$, we first compute the difference between the representations produced by the original and cue-removed inputs:
\begin{equation}
\delta_l
=
h_l^{+} - h_l^{-}.
\end{equation}
Let $\mathcal{U}_l$ denote the set of representation units at layer $l$, and let $\delta_l(u)$ be the channel-wise difference associated with unit $u$. For convolutional networks, the units correspond to spatial feature-map locations. For vision transformers, they include image tokens and any global token. We summarize the magnitude of the cue-induced difference at each unit as
\begin{equation}
a_l(u)
=
\left\lVert \delta_l(u) \right\rVert_2.
\end{equation}
The layer-specific cue mask $M_l$ identifies the units aligned with the cue's original spatial support. For units without a direct spatial correspondence, such as a global token, we set $M_l(u)=0$. We define evidence locality as the fraction of the total cue-induced difference that remains within the original support:
\begin{equation}
L_l(q;x)
=
\frac{
\sum_{u \in \mathcal{U}*l}
M_l(u),a_l(u)
}{
\sum*{u \in \mathcal{U}_l}
a_l(u) + \epsilon
},
\end{equation}
where $\epsilon$ is a small constant for numerical stability. A high $L_l(q;x)$ indicates that the difference between the two forward passes remains concentrated near the cue's original location. A low value indicates that a larger fraction of the difference appears outside that support. Importantly, this score measures the spatial concentration of representation differences; it does not by itself determine whether those differences influence the final prediction.
\paragraph{Evidence withdrawability.}
We next measure how much of the cue's prediction effect can be removed by the intervention at layer $l$. Recall that the input-level effect of cue removal is
\begin{equation}
\Delta_q(x)
=
z_c(x) - z_c\left(x^{\setminus q}\right).
\end{equation}
Using the withdrawal output $z_l^{\mathrm{wd}}$, we define the normalized withdrawability score as
\begin{equation}
W_l(q;x)
=
\frac{
z_c(x) - z_{l,c}^{\mathrm{wd}}
}{
z_c(x) - z_c\left(x^{\setminus q}\right)
}.
\end{equation}
A value of $W_l(q;x)$ close to $1$ means that replacing the cue-aligned states at layer $l$ reproduces most of the prediction change caused by removing the cue at the input. A value close to $0$ means that the same local intervention leaves the original prediction largely unchanged. We retain values outside the interval $[0,1]$, since they reveal interventions that either amplify the original effect or move the prediction in the opposite direction.
The two measurements answer different questions. Evidence locality asks whether cue-induced representation differences remain concentrated at the cue's original support, whereas evidence withdrawability asks whether modifying that support is still sufficient to change the final decision. Neither quantity can replace the other: a cue may remain detectable near its original location without that location being sufficient to control the decision, while representation differences outside the original support do not necessarily imply that the cue's decision effect has become irreversible.
We define the inference dynamics of cue $q$ for image $x$ as the joint layer-wise trajectory
\begin{equation}
\mathcal{D}*q(x)
=
\left\{
\left(
L_l(q;x),
W_l(q;x)
\right)
\right\}*{l=0}^{L}.
\end{equation}
This trajectory separates the spatial presence of cue-induced information from its ability to influence the prediction through the cue's original support. The next subsection examines how these two properties diverge with depth and uses their divergence to characterize the emergence of evidence lock-in.
\subsection{The Emergence of Evidence Lock-in}
The joint trajectory of locality and withdrawability reveals a two-stage transition. Evidence locality begins to decline first, indicating that the representation difference induced by cue $q$ is no longer concentrated within its original support. During this stage, however, withdrawability remains high: replacing the cue-aligned states can still reproduce much of the prediction change caused by removing the cue at the input. The cue is therefore becoming spatially distributed while its original support remains locally controllable.
\begin{figure*}[t]
    \centering
    \includegraphics[
        width=0.88\textwidth
    ]{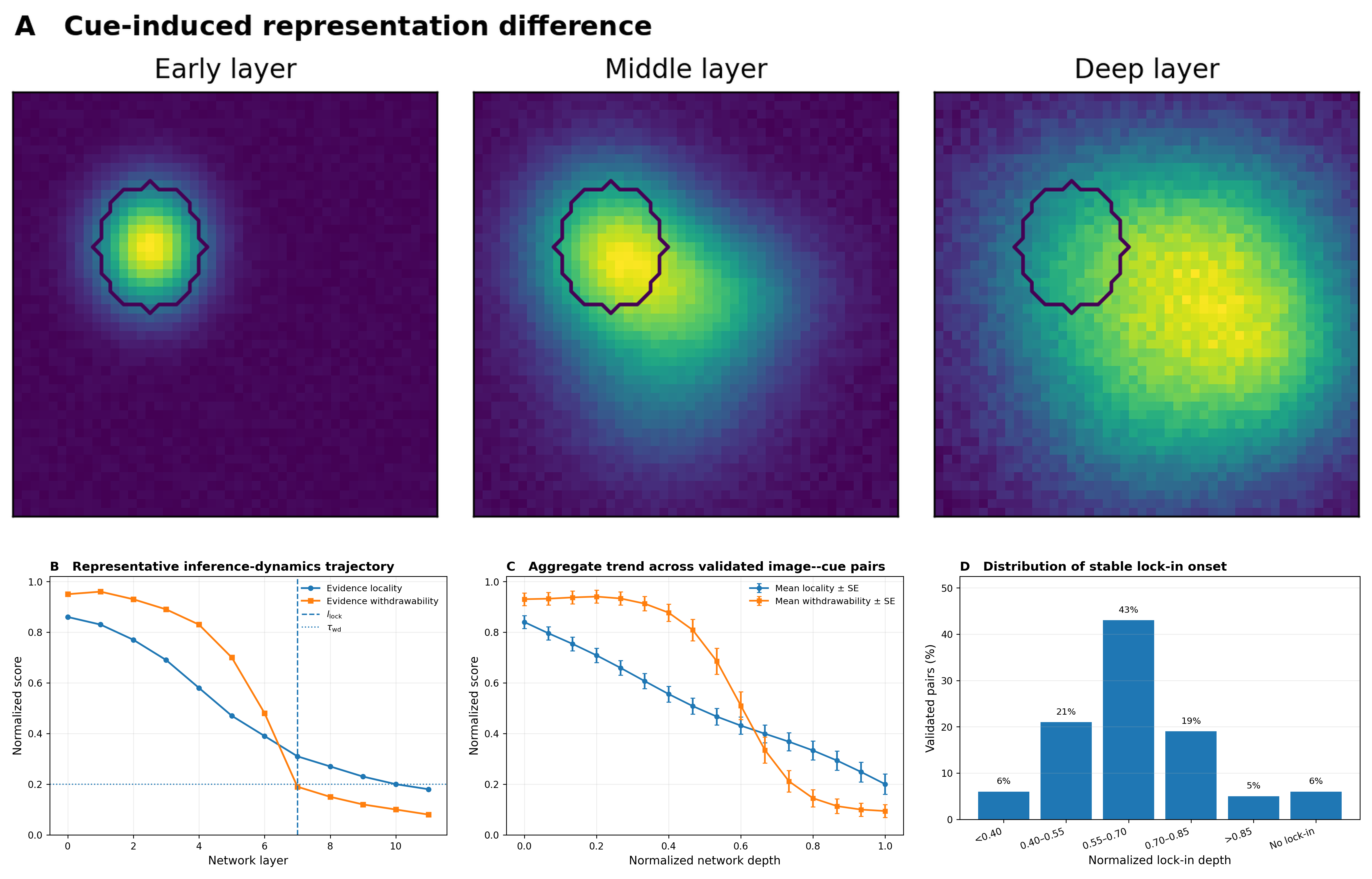}
    \caption{
        Evidence becomes less local before it becomes difficult to withdraw.
    }
    \label{fig:evidence}
\end{figure*}
As inference proceeds, withdrawability also drops and remains low. At this point, replacing the cue-aligned states no longer substantially reverses the cue-induced prediction effect. We call this persistent loss of local control \emph{evidence lock-in}. The term does not imply that all information about $q$ is irreversible. It states the narrower, directly testable conclusion that the cue's influence can no longer be withdrawn from the location where it originally entered the representation.
To distinguish a stable transition from an isolated layer-wise fluctuation, we define the lock-in onset as
\begin{equation}
    l_{\mathrm{lock}}
    =
    \min
    \left\{
        l
        \;\middle|\;
        W_j(q;x) \leq \tau_{\mathrm{wd}},
        \;
        \forall j \in
        \left\{
            l,\ldots,l+r-1
        \right\}
    \right\},
\end{equation}
where $\tau_{\mathrm{wd}}$ is the withdrawability threshold and $r$ is the required number of consecutive layers. If no such interval exists, the example is treated as not exhibiting stable lock-in.
Figure~\ref{fig:evidence} summarizes this transition. Panel A visualizes the cue-induced representation difference at increasing depths. The difference is initially concentrated near the original cue support, then gradually extends beyond it. Panel B shows the same process for a representative example: locality decreases before withdrawability, leaving an intermediate interval in which the cue has become less local but remains removable. Panel C confirms this ordering across validated image--cue pairs, while Panel D reports the distribution of stable lock-in depths.
This temporal separation is important for intervention. Acting only after withdrawability has collapsed is too late for a local correction, whereas intervening indiscriminately at the earliest layers may disrupt useful visual processing. 

\section{Evidence Reset}
The inference dynamics above reveal an actionable interval: evidence locality decreases before withdrawability collapses. During this interval, the cue has begun to spread beyond its original support, yet modifying that support can still alter the final decision. This suggests that intervention should be neither uniformly early nor postponed until deep layers. We therefore introduce \emph{Evidence Reset}
.As summarized in the lower part of Figure~\ref{fig:framework_overview}, Evidence Reset uses the dynamics identified in the previous section to choose an input-dependent layer before lock-in and reset the cue-aligned representation toward its cue-free counterpart.
\subsection{Lock-in-aware Evidence Reset}
A useful reset layer should satisfy two conditions. The cue-induced representation should have begun to leave its original support, making intervention necessary, while its prediction effect should remain locally withdrawable, making intervention effective. We capture this trade-off with the resetability score
\begin{equation}
    \rho_l(q;x)
    =
    \left(1-\overline{L}_l(q;x)\right)
    \overline{W}_l(q;x),
\end{equation}
where $\overline{L}_l$ and $\overline{W}_l$ denote the locality and withdrawability scores clipped to $[0,1]$ for layer selection. A high $\rho_l$ indicates that the cue is becoming spatially distributed but remains controllable from its original support.
For a training pair that exhibits stable lock-in, the target reset layer is selected as
\begin{equation}
    l^{\star}
    =
    \operatorname*{arg\,max}_{l < l_{\mathrm{lock}}}
    \rho_l(q;x).
\end{equation}
Restricting the search to layers before $l_{\mathrm{lock}}$ prevents the method from selecting a stage at which local intervention has already become ineffective. Unlike choosing a fixed shallow or deep layer, $l^{\star}$ adapts to the dynamics of each image--cue pair.
At inference, the cue-removed branch is unavailable. We therefore train a lightweight selector $\pi_{\phi}(l \mid x,m_q)$ to predict the reset layer from the original image and cue mask. Its output space contains the candidate layers together with a no-reset option. The latter allows the model to preserve the original forward pass when no reliable intervention point is identified.
Let $\widehat{l}$ be the selected layer, and let $h_{\widehat{l}}^{+}$ be the corresponding representation of the original image. A layer-conditioned reset module $R_{\theta}$ reconstructs the cue-free states from the surrounding representation:
\begin{equation}
    \widehat{h}_{\widehat{l}}^{-}
    =
    R_{\theta}
    \left(
        \left(1-M_{\widehat{l}}\right)
        \odot h_{\widehat{l}}^{+},
        M_{\widehat{l}},
        \widehat{l}
    \right).
\end{equation}
The cue-aligned states are masked before reconstruction so that the reset module cannot simply copy the representation to be removed. We then preserve all states outside the cue support and replace only the selected region:
\begin{equation}
    h_{\widehat{l}}^{\mathrm{reset}}
    =
    \left(1-M_{\widehat{l}}\right)
    \odot h_{\widehat{l}}^{+}
    +
    M_{\widehat{l}}
    \odot \widehat{h}_{\widehat{l}}^{-}.
\end{equation}
The reset representation is propagated through the remaining network:
\begin{equation}
    z^{\mathrm{reset}}
    =
    f_{>\widehat{l}}
    \left(
        h_{\widehat{l}}^{\mathrm{reset}}
    \right).
\end{equation}
Evidence Reset does not attempt to reconstruct an entirely cue-free network state after the cue has become locked in. Instead, it acts before that transition, when modifying the original support can still redirect the downstream computation.
\begin{table*}[t]
    \centering
    \small
    \setlength{\tabcolsep}{5.0pt}
    \caption{
    Generality and predictive validity of evidence lock-in.
    Rates are mean $\pm$ standard deviation over three runs, and
    normalized lock-in depth is median $[\mathrm{IQR}]$ across
    validated image--cue pairs. The last two columns report
    $\mathrm{WG}_{\mathrm{pre}}-\mathrm{WG}_{\mathrm{post}}$ and
    $\mathrm{NCR}_{\mathrm{post}}-\mathrm{NCR}_{\mathrm{pre}}$;
    larger values favor intervention before lock-in.
}
    \resizebox{\textwidth}{!}{
    \begin{tabular}{llcccccc}
        \toprule
        \small
        & &
        \multicolumn{3}{c}{Generality} &
        \multicolumn{3}{c}{Predictive validity} \\
        \cmidrule(lr){3-5}
        \cmidrule(lr){6-8}
        Dataset & Backbone
        & Locality-first (\%)
        & Stable lock-in (\%)
        & Median $l_{\mathrm{lock}}/L$
        & Best layer pre-lock (\%)
        & Pre--post WG gap
        & NCR reduction \\
        \midrule
        Waterbirds
        & ResNet-50
        & $84.7 \pm 2.1$
        & $90.8 \pm 1.7$
        & $0.61\,[0.54,0.69]$
        & $88.2 \pm 1.9$
        & $+7.4$
        & $0.22$ \\

        Waterbirds
        & ViT-B/16
        & $88.5 \pm 1.8$
        & $93.1 \pm 1.4$
        & $0.57\,[0.50,0.65]$
        & $91.0 \pm 1.6$
        & $+8.6$
        & $0.25$ \\

        \midrule

        UrbanCars
        & ResNet-50
        & $79.6 \pm 2.7$
        & $86.2 \pm 2.2$
        & $0.66\,[0.57,0.75]$
        & $83.4 \pm 2.4$
        & $+6.1$
        & $0.18$ \\

        UrbanCars
        & ViT-B/16
        & $82.9 \pm 2.3$
        & $89.4 \pm 1.9$
        & $0.62\,[0.54,0.71]$
        & $86.7 \pm 2.1$
        & $+7.0$
        & $0.21$ \\
        \bottomrule
    \end{tabular}
    }
    \label{tab:lockin_validity}
\end{table*}
\subsection{Training Objective and Inference}
We keep the pretrained classifier $f$ fixed and train only the layer selector and reset module. Training uses class-preserving pairs $(x,x^{\setminus q})$, for which removing $q$ does not change the semantic label.
The selector is supervised by the dynamics-derived target layer $l^{\star}$:
\begin{equation}
    \mathcal{L}_{\mathrm{sel}}
    =
    -\log
    \pi_{\phi}
    \left(
        l^{\star}
        \mid x,m_q
    \right).
\end{equation}
Pairs without a stable, actionable reset interval are assigned to the no-reset option rather than being forced to undergo an intervention.
At the selected layer, the cue-removed branch provides a direct target for representation reconstruction. We define
\begin{equation}
    \mathcal{L}_{\mathrm{feat}}
    =
    \frac{
        \left\|
            M_{l^{\star}}
            \odot
            \left(
                \widehat{h}_{l^{\star}}^{-}
                -
                \operatorname{sg}
                \left(
                    h_{l^{\star}}^{-}
                \right)
            \right)
        \right\|_{1}
    }{
        \left\|M_{l^{\star}}\right\|_{1}
    },
\end{equation}
where $\operatorname{sg}(\cdot)$ denotes stop-gradient. This loss teaches the reset module to approximate the representation produced at the same locations when cue $q$ is absent.
We further align the downstream prediction with the cue-free branch:
\begin{equation}
    \mathcal{L}_{\mathrm{pred}}
    =
    D_{\mathrm{KL}}
    \left(
        \operatorname{sg}
        \left(
            p\left(x^{\setminus q}\right)
        \right)
        \,\middle\|\,
        p^{\mathrm{reset}}(x)
    \right),
\end{equation}
where $p(x^{\setminus q})$ and $p^{\mathrm{reset}}(x)$ are the softmax distributions produced by the cue-removed and reset executions, respectively. A standard classification loss preserves the task label:
\begin{equation}
    \mathcal{L}_{\mathrm{cls}}
    =
    \operatorname{CE}
    \left(
        z^{\mathrm{reset}},
        y
    \right).
\end{equation}
The complete training objective is
\begin{equation}
    \mathcal{L}
    =
    \mathcal{L}_{\mathrm{cls}}
    +
    \lambda_{\mathrm{feat}}
    \mathcal{L}_{\mathrm{feat}}
    +
    \lambda_{\mathrm{pred}}
    \mathcal{L}_{\mathrm{pred}}
    +
    \lambda_{\mathrm{sel}}
    \mathcal{L}_{\mathrm{sel}}.
\end{equation}
At inference, Evidence Reset requires only the original image $x$ and the mask $m_q$ of the selected cue. The selector first predicts $\widehat{l}$. If the no-reset option is selected, the classifier follows its original forward pass. Otherwise, the representation is reset once at layer $\widehat{l}$ and inference continues through the remaining network. Cue-removed images, layer-wise withdrawal scans, and ground-truth lock-in depths are used only for training supervision and are not required at test time.

\section{Experiments}
\subsection{Experimental Setup}
\paragraph{Datasets and cue construction.}
We use Waterbirds~\cite{sagawa2020distributionallyrobustneuralnetworks} and
UrbanCars~\cite{Li_2023_CVPR} as our primary benchmarks because they
provide spatially identifiable and controllable visual shortcuts.
Waterbirds correlates bird categories with land or water backgrounds;
we treat the background as the target cue and obtain its support from
the provided bird segmentation mask. UrbanCars contains two independent
shortcuts, the scene background and a co-occurring object, enabling both
single-cue and multi-cue evaluation. For these two benchmarks, matched
cue-removed references are constructed using their native compositing
pipelines: the task-relevant foreground is kept fixed while only the
selected shortcut component is removed or independently resampled. Additional results on Spawrious-M2M-Hard~\cite{lynch2023spawriousbenchmarkfinecontrol}
and the ImageNet-9 Background Challenge~\cite{xiao2020noisesignalroleimage}, together
with alternative cue-removal procedures, are reported in the
supplementary material.
\paragraph{Models and comparisons.}
We evaluate ImageNet-pretrained ResNet-50~\cite{he2016deep} and
ViT-B/16~\cite{dosovitskiy2021image}, covering convolutional and
transformer architectures. For Evidence Reset, we first fine-tune the
backbone using empirical risk minimization and then keep it fixed while
training the layer selector and reset module.
We compare with ERM, GroupDRO~\cite{sagawa2019distributionally},
JTT~\cite{pmlr-v139-liu21f}, DFR~\cite{kirichenko2022last}, and
Last Layer Ensemble (LLE)~\cite{Li_2023_CVPR}. To isolate the contribution
of lock-in-aware layer selection, we additionally compare with
input-level cue removal using the same editing operator, the best fixed
reset layer selected on the validation set, and an oracle reset using
the ground-truth dynamics-derived layer. All reset-based variants use
the same reset module, parameter budget, training data, and optimization
schedule.

\paragraph{Metrics and protocol.}
We report average accuracy, worst-group accuracy, cue-aligned accuracy,
and normalized cue residual (NCR). NCR is the Jensen--Shannon distance
between the reset and cue-free predictions, normalized by the
corresponding original--cue-free distance; lower values indicate more
complete cue withdrawal. All thresholds, fixed layers, and
hyperparameters are selected on the validation set. We report mean
$\pm$ standard deviation over three runs and 95\% bootstrap confidence
intervals for sample-level dynamics. Cue-removed references and
layer-wise scans are used only during training and analysis, not during
test-time inference. Further implementation details, seed-wise results,
sensitivity analyses, removal validation, and computational costs are
provided in the supplementary material.

\begin{table*}[t]
    \centering
    \scriptsize
    \setlength{\tabcolsep}{2.8pt}
    \renewcommand{\arraystretch}{0.82}
    \caption{
        Waterbirds and UrbanCars results averaged over ResNet-50 and
        ViT-B/16 (three-run mean $\pm$ std.).
        WG and Aligned denote worst-group and cue-aligned accuracy;
        lower NCR is better. Rows below the midrule are controlled
        reset baselines. Best Fixed uses one validation-selected layer;
        Oracle uses sample-specific target layers.
    }
    \vspace{-1mm}
    \resizebox{0.96\textwidth}{!}{%
    \begin{tabular}{@{}lcccccccc@{}}
        \toprule
        &
        \multicolumn{4}{c}{Waterbirds} &
        \multicolumn{4}{c}{UrbanCars} \\
        \cmidrule(lr){2-5}
        \cmidrule(lr){6-9}
        Method
        & Avg. $\uparrow$
        & WG $\uparrow$
        & NCR $\downarrow$
        & Aligned $\uparrow$
        & Avg. $\uparrow$
        & WG $\uparrow$
        & NCR $\downarrow$
        & Aligned $\uparrow$ \\
        \midrule

        ERM
        & $96.8 \pm 0.3$
        & $75.8 \pm 1.5$
        & $0.89 \pm 0.02$
        & $\mathbf{98.1 \pm 0.2}$
        & $94.7 \pm 0.4$
        & $47.7 \pm 1.8$
        & $0.92 \pm 0.02$
        & $\mathbf{98.3 \pm 0.2}$ \\

        GroupDRO
        & $95.0 \pm 0.4$
        & $91.0 \pm 0.8$
        & $0.64 \pm 0.03$
        & $95.8 \pm 0.4$
        & $92.8 \pm 0.5$
        & $68.9 \pm 1.2$
        & $0.70 \pm 0.03$
        & $94.9 \pm 0.5$ \\

        JTT
        & $95.5 \pm 0.3$
        & $88.2 \pm 1.0$
        & $0.68 \pm 0.03$
        & $96.4 \pm 0.3$
        & $93.5 \pm 0.4$
        & $65.0 \pm 1.3$
        & $0.74 \pm 0.03$
        & $95.6 \pm 0.4$ \\

        DFR
        & $96.1 \pm 0.2$
        & $89.8 \pm 0.7$
        & $0.61 \pm 0.02$
        & $97.0 \pm 0.2$
        & $94.1 \pm 0.3$
        & $66.2 \pm 1.0$
        & $0.68 \pm 0.02$
        & $96.4 \pm 0.3$ \\

        LLE
        & -- & -- & -- & --
        & $94.2 \pm 0.3$
        & $72.1 \pm 0.9$
        & $0.60 \pm 0.02$
        & $96.2 \pm 0.3$ \\

        \midrule

        Input-level Cue Removal
        & $94.1 \pm 0.5$
        & $85.7 \pm 1.1$
        & $0.36 \pm 0.03$
        & $93.0 \pm 0.6$
        & $91.9 \pm 0.6$
        & $63.8 \pm 1.4$
        & $0.41 \pm 0.04$
        & $90.7 \pm 0.7$ \\

        Best Fixed Reset
        & $96.0 \pm 0.3$
        & \underline{$91.4 \pm 0.7$}
        & \underline{$0.27 \pm 0.02$}
        & $96.9 \pm 0.3$
        & $94.3 \pm 0.3$
        & \underline{$73.9 \pm 0.9$}
        & \underline{$0.30 \pm 0.02$}
        & $96.1 \pm 0.3$ \\

        \textbf{Evidence Reset}
        & $\mathbf{96.3 \pm 0.2}$
        & $\mathbf{93.8 \pm 0.6}$
        & $\mathbf{0.14 \pm 0.01}$
        & $\mathbf{97.2 \pm 0.2}$
        & $\mathbf{94.6 \pm 0.3}$
        & $\mathbf{77.3 \pm 0.8}$
        & $\mathbf{0.17 \pm 0.02}$
        & $\mathbf{96.5 \pm 0.3}$ \\

        \textit{Oracle-layer Reset}
        & $96.5 \pm 0.2$
        & $94.8 \pm 0.5$
        & $0.09 \pm 0.01$
        & $97.3 \pm 0.2$
        & $94.8 \pm 0.2$
        & $79.1 \pm 0.7$
        & $0.12 \pm 0.01$
        & $96.6 \pm 0.2$ \\
        \bottomrule
    \end{tabular}%
    }
    \vspace{-1mm}
    \label{tab:reset_main}
\end{table*}
\begin{table}[t]
    \centering
    \scriptsize
    \setlength{\tabcolsep}{2.4pt}
    \renewcommand{\arraystretch}{0.92}
    \caption{
        Evidence Reset ablations averaged over datasets and backbones.
    }
    \resizebox{\columnwidth}{!}{
    \begin{tabular}{lcccc}
        \toprule
        \multicolumn{5}{l}{
            \textbf{(a) Layer-selection strategy}
        } \\
        \midrule
        Strategy
        & \shortstack{WG Acc.\\$\uparrow$}
        & \shortstack{NCR\\$\downarrow$}
        & \shortstack{Aligned\\$\uparrow$}
        & \shortstack{Layer Err.\\$\downarrow$} \\
        \midrule

        Random layer
        & $79.7$
        & $0.39$
        & $96.0$
        & $0.31$ \\

        Best fixed layer
        & $82.7$
        & $0.29$
        & $96.5$
        & -- \\

        Locality only: $1-\bar{L}_l$
        & $83.1$
        & $0.24$
        & $96.5$
        & $0.19$ \\

        Withdrawability only: $\bar{W}_l$
        & $83.8$
        & $0.22$
        & $96.6$
        & $0.16$ \\

        \textbf{Full: $(1-\bar{L}_l)\bar{W}_l$}
        & $\mathbf{85.6}$
        & $\mathbf{0.16}$
        & $\mathbf{96.9}$
        & $\mathbf{0.10}$ \\

        \textit{Oracle layer}
        & $87.0$
        & $0.11$
        & $97.0$
        & $0.00$ \\

        \midrule
        \multicolumn{5}{l}{
            \textbf{(b) Training objective and no-reset option}
        } \\
        \midrule
        Variant
        & \shortstack{WG Acc.\\$\uparrow$}
        & \shortstack{NCR\\$\downarrow$}
        & \shortstack{Aligned\\$\uparrow$}
        & \shortstack{No-reset Prec.\\$\uparrow$} \\
        \midrule

        w/o $\mathcal{L}_{\mathrm{feat}}$
        & $84.1$
        & $0.23$
        & $96.6$
        & $86.8$ \\

        w/o $\mathcal{L}_{\mathrm{pred}}$
        & $84.6$
        & $0.20$
        & $96.6$
        & $87.1$ \\

        w/o $\mathcal{L}_{\mathrm{cls}}$
        & $84.9$
        & $0.17$
        & $95.7$
        & $87.2$ \\

        w/o no-reset option
        & $85.1$
        & $0.16$
        & $95.3$
        & -- \\

        \textbf{Full objective}
        & $\mathbf{85.6}$
        & $\mathbf{0.16}$
        & $\mathbf{96.9}$
        & $\mathbf{88.4}$ \\

        \bottomrule
    \end{tabular}
    }
    \label{tab:reset_ablation}
\end{table}

\subsection{Evidence Lock-in: Generality and Predictive Validity}
Table~\ref{tab:lockin_validity} shows that evidence lock-in generalizes
across datasets and architecture families. The locality-first ordering
appears in $79.6$--$88.5\%$ of validated image--cue pairs, while
$86.2$--$93.1\%$ exhibit stable lock-in. Its median onset ranges from
$0.57L$ to $0.66L$, with broad interquartile ranges, indicating that
lock-in is common but does not occur at a fixed network depth. This
variation directly challenges the use of a single intervention layer
for all inputs.
We further test whether the dynamics-derived onset predicts actual
correction failure using metrics that are not involved in defining
lock-in. The best intervention lies before $l_{\mathrm{lock}}$ in
$83.4$--$91.0\%$ of examples. Compared with the best post-lock
intervention, the best pre-lock intervention improves worst-group
accuracy by $6.1$--$8.6$ points and reduces normalized cue residual by
$0.18$--$0.25$. Thus, lock-in is not merely a thresholded description
of withdrawability; it identifies when an otherwise identical local
correction becomes substantially less effective. Full layer sweeps,
per-cue results, confidence intervals, and no-lock-in cases are reported
in the supplementary material.

\subsection{Evidence Reset: Main Results and Ablations}
\paragraph{Main comparison.}
Table~\ref{tab:reset_main} shows that Evidence Reset consistently
improves worst-group accuracy while preserving average and cue-aligned
performance. Compared with the best fixed reset layer, it gains
$2.4$ points on Waterbirds and $3.4$ points on UrbanCars. More
importantly, NCR decreases from $0.27$ and $0.30$ to $0.14$ and
$0.17$, respectively, showing that the corrected prediction moves
toward the cue-free outcome rather than merely changing its final
argmax. Direct input-level cue removal reduces cue dependence but
substantially harms aligned-group accuracy, demonstrating that removing
a cue at the input is not equivalent to resetting its accumulated
influence at an appropriate internal stage.

\paragraph{Dynamic versus fixed-depth correction.}
All reset-based variants share the same backbone, reset module, training
data, and optimization budget; they differ only in how the intervention
layer is selected. Evidence Reset therefore isolates the benefit of
input-dependent correction and consistently outperforms the globally
best fixed layer selected on validation data. The remaining gap to
Oracle-layer Reset is modest but non-negligible, indicating that layer
prediction, rather than cue-free feature reconstruction alone, remains
the main source of error.

\paragraph{Ablations.}
Table~\ref{tab:reset_ablation} shows that jointly using locality and
withdrawability yields more accurate layer selection, higher
worst-group accuracy, and lower cue residual than either signal alone.
Removing feature or prediction alignment weakens cue withdrawal, while
removing classification supervision or the no-reset option harms
cue-aligned performance. Additional fixed-layer sweeps, per-backbone
results, sensitivity analyses, alternative removal operators,
efficiency, and failure cases are reported in the supplementary
material.
\section{Conclusion}
We introduced \textbf{Inference Dynamics} to trace whether visual
evidence remains locally removable throughout classifier inference.
Separating locality from withdrawability reveals an
\textbf{evidence lock-in} transition: cue-induced information leaves
its original support before local correction becomes ineffective.
\textbf{Evidence Reset} uses this transition to select an
input-dependent intervention depth, outperforming fixed-depth
correction while reducing residual cue influence. These results suggest
that correction depth should follow evidence dynamics rather than be
fixed in advance. Extending the framework beyond specified cue masks
and class-preserving training pairs remains future work.

\bibliographystyle{main}
\bibliography{main}

@inproceedings{c:1,
  title={ImageNet-trained CNNs are biased towards texture; increasing shape bias improves accuracy and robustness},
  author={Geirhos, Robert and Rubisch, Patricia and Michaelis, Claudio and Bethge, Matthias and Wichmann, Felix A and Brendel, Wieland},
  booktitle={International conference on learning representations},
  year={2018}
}

@article{r:1,
  title={Shortcut learning in deep neural networks},
  author={Geirhos, Robert and Jacobsen, J{\"o}rn-Henrik and Michaelis, Claudio and Zemel, Richard and Brendel, Wieland and Bethge, Matthias and Wichmann, Felix A},
  journal={Nature Machine Intelligence},
  volume={2},
  number={11},
  pages={665--673},
  year={2020},
  publisher={Nature Publishing Group UK London}
}

@inproceedings{c:2,
  title={Distributionally Robust Neural Networks},
  author={Sagawa, Shiori and Koh, Pang Wei and Hashimoto, Tatsunori B and Liang, Percy},
  booktitle={International Conference on Learning Representations},
  year={2020},
  url={https://openreview.net/forum?id=ryxGuJrFvS}
}

@inproceedings{c:3,
  author    = {Andrew Slavin Ross and Michael C. Hughes and Finale Doshi-Velez},
  title     = {Right for the Right Reasons: Training Differentiable Models by Constraining their Explanations},
  booktitle = {Proceedings of the Twenty-Sixth International Joint Conference on Artificial Intelligence, {IJCAI-17}},
  pages     = {2662--2670},
  year      = {2017},
  doi       = {10.24963/ijcai.2017/371},
  url       = {https://doi.org/10.24963/ijcai.2017/371},
}

@inproceedings{c:4,
  title={Interpretable explanations of black boxes by meaningful perturbation},
  author={Fong, Ruth C and Vedaldi, Andrea},
  booktitle={Proceedings of the IEEE international conference on computer vision},
  pages={3429--3437},
  year={2017}
}

@inproceedings{c:5,
  title={Grad-cam: Visual explanations from deep networks via gradient-based localization},
  author={Selvaraju, Ramprasaath R and Cogswell, Michael and Das, Abhishek and Vedantam, Ramakrishna and Parikh, Devi and Batra, Dhruv},
  booktitle={Proceedings of the IEEE international conference on computer vision},
  pages={618--626},
  year={2017}
}

@inproceedings{c:6,
  title={Network dissection: Quantifying interpretability of deep visual representations},
  author={Bau, David and Zhou, Bolei and Khosla, Aditya and Oliva, Aude and Torralba, Antonio},
  booktitle={Proceedings of the IEEE conference on computer vision and pattern recognition},
  pages={6541--6549},
  year={2017}
}

@inproceedings{sobieski2025rethinking,
  title={Rethinking visual counterfactual explanations through region constraint},
  author={Sobieski, Bartlomiej and Grzywaczewski, Jakub and Sadlej, Bart{\l}omiej and Tivnan, Matthew and Biecek, Przemyslaw},
  booktitle={International Conference on Learning Representations},
  volume={2025},
  pages={12967--13019},
  year={2025}
}

@InProceedings{pmlr-v80-kim18d,
  title = 	 {Interpretability Beyond Feature Attribution: Quantitative Testing with Concept Activation Vectors ({TCAV})},
  author =       {Kim, Been and Wattenberg, Martin and Gilmer, Justin and Cai, Carrie and Wexler, James and Viegas, Fernanda and sayres, Rory},
  booktitle = 	 {Proceedings of the 35th International Conference on Machine Learning},
  pages = 	 {2668--2677},
  year = 	 {2018},
  editor = 	 {Dy, Jennifer and Krause, Andreas},
  volume = 	 {80},
  series = 	 {Proceedings of Machine Learning Research},
  month = 	 {10--15 Jul},
  publisher =    {PMLR},
  url = 	 {https://proceedings.mlr.press/v80/kim18d.html}
}

@inproceedings{cai2025gefa,
  title={{GEFA}: A General Feature Attribution Framework Using Proxy Gradient Estimation},
  author={Yi Cai and Thibaud Ardoin and Gerhard Wunder},
  booktitle={Forty-second International Conference on Machine Learning},
  year={2025},
  url={https://openreview.net/forum?id=QyG0ilz5ju}
}

@misc{kuhn2025efficientunsupervisedshortcutlearning,
  title={Efficient Unsupervised Shortcut Learning Detection and Mitigation in Transformers}, 
  author={Lukas Kuhn and Sari Sadiya and Jorg Schlotterer and Florian Buettner and Christin Seifert and Gemma Roig},
  year={2025},
  eprint={2501.00942},
  archivePrefix={arXiv},
  primaryClass={cs.LG},
  url={https://arxiv.org/abs/2501.00942}, 
}

@inproceedings{tsoy2025measuring,
  title={Measuring the Layer-Wise Impact of Image Shortcuts on Deep Model Features},
  author={Nikita Tsoy and Nikola Konstantinov},
  booktitle={Workshop on Spurious Correlation and Shortcut Learning: Foundations and Solutions},
  year={2025},
  url={https://openreview.net/forum?id=5sigk5jNqt}
}

@inproceedings{grobner2026representation,
  title={Representation Is Not Reliance: Concept-Based Causal Diagnostics of Shortcut Learning in Vision},
  author={Gr{\"o}bner, Rene and Arnold, Stefan and Clement, Tobias},
  year = {2026},
  booktitle={Mechanistic Interpretability Workshop at ICML 2026}
}

@article{sagawa2019distributionally,
  title={Distributionally robust neural networks for group shifts: On the importance of regularization for worst-case generalization},
  author={Sagawa, Shiori and Koh, Pang Wei and Hashimoto, Tatsunori B and Liang, Percy},
  journal={arXiv preprint arXiv:1911.08731},
  year={2019}
}

@InProceedings{pmlr-v139-liu21f,
  title = 	 {Just Train Twice: Improving Group Robustness without Training Group Information},
  author =       {Liu, Evan Z and Haghgoo, Behzad and Chen, Annie S and Raghunathan, Aditi and Koh, Pang Wei and Sagawa, Shiori and Liang, Percy and Finn, Chelsea},
  booktitle = 	 {Proceedings of the 38th International Conference on Machine Learning},
  pages = 	 {6781--6792},
  year = 	 {2021},
  editor = 	 {Meila, Marina and Zhang, Tong},
  volume = 	 {139},
  series = 	 {Proceedings of Machine Learning Research},
  month = 	 {18--24 Jul},
  publisher =    {PMLR},
  url = 	 {https://proceedings.mlr.press/v139/liu21f.html}
}

@article{kirichenko2022last,
  title={Last layer re-training is sufficient for robustness to spurious correlations},
  author={Kirichenko, Polina and Izmailov, Pavel and Wilson, Andrew Gordon},
  journal={arXiv preprint arXiv:2204.02937},
  year={2022}
}

@InProceedings{pmlr-v202-tiwari23a,
  title = 	 {Overcoming Simplicity Bias in Deep Networks using a Feature Sieve},
  author =       {Tiwari, Rishabh and Shenoy, Pradeep},
  booktitle = 	 {Proceedings of the 40th International Conference on Machine Learning},
  pages = 	 {34330--34343},
  year = 	 {2023},
  editor = 	 {Krause, Andreas and Brunskill, Emma and Cho, Kyunghyun and Engelhardt, Barbara and Sabato, Sivan and Scarlett, Jonathan},
  volume = 	 {202},
  series = 	 {Proceedings of Machine Learning Research},
  month = 	 {23--29 Jul},
  publisher =    {PMLR},
  url = 	 {https://proceedings.mlr.press/v202/tiwari23a.html}
}

@inproceedings{NEURIPS2024_dc4db2ff,
 author = {LaBonte, Tyler and Hill, John C. and Zhang, Xinchen and Muthukumar, Vidya and Kumar, Abhishek},
 booktitle = {Advances in Neural Information Processing Systems},
 doi = {10.52202/079017-3865},
 editor = {A. Globerson and L. Mackey and D. Belgrave and A. Fan and U. Paquet and J. Tomczak and C. Zhang},
 pages = {121598--121629},
 publisher = {Curran Associates, Inc.},
 title = {The Group Robustness is in the Details: Revisiting Finetuning under Spurious Correlations},
 url = {https://proceedings.neurips.cc/paper_files/paper/2024/file/dc4db2ff2c1aefce3b594f821ea82fe6-Paper-Conference.pdf},
 volume = {37},
 year = {2024}
}

@inproceedings{NEURIPS2024_95c6ae3f,
 author = {Varma, Maya and Delbrouck, Jean-Benoit and Chen, Zhihong and Chaudhari, Akshay and Langlotz, Curtis},
 booktitle = {Advances in Neural Information Processing Systems},
 doi = {10.52202/079017-2614},
 editor = {A. Globerson and L. Mackey and D. Belgrave and A. Fan and U. Paquet and J. Tomczak and C. Zhang},
 pages = {82235--82264},
 publisher = {Curran Associates, Inc.},
 title = {RaVL: Discovering and Mitigating Spurious Correlations in Fine-Tuned Vision-Language Models},
 url = {https://proceedings.neurips.cc/paper_files/paper/2024/file/95c6ae3f3393786203a4b6dcb9df1036-Paper-Conference.pdf},
 volume = {37},
 year = {2024}
}

@InProceedings{Kuhn_2025_ICCV,
    author    = {Kuhn, Lukas and Sadiya, Sari and Schl\"otterer, J\"org and Buettner, Florian and Seifert, Christin and Roig, Gemma},
    title     = {Efficient Unsupervised Shortcut Learning Detection and Mitigation in Transformers},
    booktitle = {Proceedings of the IEEE/CVF International Conference on Computer Vision (ICCV)},
    month     = {October},
    year      = {2025},
    pages     = {2217-2226}
}

@InProceedings{Elisha_2026_CVPR,
    author    = {Elisha, Yehonatan and Barkan, Oren and Koenigstein, Noam},
    title     = {Concept-Guided Fine-Tuning: Steering ViTs away from Spurious Correlations to Improve Robustness},
    booktitle = {Proceedings of the IEEE/CVF Conference on Computer Vision and Pattern Recognition (CVPR)},
    month     = {June},
    year      = {2026},
    pages     = {17035-17045}
}

@misc{sagawa2020distributionallyrobustneuralnetworks,
  title={Distributionally Robust Neural Networks for Group Shifts: On the Importance of Regularization for Worst-Case Generalization}, 
  author={Shiori Sagawa and Pang Wei Koh and Tatsunori B. Hashimoto and Percy Liang},
  year={2020},
  eprint={1911.08731},
  archivePrefix={arXiv},
  primaryClass={cs.LG},
  url={https://arxiv.org/abs/1911.08731}, 
}

@InProceedings{Li_2023_CVPR,
    author    = {Li, Zhiheng and Evtimov, Ivan and Gordo, Albert and Hazirbas, Caner and Hassner, Tal and Ferrer, Cristian Canton and Xu, Chenliang and Ibrahim, Mark},
    title     = {A Whac-a-Mole Dilemma: Shortcuts Come in Multiples Where Mitigating One Amplifies Others},
    booktitle = {Proceedings of the IEEE/CVF Conference on Computer Vision and Pattern Recognition (CVPR)},
    month     = {June},
    year      = {2023},
    pages     = {20071-20082}
}

@misc{lynch2023spawriousbenchmarkfinecontrol,
  title={Spawrious: A Benchmark for Fine Control of Spurious Correlation Biases}, 
  author={Aengus Lynch and Gbètondji J-S Dovonon and Jean Kaddour and Ricardo Silva},
  year={2023},
  eprint={2303.05470},
  archivePrefix={arXiv},
  primaryClass={cs.CV},
  url={https://arxiv.org/abs/2303.05470}, 
}

@misc{xiao2020noisesignalroleimage,
  title={Noise or Signal: The Role of Image Backgrounds in Object Recognition}, 
  author={Kai Xiao and Logan Engstrom and Andrew Ilyas and Aleksander Madry},
  year={2020},
  eprint={2006.09994},
  archivePrefix={arXiv},
  primaryClass={cs.CV},
  url={https://arxiv.org/abs/2006.09994}, 
}

@inproceedings{he2016deep,
  title={Deep Residual Learning for Image Recognition},
  author={He, Kaiming and Zhang, Xiangyu and Ren, Shaoqing and Sun, Jian},
  booktitle={Proceedings of the IEEE Conference on Computer Vision and Pattern Recognition},
  pages={770--778},
  year={2016}
}

@inproceedings{dosovitskiy2021image,
  title={An Image is Worth 16x16 Words: Transformers for Image Recognition at Scale},
  author={Dosovitskiy, Alexey and others},
  booktitle={International Conference on Learning Representations},
  year={2021}
}

@article{ren2024dinox,
  title={DINO-X: A Unified Vision Model for Open-World Object Detection and Understanding},
  author={Ren, Tianhe and Chen, Yihao and Jiang, Qing and others},
  journal={arXiv preprint arXiv:2411.14347},
  year={2024}
}

@article{ravi2024sam2,
  title={SAM 2: Segment Anything in Images and Videos},
  author={Ravi, Nikhila and Gabeur, Valentin and Hu, Yuan-Ting and others},
  journal={arXiv preprint arXiv:2408.00714},
  year={2024}
}

@inproceedings{ju2024brushnet,
  title={BrushNet: A Plug-and-Play Image Inpainting Model with Decomposed Dual-Branch Diffusion},
  author={Ju, Xuan and Liu, Xian and Wang, Xintao and Bian, Yuxuan and Shan, Ying and Xu, Qiang},
  booktitle={European Conference on Computer Vision},
  year={2024}
}

@article{oquab2023dinov2,
  title={DINOv2: Learning Robust Visual Features without Supervision},
  author={Oquab, Maxime and Darcet, Timoth{\'e}e and Moutakanni, Th{\'e}o and others},
  journal={Transactions on Machine Learning Research},
  year={2024}
}

@inproceedings{geiger2021causal,
  title={Causal Abstractions of Neural Networks},
  author={Geiger, Atticus and Lu, Hanson and Icard, Thomas and Potts, Christopher},
  booktitle={Advances in Neural Information Processing Systems},
  volume={34},
  year={2021}
}

@inproceedings{geiger2022iit,
  title={Inducing Causal Structure for Interpretable Neural Networks},
  author={Geiger, Atticus and Wu, Zhengxuan and Lu, Hanson and Rozner, Josh and Kreiss, Elisa and Icard, Thomas and Goodman, Noah D. and Potts, Christopher},
  booktitle={Proceedings of the 39th International Conference on Machine Learning},
  series={Proceedings of Machine Learning Research},
  volume={162},
  pages={7324--7338},
  year={2022}
}

@inproceedings{ravfogel2020inlp,
  title={Null It Out: Guarding Protected Attributes by Iterative Nullspace Projection},
  author={Ravfogel, Shauli and Elazar, Yanai and Gonen, Hila and Twiton, Michael and Goldberg, Yoav},
  booktitle={Proceedings of the 58th Annual Meeting of the Association for Computational Linguistics},
  pages={7237--7256},
  year={2020}
}

@inproceedings{belrose2023leace,
  title={LEACE: Perfect Linear Concept Erasure in Closed Form},
  author={Belrose, Nora and Schneider-Joseph, David and Ravfogel, Shauli and Cotterell, Ryan and Raff, Edward and Biderman, Stella},
  booktitle={Advances in Neural Information Processing Systems},
  year={2023}
}

@article{izadi2026caap,
  title={Causal Attribution via Activation Patching},
  author={Izadi, Amirmohammad and Banayeeanzade, Mohammadali and Mirrokni, Alireza and Hasani, Hosein and Bagherian, Mobin and Mehri, Faridoun and Soleymani Baghshah, Mahdieh},
  journal={arXiv preprint arXiv:2603.13652},
  year={2026}
}

@inproceedings{wu2023discover,
  title={Discover and Cure: Concept-aware Mitigation of Spurious Correlation},
  author={Wu, Shirley and Yuksekgonul, Mert and Zhang, Linjun and Zou, James},
  booktitle={Proceedings of the 40th International Conference on Machine Learning},
  series={Proceedings of Machine Learning Research},
  volume={202},
  year={2023}
}

@inproceedings{meng2022locating,
  title={Locating and Editing Factual Associations in GPT},
  author={Meng, Kevin and Bau, David and Andonian, Alex and Belinkov, Yonatan},
  booktitle={Advances in Neural Information Processing Systems},
  volume={35},
  pages={17359--17372},
  year={2022}
}

@article{heimersheim2024activation,
  title={How to Use and Interpret Activation Patching},
  author={Heimersheim, Stefan and Nanda, Neel},
  journal={arXiv preprint arXiv:2404.15255},
  year={2024}
}

@inproceedings{palit2023visionlanguage,
  title={Towards Vision-Language Mechanistic Interpretability: A Causal Tracing Tool for BLIP},
  author={Palit, Vedant and Pandey, Rohan and Arora, Aryaman and Liang, Paul Pu},
  booktitle={Proceedings of the IEEE/CVF International Conference on Computer Vision Workshops},
  year={2023}
}

@inproceedings{lepori2024beyond,
  title={Beyond the Doors of Perception: Vision Transformers Represent Relations Between Objects},
  author={Lepori, Michael A. and Tartaglini, Alexa R. and Vong, Wai Keen and Serre, Thomas and Lake, Brenden M. and Pavlick, Ellie},
  booktitle={Advances in Neural Information Processing Systems},
  year={2024}
}

@article{huang2025causality,
  title={Causality $\neq$ Decodability, and Vice Versa: Lessons from Interpreting Counting ViTs},
  author={Huang, Lianghuan and Chang, Yingshan},
  journal={arXiv preprint arXiv:2510.09794},
  year={2025}
}

@inproceedings{li2026causaltracing,
  title={Causal Tracing of Object Representations in Large Vision Language Models: Mechanistic Interpretability and Hallucination Mitigation},
  author={Li, Qiming and Ye, Zekai and Feng, Xiaocheng and Zhong, Weihong and Ma, Weitao and Feng, Xiachong},
  booktitle={Proceedings of the AAAI Conference on Artificial Intelligence},
  year={2026}
}

@inproceedings{peng2026insideout,
  title={Inside-Out: Measuring Generalization in Vision Transformers Through Inner Workings},
  author={Peng, Yunxiang and Ma, Mengmeng and Yao, Ziyu and Peng, Xi},
  booktitle={Proceedings of the IEEE/CVF Conference on Computer Vision and Pattern Recognition},
  year={2026}
}

@article{zukowska2026seeing,
  title={Seeing Through Circuits: Faithful Mechanistic Interpretability for Vision Transformers},
  author={{\.Z}ukowska, Nina and Stammer, Wolfgang and Schiele, Bernt and Fischer, Jonas},
  journal={arXiv preprint arXiv:2604.14477},
  year={2026}
}

@inproceedings{ravfogel2022rlace,
  title={Linear Adversarial Concept Erasure},
  author={Ravfogel, Shauli and Twiton, Michael and Goldberg, Yoav and Cotterell, Ryan D.},
  booktitle={Proceedings of the 39th International Conference on Machine Learning},
  series={Proceedings of Machine Learning Research},
  volume={162},
  pages={18400--18421},
  year={2022}
}

@inproceedings{jiang2025interpreting,
  title={Interpreting and Editing Vision-Language Representations to Mitigate Hallucinations},
  author={Jiang, Nicholas and Kachinthaya, Anish and Petryk, Suzanne and Gandelsman, Yossi},
  booktitle={International Conference on Learning Representations},
  year={2025}
}

\section{Appendix}

\section*{A. Extended Related Work and Positioning}

The questions studied in this work lie at the intersection of shortcut
robustness, visual explanation, causal intervention on internal
representations, and latent representation editing. These areas share
several technical tools but differ in the quantities they seek to
identify. Shortcut-mitigation methods primarily optimize final
robustness; attribution methods identify influential input regions or
concepts; activation-patching methods test the behavioral contribution
of selected internal states; and concept-erasure methods remove
information recoverable by a specified representation family.

Inference Dynamics addresses a complementary process-level question:
after a visual cue has affected the computation, at what network depth
can its decision effect still be withdrawn by modifying the
representation aligned with the cue's original support? We review the
closest research lines below and clarify how the proposed
locality--withdrawability trajectory differs from their respective
targets.

\subsection*{A1. Shortcut Learning and Robust Classification}

Deep visual classifiers frequently exploit predictive regularities that
are easier to learn than the intended task semantics. Well-known
examples include texture bias, background correlation, co-occurring
objects, and acquisition-specific artifacts
\cite{c:1,r:1,c:2,xiao2020noisesignalroleimage}. Such shortcuts can
support high in-distribution accuracy while failing when the
correlation between task labels and nuisance attributes changes.

Early work on right-for-the-right-reasons learning constrains model
explanations so that predictions depend on intended evidence
\cite{c:3}. A broader line of robustness methods mitigates shortcut
learning through group-aware optimization, sample reweighting,
retraining, data augmentation, or representation regularization.
GroupDRO optimizes worst-group risk when group annotations are available
\cite{c:2}. JTT identifies difficult examples through an initial
training stage and upweights them during retraining
\cite{pmlr-v139-liu21f}. DFR retrains the final classifier layer on a
balanced retraining set \cite{kirichenko2022last}, while LLE addresses
multiple coexisting shortcuts through an ensemble of last-layer
classifiers \cite{Li_2023_CVPR}. Feature Sieve suppresses simple
features during training \cite{pmlr-v202-tiwari23a}, and subsequent
work has shown that implementation and fine-tuning details can
substantially affect group-robustness conclusions
\cite{NEURIPS2024_dc4db2ff}.

More recent methods provide increasingly interpretable or spatially
resolved shortcut supervision. Discover and Cure identifies unstable
human-interpretable concepts and modifies training to reduce reliance on
them \cite{wu2023discover}. RaVL discovers region-level spurious
correlations in vision--language models and mitigates them using a
region-aware objective \cite{NEURIPS2024_95c6ae3f}. Kuhn et al. detect
and suppress shortcut-related transformer representations without
requiring shortcut annotations \cite{Kuhn_2025_ICCV}.
Concept-Guided Fine-Tuning aligns ViT relevance with grounded semantic
concepts while suppressing spurious correlations
\cite{Elisha_2026_CVPR}.

These approaches improve a model so that it relies less on undesirable
evidence. Our objective is different from determining which
robustness-training strategy produces the strongest final classifier.
We analyze a frozen classifier and characterize how the controllability
of one validated visual cue changes during a single forward pass.
Existing mitigation methods typically apply one training objective, one
representation constraint, or one globally chosen intervention stage.
They do not determine, for each image--cue pair, when the cue's original
support ceases to control its accumulated prediction effect.

\subsection*{A2. Attribution, Concept Analysis, and Visual
Counterfactuals}

Visual explanation methods identify image regions, hidden units, or
human-interpretable concepts associated with a prediction. Grad-CAM
localizes class-sensitive regions through gradients flowing into
convolutional feature maps \cite{c:5}. Network Dissection associates
individual units with semantic concepts \cite{c:6}, while TCAV measures
directional sensitivity to user-defined concepts in hidden space
\cite{pmlr-v80-kim18d}. More recent feature-attribution frameworks
extend these ideas to settings in which standard gradients are
unavailable or unreliable \cite{cai2025gefa}.

Perturbation-based explanations instead modify an input region and
measure the resulting output change. Meaningful Perturbation searches
for a compact deletion mask whose removal changes the prediction
\cite{c:4}. Visual counterfactual explanations generate a semantically
plausible input whose prediction differs from that of the original
image. Region-constrained counterfactual methods further restrict the
edit to an intended spatial region, reducing unnecessary changes outside
the explanatory support \cite{sobieski2025rethinking}.

These methods answer two important questions: where evidence is
localized, and which input modification can alter the prediction.
Neither question determines whether the corresponding decision effect
remains controllable from the same spatial support after the cue has
entered the hidden computation. An attribution map may continue to
highlight a cue after that location has ceased to be sufficient for
controlling the prediction. Conversely, representation differences may
appear outside the original support without every such difference being
behaviorally relevant.

The cue-removed image in our framework is therefore not used merely as
a visual counterfactual explanation. It provides a matched reference
execution from which we construct layer-wise internal interventions.
The resulting trajectory measures how the relationship between spatial
concentration and decision-level control changes with network depth,
rather than only reporting the output difference produced by an input
edit.

\subsection*{A3. Causal Abstraction, Interchange Interventions, and
Activation Patching}

The closest methodological connection to layer-wise evidence withdrawal
is causal intervention on internal neural representations.
Causal-abstraction analysis tests whether variables in a neural network
realize the intervention behavior of a higher-level causal model
\cite{geiger2021causal}. Interchange intervention replaces a selected
representation in a base execution with the corresponding
representation produced by a source execution. Interchange Intervention
Training further uses these replacements as supervision for learning
networks whose internal variables follow a specified causal structure
\cite{geiger2022iit}.

Activation patching and causal tracing use a closely related operation
without necessarily assuming a complete high-level causal model.
Causal tracing restores selected hidden states from a clean execution
into a corrupted execution and measures recovery of a target behavior
\cite{meng2022locating}. Subsequent work has emphasized that the
interpretation of activation-patching results depends on how the source
and target executions are constructed, which component is patched, and
which output metric is measured \cite{heimersheim2024activation}.
These considerations motivate our use of matched image pairs,
context-matched hidden states, independent interventions at each layer,
and an explicit distinction between the operational measurement and its
broader interpretation.

Internal causal interventions have increasingly been adopted in vision
and multimodal models. Palit et al. introduce a causal-tracing tool for
BLIP and analyze how visual and linguistic information contributes to
multimodal predictions \cite{palit2023visionlanguage}. Mechanistic
analysis of ViTs has identified distinct perceptual and relational
structures in visual computation \cite{lepori2024beyond}. Work on
counting ViTs further shows that information decodability and causal use
can emerge at different depths \cite{huang2025causality}.

Causal Attribution via Activation Patching directly injects
patch-associated internal representations into a neutral visual context
to estimate the causal contribution of image patches
\cite{izadi2026caap}. Fine-grained causal tracing has also been used to
localize object-related effects in large vision--language models and to
modify intermediate representations for hallucination mitigation
\cite{li2026causaltracing}. Recent circuit-level approaches extend
activation patching from individual layers or tokens to structured
computational pathways. Inside-Out uses internal causal structure as a
signal for generalization under distribution shift
\cite{peng2026insideout}, while Vi-CD extracts edge-level circuits in
vision transformers for interpretation and steering
\cite{zukowska2026seeing}.

These studies establish internal activation replacement as a powerful
tool for visual attribution, mechanistic localization, and circuit
discovery. Our contribution is not the activation-replacement operation
itself. The distinction lies in the intervention question and the
quantity being measured.

Most activation-patching studies ask whether restoring or transplanting
an internal component recovers a target behavior, thereby identifying
where causally relevant information is stored or routed. Inference
Dynamics holds the visual cue support fixed and repeats a matched
withdrawal intervention independently across network depth. It asks when
that support stops being sufficient to undo an effect that has already
entered the computation. The output is therefore not only an
attribution score or causal circuit, but a depth-indexed trajectory of
local control and a sample-specific transition boundary.

Evidence Reset introduces a further distinction. Standard matched-state
patching requires the source execution whenever the intervention is
performed. Evidence Reset uses matched states only for offline
supervision. At inference, it predicts a pre-lock layer and reconstructs
the cue-free local state without executing the cue-removed branch.

\subsection*{A4. Layer-wise Shortcut Analysis, Representation, and
Reliance}

Several recent studies move beyond final predictions and examine how
shortcut information changes across network layers. Tsoy and
Konstantinov measure the layer-wise impact of synthetic image shortcuts
on deep model features \cite{tsoy2025measuring}. Kuhn et al. identify
internal shortcut concepts in transformers and use them for shortcut
detection and mitigation \cite{Kuhn_2025_ICCV}. Concept-based causal
diagnostics further distinguish whether a shortcut is represented from
whether intervening on it changes the prediction
\cite{grobner2026representation}.

An important conclusion emerging from this literature is that
representing a concept is not equivalent to relying on it. Information
may be accurately decoded from a late representation while being
functionally inactive, or may exert a causal effect before it becomes
easily linearly decodable \cite{huang2025causality}. Circuit-level
analyses similarly show that behavioral information can depend on
specific pathways rather than on the mere presence of a decodable
feature \cite{peng2026insideout,zukowska2026seeing}.

Inference Dynamics builds on this distinction but introduces a
different pair of measurements. Evidence locality measures the fraction
of the paired cue-induced representation difference that remains inside
the cue's mapped support. Evidence withdrawability measures whether
replacing that support is still sufficient to reproduce the
input-level cue-removal effect. Neither quantity is a probe score:
locality does not assert that every off-support difference is causally
used, and withdrawability does not require the cue to be linearly
decodable.

The joint trajectory reveals an intermediate regime that a single
layer-wise shortcut score does not capture. The cue-induced
representation difference can become less spatially concentrated while
the original support continues to control the final prediction. Stable
lock-in occurs only when this local control subsequently collapses and
remains low. This ordering identifies an actionable intervention
interval rather than only describing where shortcut information is
represented.

Circuit-discovery methods analyze internal computation across depth, but
their unit of explanation is typically a subgraph of attention heads,
MLPs, residual-stream components, or edges
\cite{peng2026insideout,zukowska2026seeing}. Our unit is an externally
specified visual cue and its mapped spatial support. Circuit discovery
asks which internal pathways are sufficient for a behavior; Inference
Dynamics asks how long a particular visual support remains capable of
withdrawing the behavior it induced. These perspectives are
complementary and could be combined by tracing the circuits through
which support-level control is lost.

\subsection*{A5. Concept Erasure and Latent Representation Editing}

Concept-erasure methods intervene directly on hidden representations to
remove information associated with a specified attribute. INLP
iteratively trains linear classifiers for the target attribute and
projects representations into their null spaces
\cite{ravfogel2020inlp}. R-LACE formulates linear concept erasure as an
adversarial projection problem \cite{ravfogel2022rlace}. LEACE derives
a closed-form transformation that prevents linear classifiers from
recovering the selected concept while minimizing representational
distortion \cite{belrose2023leace}.

Latent editing has also been applied to visual and multimodal
representations. Jiang et al. identify hallucination-related directions
in vision--language representations and remove those directions to
reduce hallucinated objects \cite{jiang2025interpreting}. Fine-grained
causal tracing and intermediate-representation injection similarly
modify internal multimodal states to alter object perception
\cite{li2026causaltracing}. Shortcut-mitigation methods such as RaVL,
transformer shortcut suppression, and Concept-Guided Fine-Tuning also
steer or regularize internal visual representations toward more reliable
evidence
\cite{NEURIPS2024_95c6ae3f,Kuhn_2025_ICCV,Elisha_2026_CVPR}.

Concept erasure and Evidence Reset differ in both their target and
geometry. Concept erasure asks whether a designated attribute remains
recoverable from the edited representation, commonly under a linear
probe family. Evidence Reset asks whether the decision effect caused by
one image cue has been removed from the downstream prediction. A
representation can become nearly non-decodable by a linear probe while
retaining a non-negligible decision residual because cue-induced effects
may be encoded nonlinearly, distributed across other units, or already
incorporated into downstream computation.

Most erasure transformations are fitted at a fixed representation and
apply a dataset-level concept direction. Evidence Reset instead
reconstructs image-conditioned states at the specified spatial support.
It preserves the original representation outside that support and
selects the intervention depth separately for each image--cue pair. Its
supervision is the representation produced by the same classifier on a
matched image without the cue, rather than a generic null space or
population-level mean.

The controlled latent-intervention comparison consequently evaluates
three distinct properties: post-intervention cue decodability,
task-level behavior, and residual distance to the cue-free prediction.
This separation prevents successful linear concept removal from being
interpreted automatically as complete withdrawal of decision reliance.

\subsection*{A6. Positioning of Inference Dynamics and Evidence Reset}

The closest research families can be organized by the question they
answer and by the output they produce.

\begin{table*}[t]
    \centering
    \small
    \setlength{\tabcolsep}{4pt}
    \renewcommand{\arraystretch}{1.15} 
    \caption{
        Positioning relative to closely related research families.
        ``Depth treatment'' describes whether network depth is usually
        a fixed analysis choice, a searched causal component, or the
        primary sample-specific object of study.
    }
    \begin{tabular}{@{} 
        p{0.15\textwidth} 
        p{0.22\textwidth} 
        p{0.22\textwidth} 
        p{0.16\textwidth} 
        p{0.21\textwidth} 
        @{}
    }
        \toprule
        \textbf{Research family}
        &
        \textbf{Primary question}
        &
        \textbf{Typical intervention or signal}
        &
        \textbf{Depth treatment}
        &
        \textbf{Primary output}
        \\
        \midrule

        Attribution and concept localization
        &
        Which pixels, regions, units, or concepts are associated with
        the prediction?
        &
        Gradients, relevance, perturbation, or concept sensitivity
        &
        Usually one or several selected representations
        &
        Saliency, attribution, or concept score
        \\

        Visual counterfactual explanation
        &
        What plausible input modification changes the prediction?
        &
        Input deletion, replacement, or generative editing
        &
        Internal depth is not the main object
        &
        Counterfactual image and output change
        \\

        Activation patching and causal tracing
        &
        Does restoring or transplanting an internal component recover a
        target behavior?
        &
        Source-to-target activation replacement
        &
        Layers or components are searched or compared
        &
        Causal contribution, localization, or circuit
        \\

        Concept erasure and latent editing
        &
        Can a specified concept be removed or the model behavior be
        steered?
        &
        Projection, direction removal, injection, or latent steering
        &
        Commonly fitted at a fixed representation
        &
        Reduced decodability or edited behavior
        \\

        Shortcut mitigation
        &
        Can training reduce reliance on spurious correlations?
        &
        Reweighting, retraining, augmentation, regularization, or
        relevance alignment
        &
        Usually fixed by method design
        &
        Improved average or worst-group robustness
        \\

        Inference Dynamics and Evidence Reset
        &
        When does the cue's original support cease to control its
        accumulated decision effect?
        &
        Matched local withdrawal followed by learned cue-free
        reconstruction
        &
        Full depth trajectory and image--cue-specific selection
        &
        Locality--withdrawability trajectory, lock-in onset, and
        adaptive reset
        \\

        \bottomrule
    \end{tabular}
\end{table*}
The relationship to prior work can therefore be stated precisely.
Inference Dynamics is not another input saliency method because its
object is the controllability of an internal decision effect rather than
the spatial importance of input pixels. It is not merely a visual
counterfactual method because the cue-removed image serves as a matched
source execution rather than as the final explanation.

At the level of intervention, Inference Dynamics belongs to the broader
activation-patching family. Its distinction is that support-level
withdrawability across depth is the central measurement, rather than
using patching only to localize a component or circuit. It is also
distinct from concept erasure because eliminating decodable cue
information is not equivalent to recovering the cue-free decision.

The methodological contribution is therefore composed of three linked
elements. First, we jointly trace where cue-induced representation
differences are concentrated and whether the cue's original support
continues to control the prediction. Second, we identify a stable,
sample-specific loss of support-level control and use it to define an
actionable pre-lock interval. Third, Evidence Reset converts the offline
matched-state analysis into a deployable intervention that predicts its
depth and reconstructs cue-free states without requiring the matched
reference during test-time inference.

\section*{B. Operational Setting and Definitions}

This section specifies the operational setting, terminology, and
information assumptions used by Inference Dynamics and Evidence Reset.
We study a controlled cue-intervention setting in which a candidate
visual cue and its spatial support are specified. This design isolates
the question of interest: how the influence of a validated cue evolves
across network depth, when its original support ceases to provide an
effective control point, and where an internal correction should be
applied.

\subsection*{B1. Image--Cue Pairs and Cue Validation}

Let \(f\) be an image classifier that maps an input image \(x\) to
class logits:

\begin{equation}
    \mathbf{z}(x)=f(x)\in\mathbb{R}^{C}.
\end{equation}

The class predicted from the original image is

\begin{equation}
    c=\arg\max_{k} z_k(x).
\end{equation}

For a candidate visual cue \(q\), let \(m_q\) denote its input-space
support and let \(x^{\setminus q}\) denote a matched reference in which
the cue is removed while the remaining image content is preserved as
closely as possible. The cue may correspond to a background region, a
co-occurring object, an object part, or another spatially identifiable
visual pattern.

We use the term \emph{evidence} in a model-relative sense. Cue \(q\) is
treated as evidence for prediction \(c\) when removing it reduces the
model's support for that prediction. Its input-level effect is

\begin{equation}
    \Delta_q(x)
    =
    z_c(x)-z_c\left(x^{\setminus q}\right).
\end{equation}

Only image--cue pairs satisfying

\begin{equation}
    \Delta_q(x)\geq \tau_{\mathrm{rel}}
\end{equation}

are retained for the inference-dynamics analysis. This relevance test
does not require the predicted label to change. A cue may contribute
substantially to the prediction while the remaining evidence is still
sufficient to preserve the final argmax. For training Evidence Reset,
we additionally use class-preserving pairs for which removal of the cue
does not alter the task label.

\subsection*{B2. Paired Representations and Local Evidence Withdrawal}

For each validated image--cue pair, the original image and its
cue-removed reference are processed independently by the same
classifier. Let \(f_{\leq l}\) denote the network up to candidate layer
\(l\). The corresponding paired representations are

\begin{equation}
    h_l^{+}=f_{\leq l}(x),
    \qquad
    h_l^{-}=f_{\leq l}\left(x^{\setminus q}\right).
\end{equation}

The input-space cue mask \(m_q\) is mapped to the representation units
of layer \(l\), yielding a binary or soft support mask \(M_l\). For a
convolutional network, \(M_l\) identifies spatial feature-map
locations. For a vision transformer, it identifies image tokens whose
input patches overlap the cue support. Units without a direct spatial
correspondence, such as a global classification token, are not included
in the mapped cue support.

Evidence withdrawal at layer \(l\) replaces only the representation
aligned with the original cue support:

\begin{equation}
    \widetilde{h}_l^{\,q}
    =
    \left(\mathbf{1}-M_l\right)\odot h_l^{+}
    +
    M_l\odot h_l^{-}.
\end{equation}

The modified representation is then propagated through the remaining
network:

\begin{equation}
    \mathbf{z}^{\mathrm{wd}}_l
    =
    f_{>l}\left(\widetilde{h}_l^{\,q}\right).
\end{equation}

Each candidate layer is intervened on independently. In particular, the
intervention at layer \(l+1\) is constructed from the intact paired
forward passes rather than from the output of the intervention at layer
\(l\). The resulting trajectory therefore measures the control
available at each network depth, rather than the cumulative effect of a
sequence of interventions.

The replacement is obtained from the same model processing a matched
image without the selected cue. It consequently provides a
context-matched state for the selected locations, in contrast to zero
ablation or noise injection, which may introduce states that do not
correspond to a plausible execution of the model.

\subsection*{B3. Evidence Locality}

We first measure how the representation difference associated with cue
removal is distributed across the units of each layer. Let

\begin{equation}
    \delta_l
    =
    h_l^{+}-h_l^{-}
\end{equation}

denote the paired representation difference. For representation unit
\(u\), its channel-wise magnitude is

\begin{equation}
    a_l(u)
    =
    \left\|
        \delta_l(u)
    \right\|_2.
\end{equation}

Let \(\mathcal{U}_l\) denote the set of representation units at layer
\(l\). Evidence locality is defined as the fraction of the total paired
representation difference that remains within the mapped cue support:

\begin{equation}
    L_l(q;x)
    =
    \frac{
        \displaystyle
        \sum_{u\in\mathcal{U}_l}
        M_l(u)a_l(u)
    }{
        \displaystyle
        \sum_{u\in\mathcal{U}_l}
        a_l(u)+\epsilon
    },
\end{equation}

where \(\epsilon\) is a small numerical-stability constant.

A high locality score indicates that the difference between the
cue-present and cue-removed executions remains concentrated near the
cue's original support. A lower score indicates that an increasing
fraction of the paired difference appears in other representation
units. Throughout the paper, describing a cue as becoming
\emph{less local} or \emph{more distributed} refers to this
representation-space concentration property.

Locality does not by itself measure whether the observed representation
differences determine the final decision. Differences outside the
original support may arise through receptive-field growth, attention,
normalization, or contextual interaction. Their decision-level
consequence is therefore assessed separately through evidence
withdrawability.

\subsection*{B4. Evidence Withdrawability}

Evidence withdrawability measures whether replacing the states aligned
with the cue's original support remains sufficient to reproduce the
prediction change caused by removing the cue at the input. It is defined
as

\begin{equation}
    W_l(q;x)
    =
    \frac{
        z_c(x)-z^{\mathrm{wd}}_{l,c}
    }{
        z_c(x)-z_c\left(x^{\setminus q}\right)
    }.
\end{equation}

Because the analysis retains only pairs satisfying
\(\Delta_q(x)\geq\tau_{\mathrm{rel}}\), the denominator is positive and
bounded away from zero by the relevance threshold.

A value close to one indicates that replacement at layer \(l\)
reproduces most of the input-level cue-removal effect. A value close to
zero indicates that modifying the original cue-aligned support has
little effect on the final prediction. Values outside \([0,1]\) are
retained during analysis. Values greater than one indicate an
intervention stronger than input-level cue removal, whereas negative
values indicate movement in the opposite direction.

Locality and withdrawability answer different questions. Locality asks
where the paired representation difference is concentrated.
Withdrawability asks whether the original cue support still controls
the cue-induced prediction effect. A cue may become less spatially
concentrated while remaining withdrawable, or it may remain detectable
near its original location after that location has ceased to control the
decision. The inference dynamics of an image--cue pair are therefore
represented by their joint layer-wise trajectory:

\begin{equation}
    \mathcal{D}_q(x)
    =
    \left\{
        \left(L_l(q;x),W_l(q;x)\right)
    \right\}_{l=0}^{L}.
\end{equation}

\subsection*{B5. Stable Evidence Lock-in}

We use stable evidence lock-in to denote a persistent loss of
withdrawability from the cue's original mapped support. The onset is
defined as

\begin{equation}
    l_{\mathrm{lock}}
    =
    \min
    \left\{
        l
        \ \middle|\
        W_j(q;x)\leq\tau_{\mathrm{wd}},
        \ 
        \forall j\in
        \{l,\ldots,l+r-1\}
    \right\},
\end{equation}

where \(\tau_{\mathrm{wd}}\) is the withdrawability threshold and \(r\)
is the required number of consecutive candidate layers.

The consecutive-layer condition distinguishes a stable transition from
an isolated fluctuation. No monotonicity assumption is imposed on
\(W_l\). If the threshold condition is not satisfied for \(r\)
consecutive candidate layers, the pair is categorized as not exhibiting
stable lock-in.

Operationally, \(l_{\mathrm{lock}}\) identifies the earliest depth after
which the cue's original mapped support no longer functions as an
effective local control interface under the matched-state replacement.
The definition concerns control from that support; it does not require
the cue representation to disappear, nor does it assume that every
possible intervention would fail at the same depth.

The onset definition uses only the layer-wise withdrawal trajectory.
Its practical value for correction is evaluated separately through the
outputs of the learned reset module, cue-free prediction residuals, and
task-performance measures. These outcomes are not used to determine
\(l_{\mathrm{lock}}\), keeping the definition of the transition
separate from the evaluation of whether acting before that transition
improves correction.

\subsection*{B6. Information Available During Analysis, Training, and Inference}

Inference Dynamics is used to construct process-level supervision,
whereas test-time Evidence Reset uses only the information available
from the original execution. The information flow is summarized below.

\begin{table}[t]
    \centering
    \small
    \setlength{\tabcolsep}{3pt}
    \renewcommand{\arraystretch}{1.08}
    \caption{Information used at different stages of the framework.
    A checkmark indicates that the information is available or used
    at the corresponding stage.}
    \begin{tabularx}{\columnwidth}{
        @{}
        >{\raggedright\arraybackslash}p{0.40\columnwidth}
        *{3}{>{\centering\arraybackslash}X}
        @{}
    }
        \toprule
        \textbf{Information}
        &
        \shortstack{\textbf{Dynamics}\\\textbf{analysis}}
        &
        \shortstack{\textbf{Module}\\\textbf{training}}
        &
        \textbf{Inference}
        \\
        \midrule

        Original image \(x\)
        & \checkmark & \checkmark & \checkmark
        \\

        Cue mask \(m_q\)
        & \checkmark & \checkmark & \checkmark
        \\

        Cue-removed reference \(x^{\setminus q}\)
        & \checkmark & \checkmark & --
        \\

        Paired hidden states
        & \checkmark & \checkmark & --
        \\

        Full withdrawal trajectory
        & \checkmark & \checkmark & --
        \\

        Dynamics-derived target layer
        & \checkmark & \checkmark & --
        \\

        Task label
        & -- & \checkmark & --
        \\

        Selector and reset module
        & -- & \textit{trained} & \checkmark
        \\

        \bottomrule
    \end{tabularx}
\end{table}

At inference, Evidence Reset receives the original image \(x\) and the
mask \(m_q\) of the cue under examination. The selector predicts either
a candidate reset layer or the no-reset option. If a reset layer is
selected, the reset module modifies the corresponding cue-aligned
states once, after which the unchanged downstream classifier completes
the forward pass.

The cue-removed image, paired execution, complete withdrawal scan, and
dynamics-derived target layer are used only for analysis or training
supervision. They are not required during test-time inference.

The cue mask is supplied by an external cue-localization source. In
Waterbirds, the background support is obtained from the provided bird
segmentation. In UrbanCars, the background and co-occurring-object
supports are available from the dataset construction. The framework
therefore isolates adaptive intervention-depth selection under a
specified candidate cue, rather than combining cue discovery and
internal correction into a single problem.

\subsection*{B7. Edge Cases and Multi-cue Convention}

Several trajectory patterns require explicit treatment.

\paragraph{No stable lock-in.}
If withdrawability does not remain below
\(\tau_{\mathrm{wd}}\) for \(r\) consecutive candidate layers, no
lock-in onset is assigned. Such cases are retained as a distinct
trajectory category rather than being assigned an artificial terminal
depth.

\paragraph{Early lock-in.}
If stable lock-in begins before any valid pre-lock candidate layer is
available, the pair does not provide a usable pre-lock reset interval.

\paragraph{Persistently high withdrawability.}
If \(W_l\) remains high throughout the network, the cue effect remains
controllable from its mapped support under the tested intervention.
These cases are distinguished from both stable-lock-in cases and pairs
that fail the initial relevance test.

\paragraph{Non-monotonic trajectories.}
Locality and withdrawability may fluctuate across adjacent layers. The
stable-onset criterion depends on persistence rather than on strict
monotonic decline.

\paragraph{Invalid mapped support.}
If the mapped cue support contains no valid spatial unit at a candidate
layer, that layer is excluded from the withdrawal trajectory. An empty
intervention is not interpreted as evidence of low withdrawability.

\paragraph{Multiple cues.}
The basic analysis unit is an image--cue pair. For an image containing
multiple specified cues \(q_1,\ldots,q_K\), the framework constructs a
separate trajectory for each cue:

\begin{equation}
    \mathcal{D}_{q_1}(x),
    \mathcal{D}_{q_2}(x),
    \ldots,
    \mathcal{D}_{q_K}(x).
\end{equation}

The resulting locality, withdrawability, lock-in onset, and reset layer
are cue-conditional. Unless a combined cue mask is explicitly used, a
single Evidence Reset execution acts on one specified cue. This
convention makes it possible to measure whether different shortcuts in
the same image exhibit different internal dynamics without forcing them
to share one intervention depth.

\section*{C. Complete Method and Implementation Details}

This section provides the complete procedure for constructing
inference-dynamics supervision, training the layer selector and reset
module, and applying Evidence Reset at inference. We first fine-tune an
ImageNet-pretrained classifier on each target dataset. The resulting
classifier is then frozen, and only the layer selector and reset module
are optimized during Evidence Reset training.

\subsection*{C1. Candidate Intervention Layers}

We define candidate intervention points at semantically complete block
boundaries. Let

\begin{equation}
    \mathcal{L}
    =
    \left\{
        l_1,l_2,\ldots,l_K
    \right\}
\end{equation}

denote the ordered candidate-layer set.

For ResNet-50, we use the output of the convolutional stem and the
output of each residual bottleneck block. ResNet-50 contains
\(3+4+6+3=16\) bottleneck blocks, resulting in

\begin{equation}
    K_{\mathrm{ResNet}}=17
\end{equation}

candidate intervention points. The candidate ordering is

\begin{equation}
    \mathcal{L}_{\mathrm{ResNet}}
    =
    \left\{
        \mathrm{stem},
        \mathrm{res2.1},\ldots,\mathrm{res2.3},
        \mathrm{res3.1},\ldots,\mathrm{res3.4},
        \mathrm{res4.1},\ldots,\mathrm{res4.6},
        \mathrm{res5.1},\ldots,\mathrm{res5.3}
    \right\}.
\end{equation}

For ViT-B/16, we use the patch-embedding output and the output of each
of the 12 transformer blocks, giving

\begin{equation}
    K_{\mathrm{ViT}}=13.
\end{equation}

The corresponding candidate set is

\begin{equation}
    \mathcal{L}_{\mathrm{ViT}}
    =
    \left\{
        \mathrm{embed},
        \mathrm{block1},
        \ldots,
        \mathrm{block12}
    \right\}.
\end{equation}

For an input \(x\), the representation at candidate layer \(l\) is

\begin{equation}
    h_l(x)
    =
    f_{\leq l}(x).
\end{equation}

For a validated image--cue pair, the paired candidate states are

\begin{equation}
    h_l^{+}
    =
    f_{\leq l}(x),
    \qquad
    h_l^{-}
    =
    f_{\leq l}\left(x^{\setminus q}\right).
\end{equation}

Candidate layers are indexed in forward order. The normalized layer
depth used for cross-architecture comparisons is

\begin{equation}
    d_l
    =
    \frac{k-1}{K-1},
    \qquad
    l=l_k.
\end{equation}

Thus, \(d_l=0\) corresponds to the earliest candidate state and
\(d_l=1\) corresponds to the final candidate block.

We intervene at block outputs rather than inside residual branches,
attention sublayers, or feed-forward sublayers. This provides a
consistent intervention interface and ensures that the unmodified
network suffix can be applied directly to every reset state.

\subsection*{C2. Layer-wise Mask Projection}

Let \(m_q\in\{0,1\}^{H\times W}\) denote the input-space cue mask. A
deterministic projection operator maps this mask to the representation
units of each candidate layer:

\begin{equation}
    \widetilde{M}_l
    =
    \mathcal{P}_l(m_q).
\end{equation}

For ResNet-50, the input mask is resized to the spatial resolution of
the candidate feature map using area interpolation. If the feature map
at layer \(l\) has spatial resolution \(H_l\times W_l\), then

\begin{equation}
    \widetilde{M}_l
    =
    \operatorname{Resize}_{\mathrm{area}}
    \left(
        m_q,
        H_l,
        W_l
    \right).
\end{equation}

The resized values represent the fraction of each feature-map cell
covered by the input cue. The intervention mask is obtained using

\begin{equation}
    M_l(u)
    =
    \mathbf{1}
    \left[
        \widetilde{M}_l(u)
        \geq
        \eta_{\mathrm{mask}}
    \right],
\end{equation}

where

\begin{equation}
    \eta_{\mathrm{mask}}=0.25.
\end{equation}

For ViT-B/16, the input image is partitioned into non-overlapping
\(16\times16\) patches. For patch token \(u\), the overlap ratio is

\begin{equation}
    \widetilde{M}_l(u)
    =
    \frac{
        \left|
            \mathcal{R}_u
            \cap
            \operatorname{supp}(m_q)
        \right|
    }{
        \left|
            \mathcal{R}_u
        \right|
    },
\end{equation}

where \(\mathcal{R}_u\) is the input region corresponding to token
\(u\). A token is included in the cue support when

\begin{equation}
    M_l(u)
    =
    \mathbf{1}
    \left[
        \widetilde{M}_l(u)
        \geq
        0.25
    \right].
\end{equation}

The class token is assigned

\begin{equation}
    M_l(\mathrm{CLS})=0.
\end{equation}

It may contribute to the total paired representation difference used in
the locality denominator, but it is never treated as part of the
original spatial cue support.

Mask values are broadcast across representation channels during feature
replacement. A candidate layer is valid for an image--cue pair only
when

\begin{equation}
    \left\|M_l\right\|_1>0.
\end{equation}

Layers with empty projected support are excluded from that pair's
trajectory. They are not assigned a zero withdrawability score.

\subsection*{C3. Dynamics-derived Reset Targets}

For layer selection, locality and withdrawability are clipped to the
interval \([0,1]\):

\begin{equation}
    \overline{L}_l
    =
    \operatorname{clip}
    \left(
        L_l,
        0,
        1
    \right),
\end{equation}

\begin{equation}
    \overline{W}_l
    =
    \operatorname{clip}
    \left(
        W_l,
        0,
        1
    \right).
\end{equation}

Their unclipped values are retained for analysis and reporting. The
resetability score is

\begin{equation}
    \rho_l(q;x)
    =
    \left(
        1-\overline{L}_l(q;x)
    \right)
    \overline{W}_l(q;x).
\end{equation}

A high resetability score identifies a candidate layer at which the
paired difference is no longer highly concentrated within the original
cue support, while modifying that support still produces a substantial
prediction change.

For an example with stable lock-in onset \(l_{\mathrm{lock}}\), the
valid pre-lock candidate set is

\begin{equation}
    \mathcal{L}_{\mathrm{pre}}
    =
    \left\{
        l\in\mathcal{L}
        \ \middle|\
        l<l_{\mathrm{lock}},
        \ 
        \left\|M_l\right\|_1>0
    \right\}.
\end{equation}

The target layer is

\begin{equation}
    l^{\star}
    =
    \arg\max_{l\in\mathcal{L}_{\mathrm{pre}}}
    \rho_l(q;x).
\end{equation}

To avoid forcing a reset when the trajectory contains no sufficiently
informative intervention point, we require

\begin{equation}
    \max_{l\in\mathcal{L}_{\mathrm{pre}}}
    \rho_l(q;x)
    \geq
    \tau_{\mathrm{act}},
\end{equation}

where

\begin{equation}
    \tau_{\mathrm{act}}=0.10.
\end{equation}

The selector output space contains the candidate layers and an
additional no-reset class:

\begin{equation}
    \mathcal{Y}_{\mathrm{sel}}
    =
    \mathcal{L}
    \cup
    \left\{
        \varnothing
    \right\}.
\end{equation}

The target selector label is

\begin{equation}
    y_{\mathrm{sel}}
    =
    \begin{cases}
        l^{\star},
        &
        \begin{aligned}
        &\text{if stable lock-in exists, }
        \mathcal{L}_{\mathrm{pre}}\neq\emptyset,\\
        &\text{and }
        \max_{l\in\mathcal{L}_{\mathrm{pre}}}
        \rho_l
        \geq
        \tau_{\mathrm{act}},
        \end{aligned}
        \\[3mm]
        \varnothing,
        &
        \text{otherwise}.
    \end{cases}
\end{equation}

The no-reset label is therefore assigned in four cases: no stable
lock-in is detected, lock-in occurs before a valid pre-lock candidate,
the projected mask is invalid at all pre-lock candidates, or the maximum
resetability score is below the actionability threshold.

\subsection*{C4. Layer Selector}

The selector is a lightweight network operating directly on the
original image and cue mask. The mask is concatenated with the RGB image
as a fourth input channel:

\begin{equation}
    x_{\mathrm{sel}}
    =
    \operatorname{Concat}
    \left(
        x,
        m_q
    \right)
    \in
    \mathbb{R}^{4\times H\times W}.
\end{equation}

The selector consists of four convolutional blocks with output widths

\begin{equation}
    32,\ 64,\ 128,\ 256.
\end{equation}

Each block contains a \(3\times3\) convolution with stride 2, group
normalization, and GELU activation. Global average pooling produces a
256-dimensional representation:

\begin{equation}
    v_{\mathrm{sel}}
    =
    \operatorname{GAP}
    \left(
        g_{\phi}(x_{\mathrm{sel}})
    \right).
\end{equation}

The pooled representation is passed through a two-layer prediction head
with hidden dimension 256 and dropout rate 0.1:

\begin{equation}
    s_{\mathrm{sel}}
    =
    W_2
    \operatorname{GELU}
    \left(
        W_1v_{\mathrm{sel}}+b_1
    \right)
    +
    b_2.
\end{equation}

The final output dimension is \(K+1\). The corresponding selection
distribution is

\begin{equation}
    \pi_{\phi}
    \left(
        l\mid x,m_q
    \right)
    =
    \operatorname{softmax}
    \left(
        s_{\mathrm{sel}}
    \right)_l.
\end{equation}

The predicted layer is

\begin{equation}
    \widehat{l}
    =
    \arg\max_{l\in\mathcal{Y}_{\mathrm{sel}}}
    \pi_{\phi}
    \left(
        l\mid x,m_q
    \right).
\end{equation}

The selector contains approximately \(0.46\) million trainable
parameters. A separate final prediction head is used for ResNet-50 and
ViT-B/16 because the two architectures contain different numbers of
candidate layers; the convolutional encoder is otherwise identical.

The selector objective is

\begin{equation}
    \mathcal{L}_{\mathrm{sel}}
    =
    -
    \log
    \pi_{\phi}
    \left(
        y_{\mathrm{sel}}
        \mid
        x,m_q
    \right).
\end{equation}

The selector receives no cue-removed image, paired hidden state, or
withdrawability score as input.

\subsection*{C5. Reset Module for ResNet-50}

ResNet candidate representations have different channel dimensions
across stages. We therefore use stage-specific input and output
projections together with a shared reconstruction core. Candidate
blocks within the same ResNet stage share the same projection pair.

For a representation with \(C_l\) channels, the input projection is

\begin{equation}
    v_l
    =
    P_l^{\mathrm{in}}
    \left[
        \left(
            \mathbf{1}-M_l
        \right)
        \odot
        h_l^{+}
    \right],
\end{equation}

where \(P_l^{\mathrm{in}}\) is a \(1\times1\) convolution mapping
\(C_l\) channels to a shared hidden width

\begin{equation}
    d_{\mathrm{reset}}=256.
\end{equation}

The projected mask is concatenated as an additional spatial channel,
and a learned layer embedding is broadcast over the feature map:

\begin{equation}
    v_l^{\mathrm{in}}
    =
    \operatorname{Concat}
    \left(
        v_l,
        M_l,
        E_l
    \right).
\end{equation}

The shared reconstruction core contains two residual convolutional
blocks. Each block uses two \(3\times3\) convolutions, group
normalization, and GELU activation. The core therefore accesses
surrounding unmasked features through a finite local receptive field
while preserving the spatial resolution of the selected layer.

The reconstructed cue-free state is

\begin{equation}
    \widehat{h}_l^{-}
    =
    P_l^{\mathrm{out}}
    \left(
        R_{\theta}^{\mathrm{CNN}}
        \left(
            v_l^{\mathrm{in}}
        \right)
    \right),
\end{equation}

where \(P_l^{\mathrm{out}}\) is a \(1\times1\) convolution mapping the
shared hidden width back to \(C_l\) channels.

The stem, res2, res3, res4, and res5 candidates use five
stage-specific projection pairs. The reconstruction core is shared
across all 17 candidate layers. The complete ResNet reset module
contains approximately \(4.4\) million trainable parameters.

\subsection*{C6. Reset Module for ViT-B/16}

All ViT-B/16 candidate representations use the same embedding dimension

\begin{equation}
    C_l=768.
\end{equation}

A shared linear projection maps each token to the reset hidden
dimension:

\begin{equation}
    v_l
    =
    P^{\mathrm{in}}
    \left[
        \left(
            \mathbf{1}-M_l
        \right)
        \odot
        h_l^{+}
    \right]
    \in
    \mathbb{R}^{(N+1)\times256}.
\end{equation}

A learned mask embedding is added to masked image tokens, and a learned
layer embedding is added to all tokens:

\begin{equation}
    v_l^{\mathrm{in}}
    =
    v_l
    +
    M_l\odot e_{\mathrm{mask}}
    +
    e_l.
\end{equation}

The class token is retained as contextual input but is never included in
the replacement mask.

The shared reconstruction core consists of two transformer encoder
blocks with four attention heads, hidden dimension 256, and feed-forward
dimension 1024. Each block uses pre-layer normalization, multi-head
self-attention, a two-layer GELU feed-forward network, and residual
connections.

The reconstructed representation is

\begin{equation}
    \widehat{h}_l^{-}
    =
    P^{\mathrm{out}}
    \left(
        R_{\theta}^{\mathrm{ViT}}
        \left(
            v_l^{\mathrm{in}}
        \right)
    \right),
\end{equation}

where \(P^{\mathrm{out}}\) maps the 256-dimensional reset state back to
768 dimensions. The ViT reset module contains approximately \(2.0\)
million trainable parameters and is shared across all candidate
transformer layers.

\subsection*{C7. Reset State Construction}

For either architecture, the reset module receives a representation in
which the cue-aligned states have been removed. It therefore cannot
directly copy the states that it is trained to replace.

The complete reset representation is

\begin{equation}
    h_l^{\mathrm{reset}}
    =
    \left(
        \mathbf{1}-M_l
    \right)
    \odot
    h_l^{+}
    +
    M_l
    \odot
    \widehat{h}_l^{-}.
\end{equation}

All unmasked states are copied exactly from the original execution. The
modified representation is propagated through the frozen network
suffix:

\begin{equation}
    \mathbf{z}^{\mathrm{reset}}_l
    =
    f_{>l}
    \left(
        h_l^{\mathrm{reset}}
    \right).
\end{equation}

During training, the reset module is applied at the ground-truth
dynamics-derived layer \(l^{\star}\), rather than at the selector's
current prediction. This teacher-forcing strategy avoids unstable
training caused by early selector errors and keeps the reset target
well-defined. At inference, the predicted layer \(\widehat{l}\) is used.

\subsection*{C8. Feature Reconstruction Objective}

For a training pair with a valid reset target, the cue-removed branch
provides the target hidden representation. Let
\(\operatorname{sg}(\cdot)\) denote stop-gradient. We use the masked
mean absolute reconstruction error:

\begin{equation}
    \mathcal{L}_{\mathrm{feat}}
    =
    \frac{
        \left\|
            M_{l^{\star}}
            \odot
            \left(
                \widehat{h}_{l^{\star}}^{-}
                -
                \operatorname{sg}
                \left(
                    h_{l^{\star}}^{-}
                \right)
            \right)
        \right\|_1
    }{
        C_{l^{\star}}
        \left\|M_{l^{\star}}\right\|_1
        +
        \epsilon
    },
\end{equation}

where

\begin{equation}
    \epsilon=10^{-8}.
\end{equation}

The denominator normalizes over both selected representation units and
channels, preventing cue size or feature dimensionality from changing
the relative scale of the loss.

\subsection*{C9. Prediction Alignment Objective}

The cue-free prediction distribution is

\begin{equation}
    p^{-}
    =
    \operatorname{softmax}
    \left(
        \mathbf{z}\left(x^{\setminus q}\right)
    \right),
\end{equation}

and the reset prediction distribution is

\begin{equation}
    p^{\mathrm{reset}}
    =
    \operatorname{softmax}
    \left(
        \mathbf{z}^{\mathrm{reset}}_{l^{\star}}
    \right).
\end{equation}

The prediction-alignment loss is

\begin{equation}
    \mathcal{L}_{\mathrm{pred}}
    =
    D_{\mathrm{KL}}
    \left(
        \operatorname{sg}
        \left(
            p^{-}
        \right)
        \,\middle\|\,
        p^{\mathrm{reset}}
    \right).
\end{equation}

The cue-free branch is treated as a fixed target and receives no
gradient.

\subsection*{C10. Classification-preservation Objective}

The reset output is additionally supervised using the task label:

\begin{equation}
    \mathcal{L}_{\mathrm{cls}}
    =
    \operatorname{CE}
    \left(
        \mathbf{z}^{\mathrm{reset}}_{l^{\star}},
        y
    \right).
\end{equation}

This term discourages the reset module from matching the cue-free branch
at the expense of task-relevant visual information.

\subsection*{C11. Complete Training Objective}

Let

\begin{equation}
    I_{\mathrm{reset}}
    =
    \mathbf{1}
    \left[
        y_{\mathrm{sel}}
        \neq
        \varnothing
    \right]
\end{equation}

indicate whether a valid reset target exists. The complete objective is

\begin{equation}
    \mathcal{L}
    =
    \lambda_{\mathrm{sel}}
    \mathcal{L}_{\mathrm{sel}}
    +
    I_{\mathrm{reset}}
    \left(
        \mathcal{L}_{\mathrm{cls}}
        +
        \lambda_{\mathrm{feat}}
        \mathcal{L}_{\mathrm{feat}}
        +
        \lambda_{\mathrm{pred}}
        \mathcal{L}_{\mathrm{pred}}
    \right).
\end{equation}

We use

\begin{equation}
    \lambda_{\mathrm{feat}}=1.0,
    \qquad
    \lambda_{\mathrm{pred}}=1.0,
    \qquad
    \lambda_{\mathrm{sel}}=1.0.
\end{equation}

No-reset examples supervise the selector but do not update the reset
module. This prevents the reconstruction network from being trained at
an arbitrary layer when no reliable reset target is available.

The pretrained classifier remains frozen. Gradients are applied only to
the selector, reset module, projection layers, and layer embeddings.

\subsection*{C12. Thresholds Used for Dynamics Construction}

The cue-relevance threshold is

\begin{equation}
    \tau_{\mathrm{rel}}=0.5.
\end{equation}

This threshold is applied to the original-class logit difference between
the cue-present and cue-removed images.

Stable lock-in is detected using

\begin{equation}
    \tau_{\mathrm{wd}}=0.20
\end{equation}

and

\begin{equation}
    r=3
\end{equation}

consecutive candidate layers. The actionability threshold used for
no-reset assignment is

\begin{equation}
    \tau_{\mathrm{act}}=0.10.
\end{equation}

All threshold values are selected on the validation split and then held
fixed for test evaluation.

\subsection*{C13. Backbone Fine-tuning}

All images are resized and cropped to \(224\times224\). During training,
we use random resized cropping and horizontal flipping. At validation
and test time, we use deterministic resizing and center cropping.

For ResNet-50, the ImageNet-pretrained classifier is fine-tuned using
stochastic gradient descent with momentum:

\begin{equation}
    \mathrm{lr}=10^{-3},
    \qquad
    \mathrm{momentum}=0.9,
    \qquad
    \mathrm{weight\ decay}=10^{-4}.
\end{equation}

The model is trained for 30 epochs with a cosine learning-rate schedule
and batch size 64.

For ViT-B/16, we use AdamW with

\begin{equation}
    \mathrm{lr}=5\times10^{-5},
    \qquad
    \mathrm{weight\ decay}=0.05.
\end{equation}

The model is trained for 20 epochs with two warm-up epochs, cosine
learning-rate decay, and batch size 64. The checkpoint with the best
validation worst-group accuracy is retained.

\subsection*{C14. Selector and Reset-module Optimization}

The selector and reset module are trained jointly using AdamW with

\begin{equation}
    \mathrm{lr}=10^{-4},
    \qquad
    \mathrm{weight\ decay}=10^{-4}.
\end{equation}

Training uses batch size 64 for 40 epochs. The learning rate is warmed
up linearly for the first five epochs and then decayed using a cosine
schedule. Gradients are clipped to a maximum global norm of

\begin{equation}
    1.0.
\end{equation}

Mixed-precision training is used. Model selection is based on validation
worst-group accuracy, with normalized cue residual used as the
secondary criterion when two checkpoints have the same worst-group
accuracy.

We use three random seeds:

\begin{equation}
    \left\{
        0,1,2
    \right\}.
\end{equation}

Training is performed on one NVIDIA A100 GPU with 40 GB of memory.

\subsection*{C15. Training Procedure}

Training consists of three stages.

\paragraph{Backbone adaptation.}
The ImageNet-pretrained classifier is fine-tuned on the target training
split. The selected checkpoint is frozen for all subsequent stages.

\paragraph{Dynamics construction.}
For each validated training pair, the cue-present and cue-removed
images are processed by the frozen classifier. Hidden states are cached
at all candidate layers. Every candidate layer is independently
intervened on to compute \(L_l\), \(W_l\), stable lock-in onset,
resetability scores, and the selector target.

The dynamics labels, valid-layer masks, and target reset layers are
precomputed once and reused throughout module training. Cue-free hidden
states at the selected target layers are cached in half precision to
reduce storage cost.

\paragraph{Module optimization.}
The selector is trained on every validated pair. For pairs with a valid
reset target, the reset module is applied at \(l^{\star}\) and optimized
using feature reconstruction, prediction alignment, and classification
preservation. For no-reset pairs, only the selector loss is evaluated.

\subsection*{C16. Inference Procedure}

At inference, the selector first processes the original image and cue
mask:

\begin{equation}
    \widehat{l}
    =
    \arg\max_{l\in\mathcal{Y}_{\mathrm{sel}}}
    \pi_{\phi}
    \left(
        l\mid x,m_q
    \right).
\end{equation}

Because the selector operates directly on the image and mask, its
decision is available before the main classifier forward pass.

When

\begin{equation}
    \widehat{l}
    =
    \varnothing,
\end{equation}

the classifier performs its standard forward pass:

\begin{equation}
    \mathbf{z}^{\mathrm{out}}
    =
    f(x).
\end{equation}

Otherwise, the classifier is executed up to the selected layer, the
cue-aligned representation is reset once, and inference continues
through the remaining network:

\begin{equation}
    \mathbf{z}^{\mathrm{out}}
    =
    f_{>\widehat{l}}
    \left(
        h_{\widehat{l}}^{\mathrm{reset}}
    \right).
\end{equation}

The cue-removed image, paired hidden states, complete layer sweep, and
ground-truth lock-in depth are not used at inference.

\subsection*{C17. Computational Cost}

For each image--cue pair, offline dynamics construction requires two
complete forward passes and one suffix execution for each candidate
layer. Let \(C_f\) denote the cost of a full classifier forward pass and
\(C_{>l}\) the cost of the suffix after layer \(l\). The approximate
cost is

\begin{equation}
    C_{\mathrm{dyn}}
    =
    2C_f
    +
    \sum_{l\in\mathcal{L}}
    C_{>l}.
\end{equation}

The network prefixes are not recomputed for each intervention because
the paired candidate states are cached.

At inference, the approximate cost is

\begin{equation}
    C_{\mathrm{test}}
    =
    C_f
    +
    C_{\pi}
    +
    \mathbf{1}
    \left[
        \widehat{l}\neq\varnothing
    \right]
    C_R(\widehat{l}),
\end{equation}

where \(C_{\pi}\) is the selector cost and \(C_R(\widehat{l})\) is the
cost of one reset-module execution. No layer-wise scan or cue-free
branch is required at test time.

\section*{D. Datasets, Cue Construction, and Pair Validation}

\begin{figure*}[t]
    \centering
    \includegraphics[
        width=\textwidth
    ]{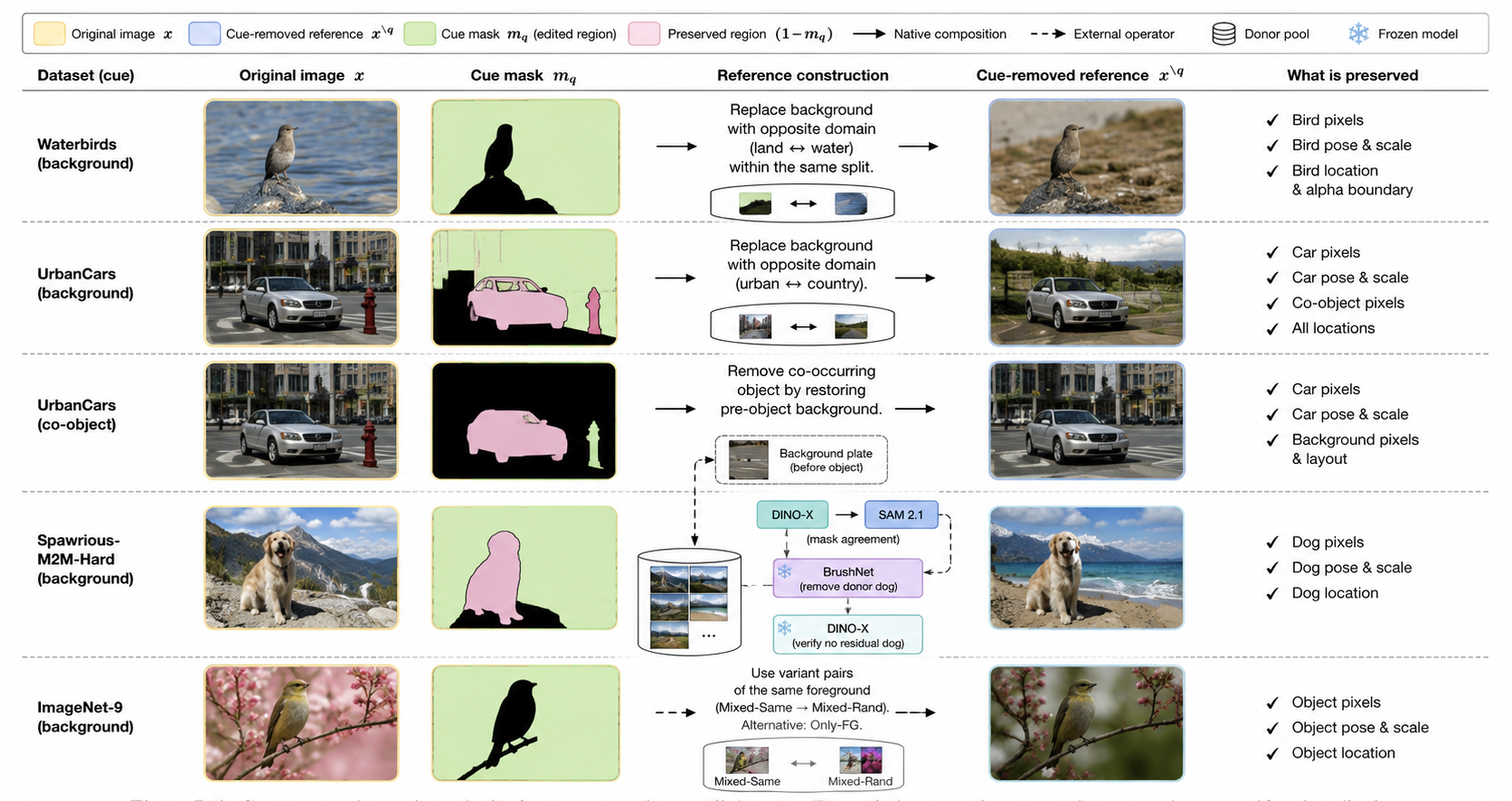}
    \caption{
        Cue construction and matched-reference generation across
        datasets. For each dataset and cue type, the original image
        \(x\) is paired with a cue-removed reference
        \(x^{\setminus q}\). The cue mask \(m_q\) specifies the region
        modified by the reference-construction procedure, while the
        task-relevant foreground pixels, pose, scale, and spatial
        configuration are preserved. Waterbirds and UrbanCars use
        native compositing information; Spawrious-M2M-Hard uses
        cross-model foreground localization and verified background
        completion; ImageNet-9 uses matched benchmark variants with the
        same foreground.
    }
    \label{fig:supp_cue_construction}
\end{figure*}

We evaluate Inference Dynamics and Evidence Reset on datasets in which
the target visual cue can be spatially identified and modified while
preserving the task-relevant object. Waterbirds and UrbanCars serve as
the primary controlled benchmarks. Spawrious-M2M-Hard and the
ImageNet-9 Background Challenge provide additional evaluation under
more varied foregrounds, backgrounds, and cue-construction procedures. Figure~\ref{fig:supp_cue_construction} summarizes the cue
definitions, matched-reference construction procedures, and preserved
image components across all datasets.

The basic analysis unit is an image--cue pair rather than an image
alone. An image containing one specified shortcut contributes one
candidate pair. An UrbanCars image contributes separate
background-cue and co-occurring-object-cue pairs. Dataset splits,
source--reference assignments, and cue masks are fixed before
layer-wise analysis and are shared across architectures and random
seeds.

\subsection*{D1. Dataset Overview}

Waterbirds contains two target classes, landbird and waterbird, and a
binary background attribute, land or water. UrbanCars contains two
target classes, urban car and country car, together with two
independently controlled shortcut attributes: background type and
co-occurring-object type. Spawrious-M2M-Hard contains four dog breeds
shown in six environments under a many-to-many class--background
correlation. ImageNet-9 groups ImageNet categories into nine
superclasses and provides controlled foreground--background
compositions.

The primary dataset sizes and candidate cue types are summarized below.

\begin{table}[t]
    \centering
    \small
    \setlength{\tabcolsep}{3pt}
    \renewcommand{\arraystretch}{1.08}
    \caption{Datasets and cue types used in the experiments. ``Official''
    indicates that we retain the split distributed with the benchmark.}
    \begin{tabularx}{\columnwidth}{
        @{}
        >{\raggedright\arraybackslash}p{0.22\columnwidth}
        >{\centering\arraybackslash}p{0.13\columnwidth}
        >{\centering\arraybackslash}p{0.13\columnwidth}
        >{\centering\arraybackslash}p{0.13\columnwidth}
        >{\raggedright\arraybackslash}X
        @{}
    }
        \toprule
        \textbf{Dataset}
        &
        \textbf{Train}
        &
        \textbf{Val.}
        &
        \textbf{Test}
        &
        \textbf{Target cue}
        \\
        \midrule

        Waterbirds
        & 4,795
        & 1,199
        & 5,794
        & Background
        \\

        UrbanCars
        & 8,000
        & 1,000
        & 1,000
        & Background; co-object
        \\

        Spawrious-M2M-Hard
        & Official
        & Official
        & Official
        & Background environment
        \\

        ImageNet-9
        & --
        & --
        & Official
        & Background
        \\

        \bottomrule
    \end{tabularx}
\end{table}

Waterbirds and UrbanCars are used for training and evaluating Evidence
Reset. Spawrious-M2M-Hard and ImageNet-9 are used to examine whether
the measured inference dynamics persist beyond the native compositing
setting of the two primary benchmarks.

\subsection*{D2. Waterbirds}

Waterbirds is constructed from foreground birds in CUB-200-2011 and
background images from Places. Bird species are grouped into landbird
and waterbird classes. Each segmented bird is placed on either a land
background or a water background. In the training split, 95\% of
waterbirds appear on water backgrounds and 95\% of landbirds appear on
land backgrounds. Validation and test splits contain more balanced
combinations of target class and background type.

Let \(m_{\mathrm{fg}}\) denote the provided bird segmentation mask. The
background cue mask is defined as

\begin{equation}
    m_q
    =
    \mathbf{1}
    -
    m_{\mathrm{fg}}.
\end{equation}

The cue therefore includes pixels belonging to the composited
background, while the complete bird foreground remains outside the
intervention support.

For the primary matched-reference operator, we use the native
compositing pipeline. The original foreground bird, including its
location, scale, appearance, and alpha boundary, is kept fixed. Its
original background is replaced by a background sampled from the
opposite background category within the same dataset split. A land
background is therefore replaced by a water background, and a water
background is replaced by a land background.

Let \(F_i\) denote the foreground bird layer, \(A_i\) its alpha mask,
and \(B_i\) its original background. The original image is

\begin{equation}
    x_i
    =
    A_i\odot F_i
    +
    \left(
        \mathbf{1}-A_i
    \right)
    \odot B_i.
\end{equation}

For a donor background \(B_j\) from the opposite background category,
the matched reference is

\begin{equation}
    x_i^{\setminus q}
    =
    A_i\odot F_i
    +
    \left(
        \mathbf{1}-A_i
    \right)
    \odot B_j.
\end{equation}

The donor background is sampled independently of the bird class, is
different from the original background image, and belongs to the same
training, validation, or test split as the source image. Pair
assignments are generated once using a fixed random seed and reused
across architectures.

This construction removes the original background cue without altering
the foreground pixels. It also avoids blacking out or blurring the
background, which would introduce an artificial input domain not
encountered during classifier training.

For each Waterbirds image, we construct one background-cue pair. The
semantic target remains the bird class, independent of the replacement
background.

\subsection*{D3. UrbanCars}

UrbanCars is a binary classification dataset with urban car and country
car as the target classes. Each image is constructed from three
separable visual components:

\begin{equation}
    x
    =
    \mathcal{C}
    \left(
        F_{\mathrm{car}},
        B,
        O
    \right),
\end{equation}

where \(F_{\mathrm{car}}\) is the target car foreground, \(B\) is the
background, \(O\) is a co-occurring object, and
\(\mathcal{C}(\cdot)\) denotes the native compositing process.

During training, the background type agrees with the target car class
with probability \(0.95\). The co-occurring-object type independently
agrees with the target class with probability \(0.95\). Consequently,
the fraction of training images for which both shortcuts agree with the
target class is

\begin{equation}
    0.95^2
    =
    0.9025.
\end{equation}

The dataset contains 4,000 training images for each target class,
yielding 8,000 training images in total. The validation and test sets
each contain 1,000 images and cover combinations of target class,
background type, and co-occurring-object type.

\paragraph{Background-cue pairs.}

For the background analysis, the target cue mask is the complement of
the union of the car and co-occurring-object masks:

\begin{equation}
    m_q^{\mathrm{bg}}
    =
    \mathbf{1}
    -
    \max
    \left(
        m_{\mathrm{car}},
        m_{\mathrm{obj}}
    \right).
\end{equation}

The maximum is taken elementwise. The matched reference keeps the car
and co-occurring object unchanged and replaces only the background
layer. The donor background is sampled from the opposite background
domain and from the same dataset split:

\begin{equation}
    x^{\setminus q_{\mathrm{bg}}}
    =
    \mathcal{C}
    \left(
        F_{\mathrm{car}},
        B_{\mathrm{opp}},
        O
    \right).
\end{equation}

The car mask, car pose, car scale, co-occurring-object identity, and
co-occurring-object location are identical in the original and
reference images.

\paragraph{Co-occurring-object-cue pairs.}

For the co-occurring-object analysis, the cue mask is the alpha support
used to insert the object:

\begin{equation}
    m_q^{\mathrm{obj}}
    =
    m_{\mathrm{obj}}.
\end{equation}

Because UrbanCars is generated compositionally, the background layer
beneath the co-occurring object is available before insertion. We
remove the object by restoring the corresponding pixels from the
pre-object background plate:

\begin{equation}
    x^{\setminus q_{\mathrm{obj}}}
    =
    \mathcal{C}
    \left(
        F_{\mathrm{car}},
        B,
        \varnothing
    \right).
\end{equation}

This construction changes neither the car foreground nor the
background. It also avoids generative inpainting in the primary
UrbanCars experiments.

Each UrbanCars image consequently produces two independently validated
image--cue pairs. Background and co-occurring-object trajectories are
computed separately even when both shortcuts are simultaneously aligned
with the target class.

\paragraph{Joint-cue references.}

For the multi-cue analysis, a joint reference is constructed by
replacing the background and removing the co-occurring object in the
same image:

\begin{equation}
    x^{\setminus
    \left\{
        q_{\mathrm{bg}},
        q_{\mathrm{obj}}
    \right\}}
    =
    \mathcal{C}
    \left(
        F_{\mathrm{car}},
        B_{\mathrm{opp}},
        \varnothing
    \right).
\end{equation}

The corresponding joint mask is

\begin{equation}
    m_q^{\mathrm{joint}}
    =
    \max
    \left(
        m_q^{\mathrm{bg}},
        m_q^{\mathrm{obj}}
    \right).
\end{equation}

Single-cue pairs remain the primary unit used to construct
cue-conditional dynamics and reset-layer supervision.

\subsection*{D4. Spawrious-M2M-Hard}

Spawrious is a synthetic yet photorealistic benchmark containing four
dog breeds in six background environments. M2M-Hard introduces a
many-to-many correlation in which subsets of breeds are associated with
subsets of environments during training. We retain the official
training, validation, and out-of-distribution test splits without
rebalancing them.

Unlike Waterbirds and UrbanCars, the released Spawrious images do not
provide separate foreground and background layers. We therefore
construct foreground masks using a cross-model open-world segmentation
pipeline based on DINO-X \cite{ren2024dinox} and SAM 2.1
\cite{ravi2024sam2}.

DINO-X provides semantic localization of the task-relevant foreground,
while SAM 2.1 refines its pixel-level boundary. The two predictions are
required to agree before a foreground mask is accepted. This prevents
the pair-construction procedure from depending exclusively on the
confidence estimate of a single segmentation model.

We first query DINO-X using the text prompt ``dog.'' Detections with
confidence below \(0.25\) are discarded. When multiple detections
remain, we select the highest-confidence candidate whose bounding-box
area lies between 5\% and 90\% of the image area.

Let \(b_i\) and \(m_i^{\mathrm{DX}}\) denote the selected DINO-X box and
mask. The box is then supplied to the SAM 2.1 Hiera-Large image
predictor. We additionally use the center of the DINO-X mask as a
positive point prompt. SAM 2.1 returns candidate masks

\begin{equation}
    \left\{
        m_{i,1}^{\mathrm{S2}},
        m_{i,2}^{\mathrm{S2}},
        m_{i,3}^{\mathrm{S2}}
    \right\}.
\end{equation}

For each SAM 2.1 candidate, we compute its agreement with the DINO-X
mask:

\begin{equation}
    G_{i,k}
    =
    \operatorname{IoU}
    \left(
        m_i^{\mathrm{DX}},
        m_{i,k}^{\mathrm{S2}}
    \right).
\end{equation}

The selected candidate is

\begin{equation}
    k^{\star}
    =
    \arg\max_k
    G_{i,k}.
\end{equation}

The final foreground mask is

\begin{equation}
    m_{\mathrm{dog}}
    =
    m_{i,k^{\star}}^{\mathrm{S2}}.
\end{equation}

The mask is retained only when

\begin{equation}
    G_{i,k^{\star}}
    \geq
    0.80.
\end{equation}

This agreement condition filters cases in which semantic localization
and boundary segmentation identify inconsistent regions. We remove
disconnected components whose area is smaller than 1\% of the selected
foreground mask and retain the largest remaining connected component.

The retained mask must additionally satisfy

\begin{equation}
    0.05
    \leq
    \frac{
        \left\|m_{\mathrm{dog}}\right\|_1
    }{
        HW
    }
    \leq
    0.90.
\end{equation}

The background cue mask is

\begin{equation}
    m_q
    =
    \mathbf{1}
    -
    m_{\mathrm{dog}}.
\end{equation}

To construct a matched reference, we preserve the source dog and replace
its environment with a background sampled independently of dog breed.
A donor image is sampled uniformly from one of the other five
environments and from the same dataset split. The donor image must
differ from the source image.

The dog in the donor image is localized and segmented using the same
DINO-X--SAM 2.1 pipeline. Its mask is dilated by five pixels to include
the immediate object boundary. The resulting region is removed using
BrushNet \cite{ju2024brushnet}, with the text condition ``an empty
natural background without dogs or other animals.''

To reduce dependence on one stochastic completion, we generate three
candidate background plates using fixed random seeds. Let
\(B_{j,s}^{\mathrm{plate}}\) denote candidate \(s\). Each candidate is
processed by DINO-X using the prompts ``dog'' and ``animal.'' Its
residual foreground score is

\begin{equation}
    R_{j,s}
    =
    \max
    \left(
        r_{j,s}^{\mathrm{dog}},
        r_{j,s}^{\mathrm{animal}}
    \right),
\end{equation}

where \(r_{j,s}^{\mathrm{dog}}\) and
\(r_{j,s}^{\mathrm{animal}}\) are the maximum detection confidences for
the two prompts.

The donor background plate is selected as

\begin{equation}
    s^{\star}
    =
    \arg\min_s
    R_{j,s}.
\end{equation}

The selected plate is

\begin{equation}
    B_j^{\mathrm{plate}}
    =
    B_{j,s^{\star}}^{\mathrm{plate}}.
\end{equation}

The donor plate is accepted only when

\begin{equation}
    R_{j,s^{\star}}
    <
    0.20.
\end{equation}

This second-pass verification prevents a partially removed donor dog or
a newly generated animal-like region from entering the matched
background.

To avoid copying a hard segmentation boundary, the source foreground
mask is converted into a soft alpha mask. We dilate the binary mask by
three pixels and apply Gaussian feathering with standard deviation two
pixels, producing \(A_i\). The matched reference is

\begin{equation}
    x_i^{\setminus q}
    =
    A_i\odot x_i
    +
    \left(
        \mathbf{1}-A_i
    \right)
    \odot
    B_j^{\mathrm{plate}}.
\end{equation}

The source dog retains its original pixels, position, scale, and pose.
Only the surrounding environment is replaced. Donor environments are
sampled uniformly and independently of source breed. Pair assignments,
foreground masks, and generated donor plates are fixed once and shared
across architectures and random seeds.

An image is excluded when no valid dog detection is produced, DINO-X
and SAM 2.1 fail the agreement requirement, the foreground is severely
truncated by the image boundary, the mask contains multiple large
disconnected components, or all BrushNet candidates fail the residual
animal verification.

\subsection*{D5. ImageNet-9 Background Challenge}

ImageNet-9 groups ImageNet classes into nine broad object categories and
provides controlled variants that separate foreground and background
information. We use the Mixed-Same, Mixed-Rand, and Only-FG variants.

Mixed-Same places a foreground object on a background obtained from the
same superclass. Mixed-Rand places the same foreground object on a
background sampled independently of its superclass. Only-FG preserves
the foreground while neutralizing the natural background.

For the primary analysis, the cue-present image is taken from
Mixed-Same and its matched reference from Mixed-Rand. Images are paired
using their shared foreground identifier so that object pixels,
location, scale, and class remain fixed:

\begin{equation}
    x_i
    =
    x_i^{\mathrm{Mixed\text{-}Same}}.
\end{equation}

\begin{equation}
    x_i^{\setminus q}
    =
    x_i^{\mathrm{Mixed\text{-}Rand}}.
\end{equation}

The foreground support is recovered from the corresponding Only-FG
image. Let \(m_{\mathrm{fg}}\) denote this foreground mask. The
background cue support is

\begin{equation}
    m_q
    =
    \mathbf{1}
    -
    m_{\mathrm{fg}}.
\end{equation}

Mixed-Same and Mixed-Rand images are retained as a pair only when they
share the same foreground identifier and spatial placement. Images are
not paired solely according to their superclass label.

Only-FG is used as a stricter alternative cue-removal reference:

\begin{equation}
    x_i^{\setminus q,\mathrm{fg}}
    =
    x_i^{\mathrm{Only\text{-}FG}}.
\end{equation}

This alternative removes background appearance rather than replacing
it with another natural background. Agreement across the two references
tests whether the measured dynamics depend on one particular donor
background.

\subsection*{D6. Cue-removal Operators}

We use different cue-removal constructions according to the information
available in each benchmark.

\paragraph{Native component removal.}

When the benchmark provides separable foreground, background, and
object layers, the selected component is removed or replaced directly
through the native generation pipeline. This is the primary operator
for Waterbirds and UrbanCars.

\paragraph{Independent component resampling.}

The selected background is replaced with a component sampled
independently of the target label. In the primary Waterbirds and
UrbanCars construction, sampling is restricted to the opposite
background domain so that the original shortcut attribute is absent.
In the alternative construction, donor backgrounds are sampled
uniformly from the complete background pool without conditioning on the
target label.

\paragraph{Foreground-only neutralization.}

The target foreground is preserved and the background is replaced by a
spatially constant neutral field. Let the neutral background be the
per-channel mean of the training images after reversing input
normalization:

\begin{equation}
    B_{\mathrm{neutral}}
    =
    \left(
        \mu_R,
        \mu_G,
        \mu_B
    \right).
\end{equation}

The resulting reference is

\begin{equation}
    x^{\setminus q,\mathrm{neutral}}
    =
    m_{\mathrm{fg}}\odot x
    +
    \left(
        \mathbf{1}-m_{\mathrm{fg}}
    \right)
    \odot B_{\mathrm{neutral}}.
\end{equation}

This operator is used only for sensitivity analysis because its output
is less natural than a native composite with a real background.

\paragraph{Cross-model foreground-preserving replacement.}

For images without native compositing layers, DINO-X and SAM 2.1 are
used jointly to isolate the task-relevant foreground. Candidate masks
are retained only when the two models achieve sufficient region-level
agreement. BrushNet removes the donor foreground, and DINO-X performs a
second verification pass to reject donor plates containing residual or
generated animal regions. The source foreground pixels are copied
directly and are not regenerated. This operator is used for
Spawrious-M2M-Hard.

All cue-removal operators preserve image resolution and use the same
normalization and preprocessing as the corresponding original
classifier input.

\subsection*{D7. Automated Pair Validation}

Candidate references are validated before inference-dynamics
trajectories are computed. Pair validation does not use locality,
withdrawability, lock-in depth, reset-layer selection, or downstream
correction performance.

Let \(m_q\) denote the edited cue support. The mean absolute image change
outside that support is

\begin{equation}
    E_{\mathrm{pres}}
    =
    \frac{
        \left\|
            \left(
                \mathbf{1}-m_q
            \right)
            \odot
            \left(
                x-x^{\setminus q}
            \right)
        \right\|_1
    }{
        3
        \left\|
            \mathbf{1}-m_q
        \right\|_1
        +
        \epsilon
    }.
\end{equation}

For native compositing pairs, we require

\begin{equation}
    E_{\mathrm{pres}}
    \leq
    \frac{1}{255}.
\end{equation}

This tolerance permits only small rounding differences introduced by
image encoding. For segmentation-assisted pairs, alpha feathering may
modify a narrow boundary around the foreground. Preservation is
therefore measured outside a five-pixel foreground boundary band, and
we require

\begin{equation}
    E_{\mathrm{pres}}
    \leq
    \frac{3}{255}.
\end{equation}

For background interventions, we additionally evaluate the semantic and
structural preservation of the task foreground using a frozen DINOv2
ViT-B/14 encoder \cite{oquab2023dinov2}. Features are extracted from
the original and reference images after masking out the corresponding
background regions.

Let \(\phi_{\mathrm{fg}}(\cdot)\) denote the pooled masked-foreground
representation. The foreground similarity is

\begin{equation}
    S_{\mathrm{fg}}
    =
    \frac{
        \phi_{\mathrm{fg}}(x)^{\top}
        \phi_{\mathrm{fg}}
        \left(
            x^{\setminus q}
        \right)
    }{
        \left\|
            \phi_{\mathrm{fg}}(x)
        \right\|_2
        \left\|
            \phi_{\mathrm{fg}}
            \left(
                x^{\setminus q}
            \right)
        \right\|_2
    }.
\end{equation}

We require

\begin{equation}
    S_{\mathrm{fg}}
    \geq
    0.93.
\end{equation}

The threshold is fixed before constructing the inference-dynamics
trajectories. It accommodates small representation changes caused by
alpha feathering and the narrow foreground-boundary band while
excluding pairs in which the task-relevant foreground undergoes a
substantial semantic or structural change. The similarity test is used
together with pixel-level preservation checks and visual auditing,
rather than as a standalone pair-quality measure.

The cue mask must contain a non-negligible number of pixels:

\begin{equation}
    \frac{
        \left\|m_q\right\|_1
    }{
        HW
    }
    \geq
    0.01.
\end{equation}

For background cues, masks covering up to 98\% of the image are
permitted because small foreground objects naturally produce large
background supports. For localized co-occurring-object cues, masks
covering more than 40\% of the image are treated as invalid.

Native dataset metadata is used to verify that the intended component
has changed. For Waterbirds and UrbanCars background pairs, the donor
background must belong to the required alternative domain. For
UrbanCars object pairs, the target object layer must be absent from the
reference. For ImageNet-9, original and reference images must share the
same foreground identifier. For Spawrious-M2M-Hard, the donor
environment must differ from the source environment and the final donor
plate must pass residual-animal verification.

\subsection*{D8. Visual Quality Audit}

Automated criteria are supplemented by a visual audit. For each dataset
and cue-removal operator, we randomly inspect 100 candidate pairs before
running the main layer-wise analysis. A pair is marked invalid when any
of the following is observed:

\begin{itemize}
    \item part of the task-relevant foreground is removed or overwritten;
    \item the selected cue remains clearly visible;
    \item an additional foreground object is introduced;
    \item the foreground boundary contains a conspicuous halo or seam;
    \item the reference contains a large inpainting or compositing artifact;
    \item the semantic target is no longer visually preserved.
\end{itemize}

If more than 5\% of the audited pairs from an operator fail inspection,
the corresponding pair-generation settings are revised and the audit is
repeated. The final construction configuration is then frozen before
model-relative cue-relevance filtering.

The visual audit evaluates the pair-construction procedure. It is not
used to manually select examples according to their layer-wise
trajectories or correction results.

\subsection*{D9. Cue Relevance and Final Pair Selection}

After passing construction-quality checks, a candidate pair is evaluated
using the frozen classifier. Let

\begin{equation}
    c
    =
    \arg\max_k
    z_k(x)
\end{equation}

denote the class predicted from the original image. The input-level cue
effect is

\begin{equation}
    \Delta_q(x)
    =
    z_c(x)
    -
    z_c
    \left(
        x^{\setminus q}
    \right).
\end{equation}

The pair is retained when

\begin{equation}
    \Delta_q(x)
    \geq
    \tau_{\mathrm{rel}},
\end{equation}

where

\begin{equation}
    \tau_{\mathrm{rel}}
    =
    0.5.
\end{equation}

Cue removal is not required to change the predicted label. A cue may
substantially affect the prediction logit while the remaining visual
evidence is still sufficient to preserve the same argmax.

The semantic class label is fixed by the benchmark construction. For
Evidence Reset training, only pairs whose task foreground remains
class-preserving are used. The frozen classifier is not required to
predict the cue-removed reference correctly, because that execution
serves as a matched representation and prediction target rather than as
a replacement ground-truth annotation. Task-label supervision is
retained separately during reset training.

Let \(Q_i^q\) indicate that image--cue pair \((i,q)\) passes all
construction-quality checks. For architecture \(a\), the number of
validated pairs is

\begin{equation}
    N_{\mathrm{valid}}^{(a)}
    =
    \sum_i
    \sum_{q\in\mathcal{Q}_i}
    \mathbf{1}
    \left[
        Q_i^q=1
    \right]
    \mathbf{1}
    \left[
        \Delta_q^{(a)}(x_i)
        \geq
        \tau_{\mathrm{rel}}
    \right],
\end{equation}

where \(\mathcal{Q}_i\) is the set of specified cues associated with
image \(i\).

The final number of validated pairs is architecture-dependent because
cue relevance is measured relative to the frozen classifier. Dataset
construction, donor assignment, masks, and quality-validation outcomes
remain identical across architectures. Only the model-relative
relevance condition may differ.

\subsection*{D10. Pair Accounting}

Waterbirds contributes one candidate background pair per image.
UrbanCars contributes two candidate single-cue pairs per image: one for
the background and one for the co-occurring object. Before
model-relative filtering, the corresponding training-set counts are

\begin{equation}
    N_{\mathrm{cand}}^{\mathrm{Waterbirds}}
    =
    4{,}795
\end{equation}

and

\begin{equation}
    N_{\mathrm{cand}}^{\mathrm{UrbanCars}}
    =
    2
    \times
    8{,}000
    =
    16{,}000.
\end{equation}

Joint UrbanCars references are counted separately and are not included
in the 16,000 single-cue candidate pairs. Spawrious-M2M-Hard and
ImageNet-9 contribute one candidate background pair for every source
image for which a valid matched foreground and background reference can
be constructed.

For every candidate pair, we store the following metadata:

\begin{itemize}
    \item dataset split and source-image identifier;
    \item target class and original model prediction;
    \item cue type and original cue-domain label;
    \item cue-mask area;
    \item donor-image or reference-image identifier;
    \item preservation error \(E_{\mathrm{pres}}\);
    \item foreground similarity \(S_{\mathrm{fg}}\);
    \item input-level cue effect \(\Delta_q(x)\);
    \item automated quality-validation outcome;
    \item visual-audit configuration;
    \item model-relative relevance outcome.
\end{itemize}

For Spawrious-M2M-Hard, we additionally store the DINO-X detection
confidence, DINO-X--SAM 2.1 mask agreement, selected BrushNet seed, and
residual animal-detection confidence. This metadata fixes pair
construction before the layer-wise analysis and permits the same
image--cue references to be reused across all controlled comparisons.
\section*{E. Metrics and Statistical Protocol}

We evaluate the framework from three complementary perspectives:
task-level robustness, process-level inference dynamics, and the
quality of cue withdrawal after intervention. Unless otherwise stated,
metrics are computed separately for each dataset, backbone, cue type,
and random seed before any cross-setting aggregation is performed.

All model predictions are obtained from the unmodified classifier or
from a single Evidence Reset intervention. Cue-removed references and
complete layer-wise scans are used only for analysis and supervision;
they are not required for the reported test-time execution of Evidence
Reset.

\subsection*{E1. Task Performance and Group Robustness}

Let

\begin{equation}
    \mathcal{D}
    =
    \left\{
        \left(
            x_i,
            y_i,
            g_i
        \right)
    \right\}_{i=1}^{N}
\end{equation}

denote an evaluation set, where \(y_i\) is the task label and \(g_i\)
is the corresponding spurious-attribute group. The predicted class is

\begin{equation}
    \widehat{y}_i
    =
    \arg\max_k
    z_{i,k}.
\end{equation}

\paragraph{Average accuracy.}

Average accuracy is computed over all evaluation examples:

\begin{equation}
    \operatorname{Acc}_{\mathrm{avg}}
    =
    \frac{1}{N}
    \sum_{i=1}^{N}
    \mathbf{1}
    \left[
        \widehat{y}_i
        =
        y_i
    \right].
\end{equation}

\paragraph{Group accuracy.}

For group \(g\), let

\begin{equation}
    \mathcal{D}_g
    =
    \left\{
        i
        \ \middle|\
        g_i=g
    \right\}.
\end{equation}

Its accuracy is

\begin{equation}
    \operatorname{Acc}_g
    =
    \frac{1}{
        \left|
            \mathcal{D}_g
        \right|
    }
    \sum_{i\in\mathcal{D}_g}
    \mathbf{1}
    \left[
        \widehat{y}_i
        =
        y_i
    \right].
\end{equation}

Worst-group accuracy is

\begin{equation}
    \operatorname{Acc}_{\mathrm{WG}}
    =
    \min_{g\in\mathcal{G}}
    \operatorname{Acc}_g,
\end{equation}

where \(\mathcal{G}\) is the set of non-empty evaluation groups.

For Waterbirds, groups are defined by bird class and background type:

\begin{equation}
    g_i
    =
    \left(
        y_i,
        a_i^{\mathrm{bg}}
    \right),
\end{equation}

resulting in four groups.

For UrbanCars, groups are defined by target class, background type, and
co-occurring-object type:

\begin{equation}
    g_i
    =
    \left(
        y_i,
        a_i^{\mathrm{bg}},
        a_i^{\mathrm{obj}}
    \right),
\end{equation}

resulting in eight groups.

For Spawrious-M2M-Hard, groups are defined by dog breed and environment:

\begin{equation}
    g_i
    =
    \left(
        y_i,
        a_i^{\mathrm{env}}
    \right).
\end{equation}

ImageNet-9 results are reported separately for each benchmark variant
rather than through a worst-group aggregation.

\paragraph{Cue-aligned accuracy.}

Cue-aligned accuracy measures performance on examples for which the
specified shortcut attributes agree with the target label. Let
\(\mathcal{D}_{\mathrm{align}}\) denote the aligned subset. We compute

\begin{equation}
    \operatorname{Acc}_{\mathrm{align}}
    =
    \frac{1}{
        \left|
            \mathcal{D}_{\mathrm{align}}
        \right|
    }
    \sum_{i\in\mathcal{D}_{\mathrm{align}}}
    \mathbf{1}
    \left[
        \widehat{y}_i
        =
        y_i
    \right].
\end{equation}

For Waterbirds, an example is aligned when the bird class agrees with
the land or water background attribute. For the main UrbanCars
evaluation, an example is aligned when both the background and
co-occurring-object attributes agree with the car class:

\begin{equation}
    \mathcal{D}_{\mathrm{align}}^{\mathrm{UrbanCars}}
    =
    \left\{
        i
        \ \middle|\
        a_i^{\mathrm{bg}}=y_i,
        \ 
        a_i^{\mathrm{obj}}=y_i
    \right\}.
\end{equation}

For cue-specific UrbanCars analysis, aligned accuracy is additionally
reported by conditioning only on the selected cue and marginalizing over
the other shortcut attribute.

Cue-aligned accuracy is reported together with worst-group accuracy to
measure whether shortcut correction improves conflict groups without
unnecessarily degrading examples on which the original classifier is
already correct.

\subsection*{E2. Inference-Dynamics Metrics}

Inference-dynamics metrics are computed only on validated image--cue
pairs that satisfy the cue-relevance criterion. Each pair contributes
one locality trajectory and one withdrawability trajectory over its
valid candidate layers.

\paragraph{Normalized locality.}

To compare the decline in locality across examples with different
initial concentrations, we define locality relative to the first valid
candidate layer:

\begin{equation}
    \widetilde{L}_l(q;x)
    =
    \frac{
        L_l(q;x)
    }{
        L_{l_1}(q;x)
        +
        \epsilon_L
    },
\end{equation}

where \(l_1\) is the earliest valid candidate layer and

\begin{equation}
    \epsilon_L
    =
    10^{-8}.
\end{equation}

This normalization is used only to determine the onset of a substantial
locality decline. Raw \(L_l\) values are retained for trajectory plots
and descriptive statistics.

\paragraph{Locality-decline onset.}

We define the locality-decline onset as the earliest candidate layer at
which normalized locality remains below a fixed threshold for
\(r_{\mathrm{loc}}\) consecutive valid layers:

\begin{equation}
    l_{\mathrm{loc}}
    =
    \min
    \left\{
        l
        \ \middle|\
        \widetilde{L}_j(q;x)
        \leq
        \tau_{\mathrm{loc}},
        \ 
        \forall j
        \in
        \left\{
            l,
            \ldots,
            l+r_{\mathrm{loc}}-1
        \right\}
    \right\}.
\end{equation}

We use

\begin{equation}
    \tau_{\mathrm{loc}}
    =
    0.50
\end{equation}

and

\begin{equation}
    r_{\mathrm{loc}}
    =
    2.
\end{equation}

Thus, locality onset identifies the first persistent stage at which no
more than half of the initial cue-concentrated representation
difference remains within the mapped support.

If no such interval exists, the pair is categorized as not exhibiting a
stable locality decline under this criterion.

\paragraph{Locality-first ordering.}

For pairs with both a valid locality-decline onset and a stable lock-in
onset, the locality-first indicator is

\begin{equation}
    I_{\mathrm{LF}}(x,q)
    =
    \mathbf{1}
    \left[
        l_{\mathrm{loc}}
        <
        l_{\mathrm{lock}}
    \right].
\end{equation}

The locality-first rate is

\begin{equation}
    R_{\mathrm{LF}}
    =
    \frac{
        \sum_{(x,q)\in\mathcal{V}_{\mathrm{both}}}
        I_{\mathrm{LF}}(x,q)
    }{
        \left|
            \mathcal{V}_{\mathrm{both}}
        \right|
    },
\end{equation}

where \(\mathcal{V}_{\mathrm{both}}\) contains validated pairs for which
both onsets are defined.

Pairs without a stable locality decline or stable lock-in are reported
separately and are not assigned an artificial onset.

\paragraph{Stable lock-in rate.}

Let \(\mathcal{V}\) be the full set of validated image--cue pairs and
let \(I_{\mathrm{lock}}(x,q)\) indicate that a stable lock-in onset is
defined. The stable lock-in rate is

\begin{equation}
    R_{\mathrm{lock}}
    =
    \frac{
        1
    }{
        \left|
            \mathcal{V}
        \right|
    }
    \sum_{(x,q)\in\mathcal{V}}
    I_{\mathrm{lock}}(x,q).
\end{equation}

The complementary no-stable-lock-in rate is

\begin{equation}
    R_{\mathrm{no\text{-}lock}}
    =
    1
    -
    R_{\mathrm{lock}}.
\end{equation}

\paragraph{Normalized lock-in depth.}

Let \(k_{\mathrm{lock}}\) be the position of \(l_{\mathrm{lock}}\) in
the ordered candidate-layer set and let \(K\) be the number of candidate
layers for the architecture. The normalized lock-in depth is

\begin{equation}
    d_{\mathrm{lock}}
    =
    \frac{
        k_{\mathrm{lock}}-1
    }{
        K-1
    }.
\end{equation}

Normalized depth lies in \([0,1]\) and permits comparison between
ResNet-50 and ViT-B/16. Because the distribution is generally
non-Gaussian, lock-in depth is summarized using the median and
interquartile range:

\begin{equation}
    \operatorname{IQR}
    =
    \left[
        Q_{0.25},
        Q_{0.75}
    \right].
\end{equation}

\paragraph{Trajectory aggregation.}

For cross-architecture trajectory visualization, each layer-wise
trajectory is linearly interpolated onto the normalized-depth grid

\begin{equation}
    \mathcal{T}
    =
    \left\{
        0,
        0.01,
        0.02,
        \ldots,
        1.00
    \right\}.
\end{equation}

Interpolation is performed separately for locality and withdrawability.
Each image--cue pair is first averaged across random seeds when the same
pair is available in all runs. Mean curves are then computed over
distinct source pairs, preventing multiple seeds of the same image from
being treated as independent observations. Shaded trajectory bands
represent one standard error of the mean.

\subsection*{E3. Cue-withdrawal and Prediction-alignment Metrics}

Task accuracy alone does not determine whether the corrected prediction
has approached the cue-free execution. We therefore report
distribution-level and logit-level withdrawal metrics in addition to
classification accuracy.

\paragraph{Jensen--Shannon distance.}

For two class-probability distributions \(p\) and \(q\), define

\begin{equation}
    m
    =
    \frac{1}{2}
    \left(
        p+q
    \right).
\end{equation}

The Jensen--Shannon distance used in our evaluation is

\begin{equation}
    \operatorname{JS}
    \left(
        p,
        q
    \right)
    =
    \frac{1}{2}
    D_{\mathrm{KL}}
    \left(
        p
        \,\middle\|\,
        m
    \right)
    +
    \frac{1}{2}
    D_{\mathrm{KL}}
    \left(
        q
        \,\middle\|\,
        m
    \right).
\end{equation}

Natural logarithms are used in all KL-divergence calculations.

\paragraph{Raw cue-free residual.}

Let \(p_i^{+}\), \(p_i^{-}\), and \(p_i^{\mathrm{reset}}\) denote the
prediction distributions of the original, cue-removed, and reset
executions. The raw cue-free residual is

\begin{equation}
    R_i^{\mathrm{JS}}
    =
    \operatorname{JS}
    \left(
        p_i^{\mathrm{reset}},
        p_i^{-}
    \right).
\end{equation}

Lower values indicate that the reset execution more closely matches the
complete cue-removed execution.

\paragraph{Normalized cue residual.}

The normalized cue residual is

\begin{equation}
    \operatorname{NCR}_i
    =
    \frac{
        \operatorname{JS}
        \left(
            p_i^{\mathrm{reset}},
            p_i^{-}
        \right)
    }{
        \operatorname{JS}
        \left(
            p_i^{+},
            p_i^{-}
        \right)
        +
        \epsilon_{\mathrm{NCR}}
    },
\end{equation}

where

\begin{equation}
    \epsilon_{\mathrm{NCR}}
    =
    10^{-6}.
\end{equation}

NCR is computed independently for each validated pair and then
macro-averaged:

\begin{equation}
    \operatorname{NCR}
    =
    \frac{1}{
        \left|
            \mathcal{V}
        \right|
    }
    \sum_{i\in\mathcal{V}}
    \operatorname{NCR}_i.
\end{equation}

Values are not clipped. An NCR below one indicates that reset moves the
prediction distribution closer to the cue-free reference than the
original execution. An NCR above one indicates that the reset output is
farther from the cue-free distribution.

The raw JS residual is always reported alongside NCR in the detailed
results so that conclusions do not depend only on the normalized ratio.

\paragraph{Cue-free prediction agreement.}

Cue-free prediction agreement measures whether the reset and cue-free
executions produce the same predicted class:

\begin{equation}
    \operatorname{CFA}
    =
    \frac{1}{
        \left|
            \mathcal{V}
        \right|
    }
    \sum_{i\in\mathcal{V}}
    \mathbf{1}
    \left[
        \arg\max_k
        p_{i,k}^{\mathrm{reset}}
        =
        \arg\max_k
        p_{i,k}^{-}
    \right].
\end{equation}

This metric does not require the cue-free prediction itself to be
correct. It measures agreement with the matched cue-free execution,
whereas task accuracy is measured separately using the ground-truth
label.

\paragraph{Reset logit recovery.}

Let \(c_i\) be the class predicted from the original image. The
normalized reset recovery for that class is

\begin{equation}
    \operatorname{RR}_i
    =
    \frac{
        z_{i,c_i}^{+}
        -
        z_{i,c_i}^{\mathrm{reset}}
    }{
        z_{i,c_i}^{+}
        -
        z_{i,c_i}^{-}
        +
        \epsilon_{\mathrm{RR}}
    },
\end{equation}

where

\begin{equation}
    \epsilon_{\mathrm{RR}}
    =
    10^{-8}.
\end{equation}

Values near one indicate that the learned reset reproduces the
original-class logit change caused by complete cue removal. Values
outside \([0,1]\) are retained.

Unlike layer-wise withdrawability \(W_l\), reset recovery is computed
from the output of the learned reconstruction module rather than from
oracle replacement with the paired cue-free hidden state.

\subsection*{E4. Independent Evaluation of Lock-in Predictive Validity}

We evaluate whether the lock-in onset separates layers with different
correction quality using the learned reset module. The lock-in boundary
is determined only from the matched-state withdrawal trajectory
\(W_l\). The correction metrics below are computed from independently
reconstructed reset states and are not used to define
\(l_{\mathrm{lock}}\).

For every validated pair and every valid candidate layer \(l\), we apply
the trained reset module at that layer, producing

\begin{equation}
    p_{i,l}^{\mathrm{reset}}
\end{equation}

and the corresponding task prediction

\begin{equation}
    \widehat{y}_{i,l}^{\mathrm{reset}}
    =
    \arg\max_k
    p_{i,l,k}^{\mathrm{reset}}.
\end{equation}

\paragraph{Independent best correction layer.}

For each layer, define the correction tuple

\begin{equation}
    \mathcal{C}_{i,l}
    =
    \left(
        \mathbf{1}
        \left[
            \widehat{y}_{i,l}^{\mathrm{reset}}
            \neq
            y_i
        \right],
        \ 
        \operatorname{JS}
        \left(
            p_{i,l}^{\mathrm{reset}},
            p_i^{-}
        \right)
    \right).
\end{equation}

Tuples are ordered lexicographically. Task-correct reset outputs are
therefore preferred to task-incorrect outputs; among layers with the
same task-correctness status, the layer with the smaller cue-free JS
residual is preferred.

The independently defined best correction layer is

\begin{equation}
    l_i^{\mathrm{best}}
    =
    \arg\min_{l\in\mathcal{L}_i}^{\mathrm{lex}}
    \mathcal{C}_{i,l},
\end{equation}

where \(\mathcal{L}_i\) contains the valid candidate layers for example
\(i\). If multiple layers have identical tuples up to numerical
precision, the shallower layer is selected.

This layer is used only for diagnostic analysis and is not used to
train the selector or choose the test-time intervention layer.

\paragraph{Best layer before lock-in rate.}

For examples with stable lock-in, we compute

\begin{equation}
    R_{\mathrm{best\text{-}pre}}
    =
    \frac{
        1
    }{
        \left|
            \mathcal{V}_{\mathrm{lock}}
        \right|
    }
    \sum_{i\in\mathcal{V}_{\mathrm{lock}}}
    \mathbf{1}
    \left[
        l_i^{\mathrm{best}}
        <
        l_{i,\mathrm{lock}}
    \right].
\end{equation}

This statistic tests whether the layer with the best learned-reset
outcome tends to occur before the boundary derived from oracle
withdrawability.

\paragraph{Best pre-lock and post-lock interventions.}

For each stable-lock-in pair, define the valid pre-lock and post-lock
sets as

\begin{equation}
    \mathcal{L}_i^{\mathrm{pre}}
    =
    \left\{
        l\in\mathcal{L}_i
        \ \middle|\
        l<l_{i,\mathrm{lock}}
    \right\}
\end{equation}

and

\begin{equation}
    \mathcal{L}_i^{\mathrm{post}}
    =
    \left\{
        l\in\mathcal{L}_i
        \ \middle|\
        l\geq l_{i,\mathrm{lock}}
    \right\}.
\end{equation}

The best pre-lock and post-lock layers are selected using the same
independent lexicographic correction criterion:

\begin{equation}
    l_i^{\mathrm{pre}}
    =
    \arg\min_{l\in\mathcal{L}_i^{\mathrm{pre}}}^{\mathrm{lex}}
    \mathcal{C}_{i,l},
\end{equation}

\begin{equation}
    l_i^{\mathrm{post}}
    =
    \arg\min_{l\in\mathcal{L}_i^{\mathrm{post}}}^{\mathrm{lex}}
    \mathcal{C}_{i,l}.
\end{equation}

Pairs without at least one valid candidate on both sides of the boundary
are excluded from this particular comparison and reported separately.

Using the corresponding reset predictions, the pre--post worst-group
accuracy gap is

\begin{equation}
    \Delta_{\mathrm{WG}}
    =
    \operatorname{Acc}_{\mathrm{WG}}^{\mathrm{pre}}
    -
    \operatorname{Acc}_{\mathrm{WG}}^{\mathrm{post}}.
\end{equation}

The pre--post normalized cue-residual reduction is

\begin{equation}
    \Delta_{\mathrm{NCR}}
    =
    \operatorname{NCR}^{\mathrm{post}}
    -
    \operatorname{NCR}^{\mathrm{pre}}.
\end{equation}

Larger positive values of both quantities indicate that correction
before lock-in is more effective than correction after lock-in.

Because the pre-lock and post-lock layers are selected using oracle
access to cue-free evaluation targets, these statistics characterize
the predictive meaning of the boundary and do not represent a deployable
layer-selection procedure.

\subsection*{E5. Selector and No-reset Metrics}

The layer selector is evaluated against the dynamics-derived target
label. Let \(k_i^{\star}\) be the target candidate-layer index and
\(\widehat{k}_i\) the predicted candidate-layer index.

\paragraph{Normalized layer error.}

For an example whose target is a reset layer, the layer error is

\begin{equation}
    e_i^{\mathrm{layer}}
    =
    \begin{cases}
        \dfrac{
            \left|\widehat{k}_i-k_i^{\star}\right|
        }{K-1},
        & \widehat{k}_i\neq\varnothing,
        \\[2mm]
        1,
        & \widehat{k}_i=\varnothing.
    \end{cases}
\end{equation}

The reported layer error is

\begin{equation}
    E_{\mathrm{layer}}
    =
    \frac{1}{
        \left|
            \mathcal{V}_{\mathrm{reset}}
        \right|
    }
    \sum_{i\in\mathcal{V}_{\mathrm{reset}}}
    e_i^{\mathrm{layer}},
\end{equation}

where \(\mathcal{V}_{\mathrm{reset}}\) contains examples with a valid
reset-layer target. Lower values are better.

\paragraph{Exact and near-layer accuracy.}

Exact layer accuracy is

\begin{equation}
    A_{\mathrm{exact}}
    =
    \frac{1}{
        \left|
            \mathcal{V}_{\mathrm{reset}}
        \right|
    }
    \sum_{i\in\mathcal{V}_{\mathrm{reset}}}
    \mathbf{1}
    \left[
        \widehat{k}_i
        =
        k_i^{\star}
    \right].
\end{equation}

Within-one-layer accuracy is

\begin{equation}
    A_{\pm1}
    =
    \frac{1}{
        \left|
            \mathcal{V}_{\mathrm{reset}}
        \right|
    }
    \sum_{i\in\mathcal{V}_{\mathrm{reset}}}
    \mathbf{1}
    \left[
        \left|
            \widehat{k}_i-k_i^{\star}
        \right|
        \leq
        1
    \right].
\end{equation}

A no-reset prediction is counted as incorrect for both quantities when
the target is a reset layer.

\paragraph{No-reset precision and recall.}

Let the positive class denote the no-reset target. We define

\begin{equation}
    \operatorname{TP}_{\mathrm{NR}}
    =
    \sum_i
    \mathbf{1}
    \left[
        \widehat{y}_{i,\mathrm{sel}}
        =
        \varnothing
    \right]
    \mathbf{1}
    \left[
        y_{i,\mathrm{sel}}
        =
        \varnothing
    \right],
\end{equation}

\begin{equation}
    \operatorname{FP}_{\mathrm{NR}}
    =
    \sum_i
    \mathbf{1}
    \left[
        \widehat{y}_{i,\mathrm{sel}}
        =
        \varnothing
    \right]
    \mathbf{1}
    \left[
        y_{i,\mathrm{sel}}
        \neq
        \varnothing
    \right],
\end{equation}

and

\begin{equation}
    \operatorname{FN}_{\mathrm{NR}}
    =
    \sum_i
    \mathbf{1}
    \left[
        \widehat{y}_{i,\mathrm{sel}}
        \neq
        \varnothing
    \right]
    \mathbf{1}
    \left[
        y_{i,\mathrm{sel}}
        =
        \varnothing
    \right].
\end{equation}

No-reset precision is

\begin{equation}
    P_{\mathrm{NR}}
    =
    \frac{
        \operatorname{TP}_{\mathrm{NR}}
    }{
        \operatorname{TP}_{\mathrm{NR}}
        +
        \operatorname{FP}_{\mathrm{NR}}
        +
        \epsilon
    }.
\end{equation}

No-reset recall is

\begin{equation}
    R_{\mathrm{NR}}
    =
    \frac{
        \operatorname{TP}_{\mathrm{NR}}
    }{
        \operatorname{TP}_{\mathrm{NR}}
        +
        \operatorname{FN}_{\mathrm{NR}}
        +
        \epsilon
    }.
\end{equation}

The corresponding F1 score is

\begin{equation}
    F1_{\mathrm{NR}}
    =
    \frac{
        2
        P_{\mathrm{NR}}
        R_{\mathrm{NR}}
    }{
        P_{\mathrm{NR}}
        +
        R_{\mathrm{NR}}
        +
        \epsilon
    }.
\end{equation}

We additionally report the no-reset class prevalence:

\begin{equation}
    \pi_{\mathrm{NR}}
    =
    \frac{
        1
    }{
        N
    }
    \sum_i
    \mathbf{1}
    \left[
        y_{i,\mathrm{sel}}
        =
        \varnothing
    \right].
\end{equation}

Reporting prevalence prevents a high precision value from being
interpreted without the corresponding class balance.

\subsection*{E6. Aggregation, Uncertainty, and Model Selection}

\paragraph{Random seeds.}

All trainable methods are evaluated using three independent random
seeds:

\begin{equation}
    \mathcal{S}
    =
    \left\{
        0,
        1,
        2
    \right\}.
\end{equation}

For task-performance and reset-performance metrics, we first compute one
value per seed and then report

\begin{equation}
    \overline{m}
    =
    \frac{1}{
        \left|
            \mathcal{S}
        \right|
    }
    \sum_{s\in\mathcal{S}}
    m_s
\end{equation}

together with the sample standard deviation

\begin{equation}
    \operatorname{Std}(m)
    =
    \sqrt{
        \frac{
            1
        }{
            \left|
                \mathcal{S}
            \right|-1
        }
        \sum_{s\in\mathcal{S}}
        \left(
            m_s-\overline{m}
        \right)^2
    }.
\end{equation}

\paragraph{Bootstrap confidence intervals.}

Sample-level dynamics quantities are accompanied by 95\% percentile
bootstrap confidence intervals. We use

\begin{equation}
    B
    =
    10{,}000
\end{equation}

bootstrap replicates.

Resampling is performed at the source-image level rather than at the
image--cue-pair level. Thus, when an UrbanCars image contributes both a
background pair and a co-occurring-object pair, all pairs originating
from the same source image are resampled together. This cluster
bootstrap preserves the dependence among multiple cue observations from
one image.

For grouped datasets, bootstrap sampling is stratified by target group
so that each replicate preserves the original group counts. Confidence
interval endpoints are the 2.5th and 97.5th percentiles of the bootstrap
distribution.

\paragraph{Paired method comparisons.}

Differences between Evidence Reset and a controlled baseline are
computed on identical test examples and random seeds. Statistical
uncertainty for a difference

\begin{equation}
    \Delta m
    =
    m_{\mathrm{reset}}
    -
    m_{\mathrm{baseline}}
\end{equation}

is estimated using a hierarchical paired bootstrap. Each replicate
first resamples the three random seeds with replacement and then
resamples source images within the selected seed. Both methods use the
same resampled indices.

A difference is considered statistically distinguishable from zero
when its 95\% paired-bootstrap interval excludes zero. We report
confidence intervals and effect sizes rather than relying only on a
binary significance decision.

\paragraph{Macro-aggregation across settings.}

When results are summarized across backbones, the ResNet-50 and
ViT-B/16 results are given equal weight:

\begin{equation}
    m_{\mathrm{backbone\text{-}avg}}
    =
    \frac{1}{2}
    \left(
        m_{\mathrm{ResNet}}
        +
        m_{\mathrm{ViT}}
    \right).
\end{equation}

When an ablation is summarized across datasets and backbones, we first
compute the metric independently for every dataset--backbone cell and
then take an unweighted macro-average:

\begin{equation}
    m_{\mathrm{macro}}
    =
    \frac{1}{
        \left|
            \mathcal{C}
        \right|
    }
    \sum_{c\in\mathcal{C}}
    m_c,
\end{equation}

where \(\mathcal{C}\) is the set of dataset--backbone combinations.
This prevents a dataset with more validated pairs from dominating the
aggregate result.

Cue-specific UrbanCars metrics are computed separately for background
and co-occurring-object pairs before their macro-average is reported.

\paragraph{Validation and test protocol.}

All thresholds, loss weights, candidate-layer configurations,
fixed-layer baselines, early-stopping decisions, and model checkpoints
are selected using the validation split. Test data are used only once
for final evaluation.

The globally best fixed reset layer is selected by maximizing validation
worst-group accuracy. When multiple layers have the same validation
worst-group accuracy up to \(0.1\) percentage points, the layer with
lower validation NCR is selected. Remaining ties are resolved in favor
of the shallower layer.

Oracle-layer results, complete layer sweeps, and independently defined
best-layer analyses use cue-free test references only as diagnostic
upper bounds. They do not determine the trained selector, the reported
Evidence Reset prediction, or any test-time hyperparameter.

\section*{F. Controls for Counterfactual and Receptive-field Confounds}

The inference-dynamics analysis relies on a spatial support, a matched
reference, and a layer-wise replacement operation. We therefore
evaluate whether the observed trajectories are specific to the selected
cue and remain stable under reasonable changes in intervention support
and reference construction.

\begin{figure*}[t]
    \centering
    \includegraphics[
        width=\textwidth
    ]{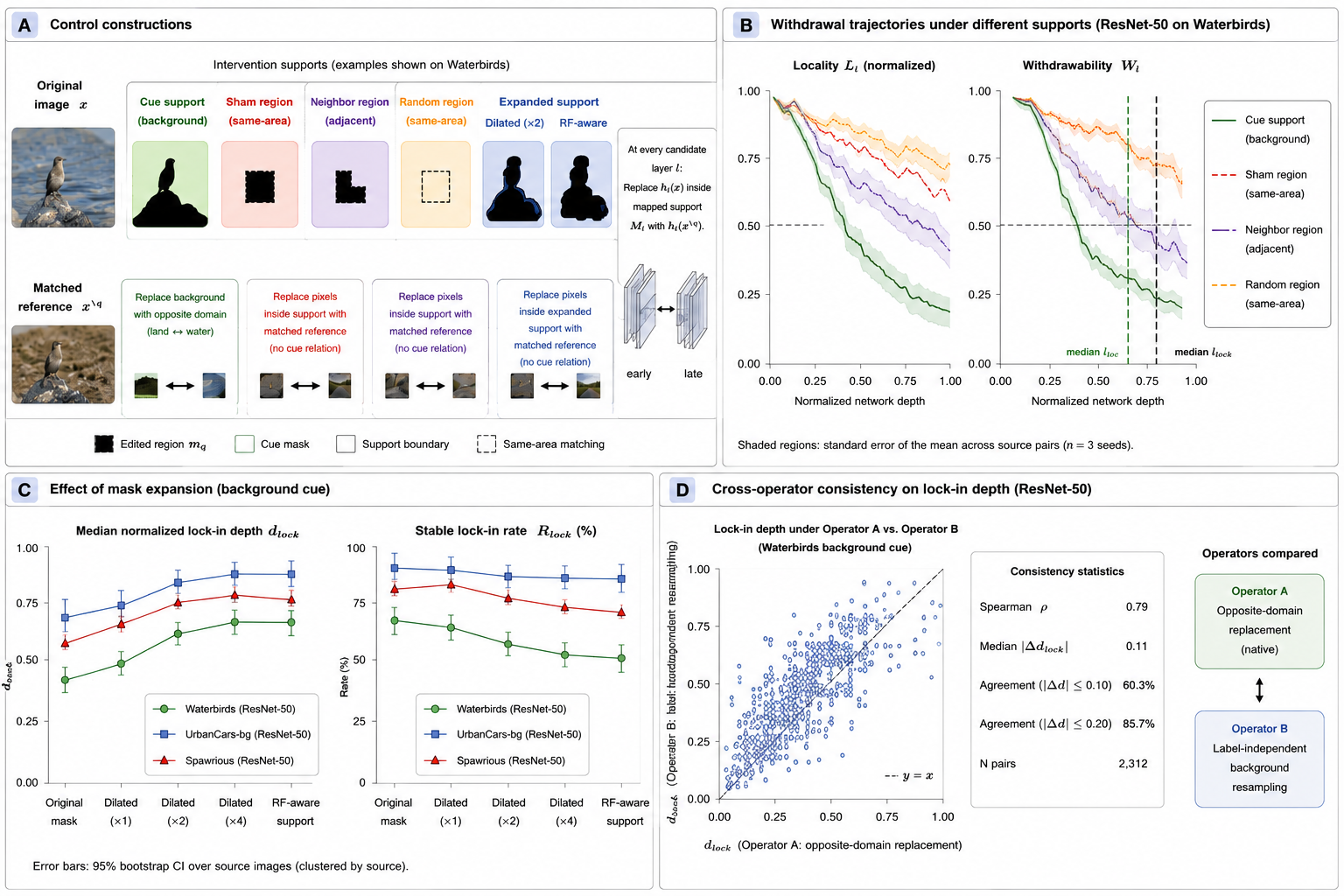}
    \caption{
        Controls for the specificity and robustness of inference
        dynamics.
        \textbf{A:} The original cue support is compared with a
        cue-preserving sham edit, spatially neighboring and random
        control supports, progressively dilated supports, and a
        receptive-field-aware support. Each control uses the same
        layer-wise replacement interface as the primary analysis.
        \textbf{B:} Representative locality and withdrawability
        trajectories under the original cue support and matched control
        constructions. Only the cue-aligned intervention exhibits the
        characteristic transition from early local control to a
        persistent deep-layer loss of withdrawability.
        \textbf{C:} Median normalized lock-in depth and stable lock-in
        rate under progressively expanded background supports.
        Expanding the support delays lock-in, but does not eliminate the
        transition.
        \textbf{D:} Image-level consistency of lock-in depth under
        native opposite-domain replacement and label-independent
        background resampling. Error bars denote 95\% bootstrap
        confidence intervals over source images.
    }
    \label{fig:supp_controls}
\end{figure*}

All controls use the same source images, frozen classifiers, candidate
layers, and relevance-filtered image--cue pairs as the corresponding
primary analysis. A control changes only one component of the
measurement procedure at a time. Figure~\ref{fig:supp_controls}
summarizes the control constructions and their principal outcomes.

\subsection*{F1. Region-specificity and Sham-edit Controls}

We first test whether the measured trajectory can be reproduced by a
generic image edit or by replacing hidden states at a region unrelated
to the selected cue. Figure~\ref{fig:supp_controls}A illustrates the
control constructions.

Let \(M_l^{q}\) denote the mapped support of the selected cue at
candidate layer \(l\). For a control region \(r\), let \(M_l^{r}\)
denote its mapped support. We apply the same state-replacement rule used
for the primary cue intervention:

\begin{equation}
    \widetilde{h}_l^{\,r}
    =
    \left(
        \mathbf{1}-M_l^{r}
    \right)
    \odot
    h_l^{+}
    +
    M_l^{r}
    \odot
    h_l^{-}.
\end{equation}

The resulting prediction is

\begin{equation}
    \mathbf{z}_l^{\,r}
    =
    f_{>l}
    \left(
        \widetilde{h}_l^{\,r}
    \right).
\end{equation}

To make support-specific effects comparable, recovery is normalized by
the input-level effect of removing the true cue:

\begin{equation}
    W_l^{r}(q;x)
    =
    \frac{
        z_c(x)-z_{l,c}^{\,r}
    }{
        z_c(x)-z_c
        \left(
            x^{\setminus q}
        \right)
    }.
\end{equation}

Thus, \(W_l^{r}\) asks how much of the true cue-removal effect can be
reproduced by modifying control region \(r\). The denominator is shared
across the true cue and all spatial controls.

\paragraph{Sham-edit control.}

The sham control uses the same image-editing pipeline and the same cue
support, but preserves the shortcut attribute. For a background cue,
for example, an original background is replaced by another background
from the same cue domain rather than from the opposite domain. Let
\(x^{\mathrm{sham}}\) denote this reference and let

\begin{equation}
    h_l^{\mathrm{sham}}
    =
    f_{\leq l}
    \left(
        x^{\mathrm{sham}}
    \right).
\end{equation}

The sham intervention is

\begin{equation}
    \widetilde{h}_l^{\,\mathrm{sham}}
    =
    \left(
        \mathbf{1}-M_l^{q}
    \right)
    \odot
    h_l^{+}
    +
    M_l^{q}
    \odot
    h_l^{\mathrm{sham}}.
\end{equation}

This control preserves the edited area and image-generation procedure
while removing the intended change in cue identity. It therefore
separates the effect of removing the selected cue from the effect of
processing a different but similarly constructed image.

\paragraph{Neighbor-region control.}

For a localized cue, the neighbor support is sampled from a ring
immediately outside the cue mask. It is constrained not to overlap the
cue and is matched to the number of selected spatial units:

\begin{equation}
    M_l^{\mathrm{nbr}}
    \odot
    M_l^{q}
    =
    \mathbf{0},
\end{equation}

\begin{equation}
    \left\|
        M_l^{\mathrm{nbr}}
    \right\|_1
    =
    \left\|
        M_l^{q}
    \right\|_1.
\end{equation}

When several adjacent regions satisfy these conditions, the region with
the smallest mean distance to the cue boundary is selected. This
control tests whether replacing nearby context is sufficient to
reproduce cue withdrawal.

\paragraph{Random-region control.}

Random supports are sampled from valid non-cue spatial units and are
matched to the cue support in cardinality:

\begin{equation}
    M_l^{\mathrm{rnd}}
    \odot
    M_l^{q}
    =
    \mathbf{0},
\end{equation}

\begin{equation}
    \left\|
        M_l^{\mathrm{rnd}}
    \right\|_1
    =
    \left\|
        M_l^{q}
    \right\|_1.
\end{equation}

We sample five random supports for each image--cue pair and candidate
layer. The reported trajectory is their mean:

\begin{equation}
    W_l^{\mathrm{rnd}}
    =
    \frac{1}{5}
    \sum_{s=1}^{5}
    W_l^{\mathrm{rnd},s}.
\end{equation}

Exact area matching is required. Image--cue pairs for which no valid
non-overlapping control support can be constructed are omitted from the
corresponding region-control comparison rather than being assigned an
undersized control.

Figure~\ref{fig:supp_controls}B compares the resulting trajectories.
The cue-aligned support exhibits the characteristic inference-dynamics
pattern: locality decreases first, followed by a persistent decline in
withdrawability. Sham and random controls do not reproduce this coupled
transition. Their trajectories reflect the effect of replacing generic
states or introducing an image-matched perturbation, but they do not
show the same image-specific change from early cue control to stable
deep-layer lock-in.

Neighbor-region replacement produces an intermediate response. This is
expected because adjacent features may share receptive fields or
participate in local contextual interactions with the cue. However, its
trajectory remains distinguishable from direct intervention on the cue
support. The result indicates that spatial proximity alone does not
fully account for the measured transition.

\subsection*{F2. Mask Expansion and Receptive-field-aware Supports}

The original mask follows the spatial support of the cue in input
space. As network depth increases, convolution, downsampling, and
contextual interaction may cause cue-related effects to involve units
outside this narrowly mapped support. We therefore test whether stable
lock-in disappears when the intervention region is enlarged.

Let

\begin{equation}
    M_l^{(0)}
    =
    M_l^{q}
\end{equation}

denote the original projected support. We construct progressively
expanded supports using binary dilation:

\begin{equation}
    M_l^{(s)}
    =
    \operatorname{Dilate}
    \left(
        M_l^{q},
        s
    \right),
\end{equation}

where

\begin{equation}
    s
    \in
    \left\{
        1,
        2,
        4
    \right\}.
\end{equation}

A dilation step adds one spatial ring at the resolution of the
candidate feature map. The expanded intervention is

\begin{equation}
    \widetilde{h}_l^{\,(s)}
    =
    \left(
        \mathbf{1}-M_l^{(s)}
    \right)
    \odot
    h_l^{+}
    +
    M_l^{(s)}
    \odot
    h_l^{-}.
\end{equation}

Withdrawability under the expanded support is

\begin{equation}
    W_l^{(s)}(q;x)
    =
    \frac{
        z_c(x)-z_{l,c}^{(s)}
    }{
        z_c(x)-z_c
        \left(
            x^{\setminus q}
        \right)
    }.
\end{equation}

The lock-in onset is recomputed independently for every support size
using the same withdrawability threshold and consecutive-layer
criterion as in the primary analysis.

\paragraph{Receptive-field-aware support.}

For ResNet-50, we additionally construct a support based on the
effective spatial footprint of each candidate block. Let \(R_l\) and
\(J_l\) denote the receptive-field size and effective input stride at
layer \(l\). They are propagated recursively as

\begin{equation}
    J_l
    =
    J_{l-1}s_l,
\end{equation}

\begin{equation}
    R_l
    =
    R_{l-1}
    +
    \left(
        k_l-1
    \right)
    J_{l-1},
\end{equation}

where \(k_l\) and \(s_l\) are the kernel size and stride of the
corresponding operation.

The receptive-field-aware dilation radius in feature-map units is

\begin{equation}
    r_l^{\mathrm{RF}}
    =
    \left\lceil
        \frac{
            R_l-R_0
        }{
            2J_l
        }
    \right\rceil.
\end{equation}

The resulting support is

\begin{equation}
    M_l^{\mathrm{RF}}
    =
    \operatorname{Dilate}
    \left(
        M_l^{q},
        r_l^{\mathrm{RF}}
    \right).
\end{equation}

When the theoretical receptive field covers the entire feature map, the
support is capped at the largest connected expansion that leaves at
least 10\% of spatial units outside the intervention. This prevents the
control from becoming equivalent to replacing the complete hidden
state.

For ViT-B/16, global self-attention makes a literal theoretical
receptive field uninformative after the first transformer block. We
therefore use token-neighborhood expansion rather than a full-map
support. The original cue tokens are expanded by one and two spatial
patch rings while the class token remains outside the intervention
mask.

Figure~\ref{fig:supp_controls}C shows the effect of support expansion.
Increasing the mask size shifts the median lock-in depth toward later
layers. This shift is informative: a narrow mask can cease to control
the prediction before a broader region does, indicating that the
effective control interface expands with depth.

Crucially, the shift does not remove the transition. Stable lock-in
remains common under the largest fixed dilation and under the
receptive-field-aware support. The result therefore supports a more
precise interpretation: mask extent affects the measured onset, but
receptive-field growth alone does not account for the persistent
deep-layer loss of local control.

The relative differences among datasets also remain visible after mask
expansion. UrbanCars background cues retain later lock-in onsets than
Waterbirds, while Spawrious-M2M-Hard remains between the two in the
representative ResNet-50 analysis. A single globally expanded support
therefore does not collapse the input- and dataset-dependent variation
into one fixed intervention depth.

\subsection*{F3. Alternative References and Cross-operator Consistency}

We next evaluate whether image-specific lock-in depth depends strongly
on one matched-reference construction. Two independently constructed
background references are compared for the same source image and cue
mask.

\paragraph{Operator A: native opposite-domain replacement.}

Operator A replaces the original background with a background from the
opposite cue domain while preserving the task foreground. This is the
primary reference used for controlled background cues.

\paragraph{Operator B: label-independent background resampling.}

Operator B samples a background independently of the target label from
the available background pool. Source foreground pixels, spatial
placement, and cue mask are identical to Operator A. The resulting pair
must independently pass the same preservation and cue-relevance
criteria.

Let

\begin{equation}
    d_{\mathrm{lock},i}^{A}
\end{equation}

and

\begin{equation}
    d_{\mathrm{lock},i}^{B}
\end{equation}

denote the normalized lock-in depths obtained for source pair \(i\)
under the two operators. Cross-operator analysis uses the intersection
of pairs for which both references pass validation and both trajectories
exhibit stable lock-in:

\begin{equation}
    \mathcal{V}_{AB}
    =
    \mathcal{V}_{A}
    \cap
    \mathcal{V}_{B}.
\end{equation}

We measure rank consistency using Spearman correlation:

\begin{equation}
    \rho_{AB}
    =
    \operatorname{Spearman}
    \left(
        \left\{
            d_{\mathrm{lock},i}^{A}
        \right\},
        \left\{
            d_{\mathrm{lock},i}^{B}
        \right\}
    \right).
\end{equation}

The pair-level onset deviation is

\begin{equation}
    \Delta d_i
    =
    \left|
        d_{\mathrm{lock},i}^{A}
        -
        d_{\mathrm{lock},i}^{B}
    \right|.
\end{equation}

We report its median together with agreement rates at two depth
tolerances:

\begin{equation}
    A_{\delta}
    =
    \frac{1}{
        \left|
            \mathcal{V}_{AB}
        \right|
    }
    \sum_{i\in\mathcal{V}_{AB}}
    \mathbf{1}
    \left[
        \Delta d_i
        \leq
        \delta
    \right],
\end{equation}

where

\begin{equation}
    \delta
    \in
    \left\{
        0.10,
        0.20
    \right\}.
\end{equation}

As shown in Figure~\ref{fig:supp_controls}D, the two operators produce
strongly consistent image-level lock-in rankings. Across 2,312 paired
examples, the Spearman correlation is

\begin{equation}
    \rho_{AB}
    =
    0.79,
\end{equation}

and the median absolute onset difference is

\begin{equation}
    \operatorname{Median}
    \left(
        \Delta d
    \right)
    =
    0.11.
\end{equation}

The lock-in depths differ by at most \(0.10\) normalized network depth
for 60.3\% of pairs and by at most \(0.20\) for 85.7\% of pairs:

\begin{equation}
    A_{0.10}
    =
    60.3\%,
\end{equation}

\begin{equation}
    A_{0.20}
    =
    85.7\%.
\end{equation}

The remaining variation is expected because the two references remove
different background instances and therefore induce different paired
hidden states. The high rank correlation nevertheless indicates that
examples that lock in relatively early or late under one operator tend
to do so under the other.

We additionally repeat the aggregate locality-first and stable-lock-in
analysis under each operator separately. Both operators preserve the
ordering in which locality declines before withdrawability and retain
the advantage of intervention before the detected lock-in boundary.
Thus, the measured onset is quantitatively influenced by the selected
reference, but the principal process-level pattern is not tied to one
specific donor construction.

\subsection*{F4. Interpretation of the Control Results}

The three control families distinguish different sources of variation.

First, sham and region controls show that a generic matched edit or a
replacement at an unrelated spatial support does not reproduce the same
locality--withdrawability transition as intervention on the selected
cue. Neighboring regions can partially affect the trajectory, which is
consistent with local feature interaction, but do not fully substitute
for the cue support.

Second, mask expansion shows that support extent affects the numerical
location of lock-in. Broader supports preserve control for longer and
therefore shift the onset toward later layers. However, stable lock-in
continues to occur under progressively dilated and
receptive-field-aware supports, indicating that the phenomenon is not
eliminated by accounting for a wider local footprint.

Third, cross-operator analysis shows that the image-specific ordering of
lock-in depths remains strongly consistent when the matched reference is
constructed independently. The trajectory is therefore not determined
solely by one background donor or one counterfactual operator.

Together, these controls support the specificity of the measured
inference dynamics within the tested intervention family. They also
clarify that lock-in depth should not be treated as a support-invariant
constant: its exact value depends on how broadly the model is
intervened on, while the transition from local control to persistent
loss of withdrawability remains stable across the tested controls.
\section*{G. Independent Predictive Validity of Evidence Lock-in}

Evidence lock-in is defined from the response to matched-state
withdrawal at the cue-aligned support. We next examine whether this
boundary identifies a meaningful change in the effectiveness of the
learned reset module.

The evaluation deliberately separates boundary construction from
correction assessment. The lock-in onset is computed only from the
withdrawability trajectory \(W_l\). After fixing this onset, we apply
the trained reset module independently at every valid candidate layer
and evaluate its output using task correctness and distance to the
cue-free prediction. Neither \(W_l\), evidence locality, nor the
resetability score is used to rank the learned reset outputs.

All results in this section are computed on held-out test examples.
The reset module and layer selector are trained using the training
split, while thresholds and checkpoints are selected using the
validation split. The selector is not used in the layer-sweep
diagnostic below; instead, the same trained reset module is evaluated
at every candidate layer to characterize the correction landscape
around the lock-in boundary.

\subsection*{G1. Independent Layer-wise Correction Evaluation}

For a validated image--cue pair \(i\), applying the learned reset module
at candidate layer \(l\) produces the prediction distribution

\begin{equation}
    p_{i,l}^{\mathrm{reset}}
    =
    \operatorname{softmax}
    \left(
        \mathbf{z}_{i,l}^{\mathrm{reset}}
    \right).
\end{equation}

Its predicted task class is

\begin{equation}
    \widehat{y}_{i,l}^{\mathrm{reset}}
    =
    \arg\max_k
    p_{i,l,k}^{\mathrm{reset}}.
\end{equation}

Let \(p_i^{-}\) denote the prediction distribution of the matched
cue-removed execution. Correction quality at layer \(l\) is represented
by the ordered tuple

\begin{equation}
    \mathcal{C}_{i,l}
    =
    \left(
        \mathbf{1}
        \left[
            \widehat{y}_{i,l}^{\mathrm{reset}}
            \neq
            y_i
        \right],
        \ 
        \operatorname{JS}
        \left(
            p_{i,l}^{\mathrm{reset}},
            p_i^{-}
        \right)
    \right).
\end{equation}

The tuples are ordered lexicographically. A task-correct reset output is
therefore preferred to a task-incorrect output. Among layers with the
same task-correctness status, the layer whose prediction distribution is
closer to the cue-free execution is preferred. This avoids introducing
a manually weighted combination of task loss and prediction residual.

The independently evaluated best correction layer is

\begin{equation}
    l_i^{\mathrm{best}}
    =
    \arg\min_{l\in\mathcal{L}_i}^{\mathrm{lex}}
    \mathcal{C}_{i,l},
\end{equation}

where \(\mathcal{L}_i\) contains all valid candidate layers for pair
\(i\). When two layers have equal task-correctness status and their
Jensen--Shannon residuals differ by less than \(10^{-6}\), the shallower
layer is selected.

This definition does not use \(W_l\), \(L_l\), \(\rho_l\), or the
dynamics-derived target layer. It is used only as a diagnostic measure
of where the trained reset module achieves its best held-out correction.

For pairs exhibiting stable lock-in, we report the fraction for which
the independently evaluated best correction layer occurs before the
lock-in onset:

\begin{equation}
    R_{\mathrm{best\text{-}pre}}
    =
    \frac{
        1
    }{
        \left|
            \mathcal{V}_{\mathrm{lock}}
        \right|
    }
    \sum_{i\in\mathcal{V}_{\mathrm{lock}}}
    \mathbf{1}
    \left[
        l_i^{\mathrm{best}}
        <
        l_{i,\mathrm{lock}}
    \right].
\end{equation}

Only pairs with at least one valid candidate layer on both sides of the
boundary are included in the direct pre--post comparison. Early-lock-in
pairs without a valid pre-lock layer are analyzed separately.

\begin{table*}[t]
    \centering
    \small
    \setlength{\tabcolsep}{5pt}
    \renewcommand{\arraystretch}{1.10}
    \caption{
        Independent predictive validity of evidence lock-in. The
        lock-in boundary is defined from matched-state withdrawability,
        whereas correction quality is measured using outputs of the
        learned reset module. ``Best pre-lock'' is the percentage of
        examples whose independently evaluated best correction layer
        occurs before lock-in. Rates and accuracies are mean
        \(\pm\) standard deviation over three runs. Larger WG gaps and
        NCR reductions favor intervention before lock-in.
    }
    \label{tab:g_predictive_validity}
    \begin{tabular}{llccccccc}
        \toprule
        \textbf{Dataset}
        &
        \textbf{Backbone}
        &
        \shortstack{\textbf{Best}\\\textbf{pre-lock (\%)}}
        &
        \shortstack{\textbf{Pre-lock}\\\textbf{WG}}
        &
        \shortstack{\textbf{Post-lock}\\\textbf{WG}}
        &
        \shortstack{\textbf{WG}\\\textbf{gap}}
        &
        \shortstack{\textbf{Pre-lock}\\\textbf{NCR}}
        &
        \shortstack{\textbf{Post-lock}\\\textbf{NCR}}
        &
        \shortstack{\textbf{NCR}\\\textbf{reduction}}
        \\
        \midrule

        Waterbirds
        &
        ResNet-50
        &
        \(88.2\pm1.9\)
        &
        \(94.1\pm0.6\)
        &
        \(86.7\pm1.1\)
        &
        \(+7.4\)
        &
        \(0.13\pm0.01\)
        &
        \(0.35\pm0.02\)
        &
        \(0.22\)
        \\

        Waterbirds
        &
        ViT-B/16
        &
        \(91.0\pm1.6\)
        &
        \(95.0\pm0.5\)
        &
        \(86.4\pm1.0\)
        &
        \(+8.6\)
        &
        \(0.11\pm0.01\)
        &
        \(0.36\pm0.02\)
        &
        \(0.25\)
        \\

        UrbanCars
        &
        ResNet-50
        &
        \(83.4\pm2.4\)
        &
        \(77.5\pm0.9\)
        &
        \(71.4\pm1.3\)
        &
        \(+6.1\)
        &
        \(0.18\pm0.02\)
        &
        \(0.36\pm0.03\)
        &
        \(0.18\)
        \\

        UrbanCars
        &
        ViT-B/16
        &
        \(86.7\pm2.1\)
        &
        \(79.4\pm0.8\)
        &
        \(72.4\pm1.2\)
        &
        \(+7.0\)
        &
        \(0.16\pm0.02\)
        &
        \(0.37\pm0.03\)
        &
        \(0.21\)
        \\

        \bottomrule
    \end{tabular}
\end{table*}

Table~\ref{tab:g_predictive_validity} shows that the independently
evaluated best reset layer occurs before lock-in for 83.4--91.0\% of
stable-lock-in examples. The result is consistent across both
convolutional and transformer backbones and is not dominated by one
dataset.

The strongest ordering is observed on Waterbirds with ViT-B/16, where
91.0\% of the best learned-reset layers occur before lock-in.
UrbanCars produces a lower but still substantial pre-lock rate because
its two independently controlled shortcuts create more heterogeneous
correction landscapes. Nevertheless, both UrbanCars backbones retain a
clear majority of best correction layers before the detected boundary.

The pre--post comparison yields the same conclusion under metrics that
are not used to define lock-in. The best pre-lock reset improves
worst-group accuracy by 6.1--8.6 percentage points relative to the best
post-lock reset. It also reduces NCR by 0.18--0.25. Thus, the detected
boundary separates layers at which the same learned reconstruction
mechanism has systematically different downstream effectiveness.

\subsection*{G2. Correction Quality Before and After Lock-in}

To examine how correction changes near the sample-specific boundary, we
align candidate layers by their signed offset from lock-in. Let
\(k_l\) be the index of candidate layer \(l\), and let
\(k_{i,\mathrm{lock}}\) be the lock-in index for pair \(i\). The
boundary-relative position is

\begin{equation}
    \Delta k_{i,l}
    =
    k_l
    -
    k_{i,\mathrm{lock}}.
\end{equation}

Negative offsets correspond to layers before lock-in, zero denotes the
onset layer, and positive offsets correspond to layers after lock-in.
Because ResNet-50 and ViT-B/16 have different numbers of candidate
layers, this analysis uses discrete layer offsets within each
architecture before macro-averaging the resulting metrics.

For each offset bin, we report three independently measured quantities:
the task-correct reset rate, the raw Jensen--Shannon residual to the
cue-free execution, and NCR.

The task-correct reset rate is

\begin{equation}
    A_{\mathrm{reset}}(\Delta k)
    =
    \frac{
        1
    }{
        \left|
            \mathcal{B}_{\Delta k}
        \right|
    }
    \sum_{(i,l)\in\mathcal{B}_{\Delta k}}
    \mathbf{1}
    \left[
        \widehat{y}_{i,l}^{\mathrm{reset}}
        =
        y_i
    \right],
\end{equation}

where

\begin{equation}
    \mathcal{B}_{\Delta k}
    =
    \left\{
        (i,l)
        \ \middle|\
        \Delta k_{i,l}
        =
        \Delta k
    \right\}.
\end{equation}

The boundary-aligned results are summarized in
Table~\ref{tab:g_boundary_aligned}.

\begin{table}[t]
    \centering
    \small
    \setlength{\tabcolsep}{4pt}
    \renewcommand{\arraystretch}{1.08}
    \caption{
        Learned-reset quality aligned to the sample-specific lock-in
        boundary. Results are macro-averaged across datasets and
        backbones. Raw JS and NCR are lower when the reset output is
        closer to the cue-free execution.
    }
    \label{tab:g_boundary_aligned}
    \begin{tabular}{lccc}
        \toprule
        \textbf{Relative layer}
        &
        \shortstack{\textbf{Task-correct}\\\textbf{reset (\%)}}
        &
        \shortstack{\textbf{Raw JS}\\\textbf{residual}}
        &
        \textbf{NCR}
        \\
        \midrule

        \(\Delta k\leq-3\)
        & \(94.8\pm0.5\)
        & \(0.022\pm0.002\)
        & \(0.12\pm0.01\)
        \\

        \(\Delta k=-2\)
        & \(94.5\pm0.5\)
        & \(0.024\pm0.002\)
        & \(0.13\pm0.01\)
        \\

        \(\Delta k=-1\)
        & \(94.1\pm0.6\)
        & \(0.027\pm0.002\)
        & \(0.15\pm0.01\)
        \\

        \(\Delta k=0\)
        & \(91.8\pm0.7\)
        & \(0.036\pm0.003\)
        & \(0.22\pm0.02\)
        \\

        \(\Delta k=+1\)
        & \(89.2\pm0.8\)
        & \(0.045\pm0.003\)
        & \(0.30\pm0.02\)
        \\

        \(\Delta k=+2\)
        & \(87.9\pm0.9\)
        & \(0.052\pm0.004\)
        & \(0.36\pm0.03\)
        \\

        \(\Delta k\geq+3\)
        & \(87.1\pm1.0\)
        & \(0.057\pm0.004\)
        & \(0.41\pm0.03\)
        \\

        \bottomrule
    \end{tabular}
\end{table}

Correction quality changes most sharply around the detected boundary.
At the final pre-lock layer, the learned reset remains task-correct on
94.1\% of examples and achieves an average NCR of 0.15. At the onset
layer, task-correctness decreases to 91.8\% and NCR rises to 0.22. One
layer after lock-in, the corresponding values are 89.2\% and 0.30.

The degradation continues after the onset rather than appearing as an
isolated one-layer fluctuation. At offsets of two or more post-lock
layers, the cue-free residual remains substantially higher than at
pre-lock layers. This persistence is consistent with the
consecutive-layer definition of stable lock-in.

The boundary-aligned pattern is present in all four
dataset--backbone combinations. Absolute performance differs across
datasets, but each setting shows the same ordering:

\begin{equation}
    \operatorname{NCR}_{\mathrm{pre}}
    <
    \operatorname{NCR}_{\mathrm{onset}}
    <
    \operatorname{NCR}_{\mathrm{post}}.
\end{equation}

Similarly, task-correct reset rate follows

\begin{equation}
    A_{\mathrm{reset}}^{\mathrm{pre}}
    >
    A_{\mathrm{reset}}^{\mathrm{onset}}
    >
    A_{\mathrm{reset}}^{\mathrm{post}}.
\end{equation}

These results show that sample-specific alignment exposes a common
correction transition that would be blurred by averaging all examples
at the same absolute network layer. Examples lock in at different
depths, but their learned-reset behavior changes in a consistent manner
around their own onset.

\subsection*{G3. Exceptional Trajectories and Boundary Cases}

Not every validated image--cue pair yields an actionable pre-lock
interval. We retain these cases as separate trajectory categories
rather than forcing them into the standard pre--post comparison.

We use four mutually exclusive categories:

\paragraph{Actionable stable lock-in.}

A stable lock-in onset is detected, and at least one valid candidate
layer exists before the onset. These examples support both
dynamics-derived layer supervision and direct pre--post correction
comparison.

\paragraph{Early lock-in.}

Stable lock-in begins at the first valid candidate layer or before a
valid pre-lock reset point is available. These examples exhibit a
persistent loss of local control but provide no actionable interval for
the current reset mechanism.

\paragraph{Persistently withdrawable.}

Withdrawability remains above the lock-in threshold throughout the
candidate-layer sequence. No stable onset is assigned.

\paragraph{Irregular non-persistent trajectory.}

Withdrawability crosses the threshold at one or more isolated layers but
does not remain below it for the required consecutive interval. These
examples are not labeled as stable lock-in.

The distribution and correction behavior of these categories are
reported in Table~\ref{tab:g_exceptional_cases}.

\begin{table}[t]
    \centering
    \small
    \setlength{\tabcolsep}{4pt}
    \renewcommand{\arraystretch}{1.10}
    \caption{
        Distribution and characteristic correction behavior of
        trajectory categories, macro-averaged across primary datasets
        and backbones.
    }
    \label{tab:g_exceptional_cases}
    \begin{tabularx}{\columnwidth}{
        @{}
        >{\raggedright\arraybackslash}p{0.31\columnwidth}
        >{\centering\arraybackslash}p{0.15\columnwidth}
        >{\raggedright\arraybackslash}X
        @{}
    }
        \toprule
        \textbf{Trajectory category}
        &
        \textbf{Share}
        &
        \textbf{Correction behavior}
        \\
        \midrule

        Actionable stable lock-in
        &
        83.6\%
        &
        Best reset usually occurs before onset; pre-lock intervention
        has substantially lower NCR.
        \\

        Early lock-in
        &
        6.3\%
        &
        The earliest candidate is usually the strongest available
        intervention, but residual cue dependence remains high.
        \\

        Persistently withdrawable
        &
        6.8\%
        &
        Correction quality changes gradually with depth and shows no
        stable pre--post discontinuity.
        \\

        Irregular non-persistent
        &
        3.3\%
        &
        Local fluctuations occur, but no persistent correction boundary
        is observed.
        \\

        \bottomrule
    \end{tabularx}
\end{table}

Actionable stable lock-in accounts for 83.6\% of validated pairs.
Together with early-lock-in cases, 89.9\% of pairs exhibit a stable
lock-in onset, consistent with the high stable-lock-in rates observed
across individual dataset--backbone settings.

Early-lock-in examples are more frequent for small localized cues and
for examples on which the original classifier exhibits a large
input-level cue effect. Their best available reset typically occurs at
the earliest candidate layer, but correction remains less complete than
for examples with a non-empty pre-lock interval. Assigning these cases
to no-reset prevents the selector from treating a late layer as a valid
substitute for a missing early intervention point.

Persistently withdrawable examples behave differently. Their learned
reset quality varies gradually across depth, and the best shallow and
deep interventions differ by only 0.8 percentage points in
worst-group accuracy and 0.03 in NCR. The absence of a stable boundary
is therefore accompanied by the absence of the marked post-lock
correction degradation seen in standard lock-in cases.

Irregular non-persistent trajectories account for 3.3\% of validated
pairs. Their isolated threshold crossings are often associated with
residual blocks that temporarily redirect the cue effect or with
transformer blocks that redistribute attention between neighboring
tokens. Requiring persistence across consecutive layers prevents these
local fluctuations from being interpreted as stable lock-in.

Overall, the independent layer sweep establishes three properties.
First, the best learned-reset layer lies before lock-in for the large
majority of examples. Second, task correctness and cue-free prediction
alignment degrade around the sample-specific onset and remain worse
after it. Third, examples without a stable onset do not exhibit the same
boundary-aligned degradation. The lock-in boundary therefore identifies
when correction restricted to the cue-aligned support becomes
systematically less effective, while preserving distinct treatment of
examples for which no such transition is observed.

\section*{H. Full and Disaggregated Results}

This section reports the complete results underlying the aggregate
comparisons. We separate datasets, backbones, cue types, intervention
depths, and random seeds so that the contribution of input-dependent
layer selection can be assessed in every experimental setting.

All reset-based methods use the same frozen classifier, cue masks,
cue-removed training pairs, reset-module capacity, optimization
schedule, and candidate-layer set. They differ only in the intervention
strategy. Best Fixed Reset selects one global layer using the validation
set, Evidence Reset predicts an image--cue-specific layer, and
Oracle-layer Reset uses the dynamics-derived target layer.

\subsection*{H1. Per-dataset and Per-backbone Main Results}

Tables~\ref{tab:h_waterbirds_backbone} and
\ref{tab:h_urbancars_backbone} report the main results separately for
ResNet-50 and ViT-B/16. The arithmetic mean of the two
backbone-specific means recovers the corresponding aggregate results.

\begin{table*}[t]
    \centering
    \scriptsize
    \setlength{\tabcolsep}{3.2pt}
    \renewcommand{\arraystretch}{1.10}
    \caption{
        Waterbirds results separated by backbone. Avg., WG, NCR, and
        Aligned denote average accuracy, worst-group accuracy,
        normalized cue residual, and cue-aligned accuracy,
        respectively. Results are mean \(\pm\) standard deviation over
        three runs. Lower NCR is better.
    }
    \label{tab:h_waterbirds_backbone}
    \begin{tabular}{lcccccccc}
        \toprule
        &
        \multicolumn{4}{c}{\textbf{ResNet-50}}
        &
        \multicolumn{4}{c}{\textbf{ViT-B/16}}
        \\
        \cmidrule(lr){2-5}
        \cmidrule(lr){6-9}
        \textbf{Method}
        &
        \textbf{Avg.}
        &
        \textbf{WG}
        &
        \textbf{NCR}
        &
        \textbf{Aligned}
        &
        \textbf{Avg.}
        &
        \textbf{WG}
        &
        \textbf{NCR}
        &
        \textbf{Aligned}
        \\
        \midrule

        ERM
        &
        \(96.7\pm0.3\)
        &
        \(74.6\pm1.7\)
        &
        \(0.90\pm0.02\)
        &
        \(98.0\pm0.2\)
        &
        \(96.9\pm0.2\)
        &
        \(77.0\pm1.3\)
        &
        \(0.88\pm0.02\)
        &
        \(98.2\pm0.2\)
        \\

        GroupDRO
        &
        \(94.8\pm0.4\)
        &
        \(90.4\pm0.9\)
        &
        \(0.66\pm0.03\)
        &
        \(95.6\pm0.4\)
        &
        \(95.2\pm0.3\)
        &
        \(91.6\pm0.7\)
        &
        \(0.62\pm0.02\)
        &
        \(96.0\pm0.3\)
        \\

        JTT
        &
        \(95.3\pm0.3\)
        &
        \(87.5\pm1.1\)
        &
        \(0.70\pm0.03\)
        &
        \(96.2\pm0.3\)
        &
        \(95.7\pm0.3\)
        &
        \(88.9\pm0.9\)
        &
        \(0.66\pm0.03\)
        &
        \(96.6\pm0.3\)
        \\

        DFR
        &
        \(96.0\pm0.2\)
        &
        \(89.2\pm0.8\)
        &
        \(0.63\pm0.02\)
        &
        \(96.9\pm0.2\)
        &
        \(96.2\pm0.2\)
        &
        \(90.4\pm0.6\)
        &
        \(0.59\pm0.02\)
        &
        \(97.1\pm0.2\)
        \\

        \midrule

        Input Removal
        &
        \(93.8\pm0.5\)
        &
        \(84.6\pm1.2\)
        &
        \(0.38\pm0.03\)
        &
        \(92.7\pm0.6\)
        &
        \(94.4\pm0.4\)
        &
        \(86.8\pm1.0\)
        &
        \(0.34\pm0.03\)
        &
        \(93.3\pm0.5\)
        \\

        Best Fixed Reset
        &
        \(95.9\pm0.3\)
        &
        \(90.6\pm0.8\)
        &
        \(0.29\pm0.02\)
        &
        \(96.8\pm0.3\)
        &
        \(96.1\pm0.3\)
        &
        \(92.2\pm0.6\)
        &
        \(0.25\pm0.01\)
        &
        \(97.0\pm0.3\)
        \\

        Evidence Reset
        &
        \(96.2\pm0.2\)
        &
        \(\mathbf{93.1\pm0.7}\)
        &
        \(\mathbf{0.15\pm0.01}\)
        &
        \(\mathbf{97.1\pm0.2}\)
        &
        \(96.4\pm0.2\)
        &
        \(\mathbf{94.5\pm0.5}\)
        &
        \(\mathbf{0.13\pm0.01}\)
        &
        \(\mathbf{97.3\pm0.2}\)
        \\

        Oracle-layer Reset
        &
        \(\mathbf{96.4\pm0.2}\)
        &
        \(94.1\pm0.6\)
        &
        \(0.10\pm0.01\)
        &
        \(97.2\pm0.2\)
        &
        \(\mathbf{96.6\pm0.2}\)
        &
        \(95.5\pm0.4\)
        &
        \(0.08\pm0.01\)
        &
        \(97.4\pm0.2\)
        \\

        \bottomrule
    \end{tabular}
\end{table*}

\begin{table*}[t]
    \centering
    \scriptsize
    \setlength{\tabcolsep}{3.2pt}
    \renewcommand{\arraystretch}{1.10}
    \caption{
        UrbanCars results separated by backbone. Results are mean
        \(\pm\) standard deviation over three runs. Rows below the
        midrule use controlled cue-aware interventions. Lower NCR is
        better.
    }
    \label{tab:h_urbancars_backbone}
    \begin{tabular}{lcccccccc}
        \toprule
        &
        \multicolumn{4}{c}{\textbf{ResNet-50}}
        &
        \multicolumn{4}{c}{\textbf{ViT-B/16}}
        \\
        \cmidrule(lr){2-5}
        \cmidrule(lr){6-9}
        \textbf{Method}
        &
        \textbf{Avg.}
        &
        \textbf{WG}
        &
        \textbf{NCR}
        &
        \textbf{Aligned}
        &
        \textbf{Avg.}
        &
        \textbf{WG}
        &
        \textbf{NCR}
        &
        \textbf{Aligned}
        \\
        \midrule

        ERM
        &
        \(94.5\pm0.4\)
        &
        \(45.8\pm2.0\)
        &
        \(0.94\pm0.02\)
        &
        \(98.2\pm0.2\)
        &
        \(94.9\pm0.3\)
        &
        \(49.6\pm1.6\)
        &
        \(0.90\pm0.02\)
        &
        \(98.4\pm0.2\)
        \\

        GroupDRO
        &
        \(92.6\pm0.5\)
        &
        \(67.3\pm1.4\)
        &
        \(0.72\pm0.03\)
        &
        \(94.7\pm0.5\)
        &
        \(93.0\pm0.4\)
        &
        \(70.5\pm1.0\)
        &
        \(0.68\pm0.03\)
        &
        \(95.1\pm0.4\)
        \\

        JTT
        &
        \(93.3\pm0.4\)
        &
        \(63.7\pm1.5\)
        &
        \(0.76\pm0.03\)
        &
        \(95.4\pm0.4\)
        &
        \(93.7\pm0.4\)
        &
        \(66.3\pm1.1\)
        &
        \(0.72\pm0.03\)
        &
        \(95.8\pm0.4\)
        \\

        DFR
        &
        \(93.9\pm0.3\)
        &
        \(64.9\pm1.2\)
        &
        \(0.70\pm0.02\)
        &
        \(96.2\pm0.3\)
        &
        \(94.3\pm0.3\)
        &
        \(67.5\pm0.8\)
        &
        \(0.66\pm0.02\)
        &
        \(96.6\pm0.3\)
        \\

        LLE
        &
        \(94.0\pm0.3\)
        &
        \(70.6\pm1.1\)
        &
        \(0.62\pm0.02\)
        &
        \(96.0\pm0.3\)
        &
        \(94.4\pm0.3\)
        &
        \(73.6\pm0.7\)
        &
        \(0.58\pm0.02\)
        &
        \(96.4\pm0.3\)
        \\

        \midrule

        Input Removal
        &
        \(91.6\pm0.6\)
        &
        \(62.1\pm1.6\)
        &
        \(0.43\pm0.04\)
        &
        \(90.3\pm0.7\)
        &
        \(92.2\pm0.5\)
        &
        \(65.5\pm1.2\)
        &
        \(0.39\pm0.03\)
        &
        \(91.1\pm0.6\)
        \\

        Best Fixed Reset
        &
        \(94.1\pm0.3\)
        &
        \(72.5\pm1.1\)
        &
        \(0.32\pm0.02\)
        &
        \(95.9\pm0.3\)
        &
        \(94.5\pm0.3\)
        &
        \(75.3\pm0.8\)
        &
        \(0.28\pm0.01\)
        &
        \(96.3\pm0.3\)
        \\

        Evidence Reset
        &
        \(94.4\pm0.3\)
        &
        \(\mathbf{76.0\pm0.9}\)
        &
        \(\mathbf{0.18\pm0.02}\)
        &
        \(\mathbf{96.3\pm0.3}\)
        &
        \(94.8\pm0.2\)
        &
        \(\mathbf{78.6\pm0.7}\)
        &
        \(\mathbf{0.16\pm0.02}\)
        &
        \(\mathbf{96.7\pm0.3}\)
        \\

        Oracle-layer Reset
        &
        \(\mathbf{94.6\pm0.2}\)
        &
        \(77.8\pm0.8\)
        &
        \(0.13\pm0.01\)
        &
        \(96.4\pm0.2\)
        &
        \(\mathbf{95.0\pm0.2}\)
        &
        \(80.4\pm0.6\)
        &
        \(0.11\pm0.01\)
        &
        \(96.8\pm0.2\)
        \\

        \bottomrule
    \end{tabular}
\end{table*}

Evidence Reset improves over Best Fixed Reset in every
dataset--backbone cell. On Waterbirds, the gains in worst-group
accuracy are 2.5 points for ResNet-50 and 2.3 points for ViT-B/16. On
UrbanCars, the corresponding gains are 3.5 and 3.3 points. NCR decreases
by 0.12--0.14 in all four settings.

The improvements do not come from sacrificing cue-aligned examples.
Aligned accuracy is preserved or slightly improved relative to Best
Fixed Reset, while average accuracy remains within 0.3 percentage
points of the unmodified classifier. In contrast, direct input removal
substantially reduces cue residual but also damages average and aligned
accuracy because it removes the cue before the model has extracted
task-relevant contextual information.

The gap to Oracle-layer Reset is smaller on Waterbirds than on
UrbanCars. Evidence Reset is 1.0 worst-group point below the oracle on
both Waterbirds backbones and 1.8 points below it on both UrbanCars
backbones. The remaining NCR gap is 0.05 in every setting, indicating
that layer prediction remains the main source of headroom.

\subsection*{H2. Cue-specific Results on UrbanCars}

UrbanCars contains two independently controlled shortcuts. We therefore
evaluate background and co-occurring-object interventions separately.
For each cue-specific evaluation, the selected cue is removed while the
other shortcut component remains unchanged.

\begin{table*}[t]
    \centering
    \scriptsize
    \setlength{\tabcolsep}{4pt}
    \renewcommand{\arraystretch}{1.10}
    \caption{
        UrbanCars results separated by backbone and cue type. BG and
        Obj. denote background and co-occurring-object cues. Each WG
        value is computed using the groups induced by the selected cue
        while marginalizing over the other shortcut. Lower NCR is
        better.
    }
    \label{tab:h_urbancars_cue}
    \begin{tabular}{lcccccccc}
        \toprule
        &
        \multicolumn{4}{c}{\textbf{ResNet-50}}
        &
        \multicolumn{4}{c}{\textbf{ViT-B/16}}
        \\
        \cmidrule(lr){2-5}
        \cmidrule(lr){6-9}
        \textbf{Method}
        &
        \shortstack{\textbf{BG}\\\textbf{WG}}
        &
        \shortstack{\textbf{BG}\\\textbf{NCR}}
        &
        \shortstack{\textbf{Obj.}\\\textbf{WG}}
        &
        \shortstack{\textbf{Obj.}\\\textbf{NCR}}
        &
        \shortstack{\textbf{BG}\\\textbf{WG}}
        &
        \shortstack{\textbf{BG}\\\textbf{NCR}}
        &
        \shortstack{\textbf{Obj.}\\\textbf{WG}}
        &
        \shortstack{\textbf{Obj.}\\\textbf{NCR}}
        \\
        \midrule

        Input Removal
        &
        \(64.8\pm1.4\)
        &
        \(0.40\pm0.03\)
        &
        \(59.4\pm1.8\)
        &
        \(0.46\pm0.04\)
        &
        \(68.1\pm1.1\)
        &
        \(0.36\pm0.03\)
        &
        \(62.9\pm1.4\)
        &
        \(0.42\pm0.03\)
        \\

        Best Fixed Reset
        &
        \(73.6\pm1.0\)
        &
        \(0.29\pm0.02\)
        &
        \(71.4\pm1.2\)
        &
        \(0.35\pm0.02\)
        &
        \(76.2\pm0.8\)
        &
        \(0.25\pm0.01\)
        &
        \(74.4\pm0.9\)
        &
        \(0.31\pm0.02\)
        \\

        Evidence Reset
        &
        \(\mathbf{77.1\pm0.8}\)
        &
        \(\mathbf{0.16\pm0.02}\)
        &
        \(\mathbf{74.9\pm1.0}\)
        &
        \(\mathbf{0.20\pm0.02}\)
        &
        \(\mathbf{79.7\pm0.7}\)
        &
        \(\mathbf{0.14\pm0.01}\)
        &
        \(\mathbf{77.5\pm0.8}\)
        &
        \(\mathbf{0.18\pm0.02}\)
        \\

        Oracle-layer Reset
        &
        \(79.0\pm0.7\)
        &
        \(0.11\pm0.01\)
        &
        \(76.6\pm0.9\)
        &
        \(0.15\pm0.01\)
        &
        \(81.6\pm0.6\)
        &
        \(0.09\pm0.01\)
        &
        \(79.2\pm0.7\)
        &
        \(0.13\pm0.01\)
        \\

        \bottomrule
    \end{tabular}
\end{table*}

Background cues are corrected more completely than localized
co-occurring-object cues. Across both backbones, background NCR is
0.02--0.04 lower than object NCR. Localized objects occupy fewer
representation units and interact more strongly with nearby car and
scene features, making their cue-free states harder to reconstruct.

The benefit of dynamic layer selection remains consistent for both cue
types. Relative to Best Fixed Reset, Evidence Reset improves
background-cue worst-group accuracy by 3.5 points on both backbones.
For co-occurring-object cues, the gains are 3.5 points with ResNet-50
and 3.1 points with ViT-B/16. NCR also decreases for every cue and
backbone combination.

\subsection*{H3. Complete Fixed-depth Sweep}

A single fixed intervention layer is selected on the validation set for
the Best Fixed Reset baseline. To show the complete depth dependence,
we additionally evaluate fixed resets at representative normalized
depths. For each target depth, the nearest valid candidate block is used
for the corresponding architecture.

Each table entry reports worst-group accuracy followed by NCR.

\begin{table*}[t]
    \centering
    \small
    \setlength{\tabcolsep}{6pt}
    \renewcommand{\arraystretch}{1.10}
    \caption{
        Fixed-depth reset sweep. Each cell reports
        \(\mathrm{WG}/\mathrm{NCR}\). Common normalized depths use the
        nearest valid candidate layer. Evidence Reset selects a
        different layer for each image--cue pair.
    }
    \label{tab:h_fixed_sweep}
    \begin{tabular}{lcccc}
        \toprule
        \textbf{Intervention strategy}
        &
        \shortstack{\textbf{Waterbirds}\\\textbf{ResNet-50}}
        &
        \shortstack{\textbf{Waterbirds}\\\textbf{ViT-B/16}}
        &
        \shortstack{\textbf{UrbanCars}\\\textbf{ResNet-50}}
        &
        \shortstack{\textbf{UrbanCars}\\\textbf{ViT-B/16}}
        \\
        \midrule

        Fixed depth \(0.10\)
        &
        \(87.6/0.36\)
        &
        \(88.5/0.33\)
        &
        \(68.4/0.41\)
        &
        \(70.1/0.38\)
        \\

        Fixed depth \(0.25\)
        &
        \(89.8/0.32\)
        &
        \(91.1/0.28\)
        &
        \(71.2/0.35\)
        &
        \(73.8/0.31\)
        \\

        Fixed depth \(0.50\)
        &
        \(90.3/0.30\)
        &
        \(91.8/0.26\)
        &
        \(72.1/0.33\)
        &
        \(74.9/0.29\)
        \\

        Fixed depth \(0.67\)
        &
        \(89.1/0.34\)
        &
        \(90.7/0.30\)
        &
        \(70.8/0.37\)
        &
        \(73.9/0.33\)
        \\

        Fixed depth \(0.83\)
        &
        \(85.7/0.46\)
        &
        \(87.8/0.40\)
        &
        \(66.5/0.49\)
        &
        \(69.4/0.45\)
        \\

        Fixed depth \(0.92\)
        &
        \(81.9/0.60\)
        &
        \(83.6/0.56\)
        &
        \(61.7/0.64\)
        &
        \(64.8/0.59\)
        \\

        \midrule

        Best Fixed Reset
        &
        \(90.6/0.29\)
        &
        \(92.2/0.25\)
        &
        \(72.5/0.32\)
        &
        \(75.3/0.28\)
        \\

        Evidence Reset
        &
        \(\mathbf{93.1/0.15}\)
        &
        \(\mathbf{94.5/0.13}\)
        &
        \(\mathbf{76.0/0.18}\)
        &
        \(\mathbf{78.6/0.16}\)
        \\

        Oracle-layer Reset
        &
        \(94.1/0.10\)
        &
        \(95.5/0.08\)
        &
        \(77.8/0.13\)
        &
        \(80.4/0.11\)
        \\

        \bottomrule
    \end{tabular}
\end{table*}

The fixed-layer sweep has a broad mid-depth optimum, but the best
absolute layer differs across datasets and architectures. Very early
reset leaves insufficient contextual information for reconstruction,
whereas deep reset occurs after a large fraction of examples have
already lost control from the cue-aligned support.

No common fixed depth matches the input-dependent strategy in all four
settings. Evidence Reset improves worst-group accuracy over every
representative fixed depth and reduces NCR by a substantial margin.
The advantage is particularly pronounced at deep layers, where the
learned module still changes the final prediction but cannot fully
reproduce the cue-free execution.

The validation-selected Best Fixed Reset performs better than the
nearest common depth because its precise candidate layer differs across
settings. It nevertheless remains below Evidence Reset, showing that
the result is not explained only by choosing a better global stage.

\subsection*{H4. Seed-wise Consistency}

Table~\ref{tab:h_seedwise} reports the two principal controlled methods
for every random seed. The gain column is computed as Evidence Reset
minus Best Fixed Reset for worst-group accuracy. NCR reduction is Best
Fixed NCR minus Evidence Reset NCR.

\begin{table*}[t]
    \centering
    \small
    \setlength{\tabcolsep}{5pt}
    \renewcommand{\arraystretch}{1.08}
    \caption{
        Seed-wise comparison between Best Fixed Reset and Evidence
        Reset. Positive WG gains and NCR reductions favor Evidence
        Reset.
    }
    \label{tab:h_seedwise}
    \begin{tabular}{llcccccc}
        \toprule
        \textbf{Dataset}
        &
        \textbf{Backbone / seed}
        &
        \shortstack{\textbf{Fixed}\\\textbf{WG}}
        &
        \shortstack{\textbf{Fixed}\\\textbf{NCR}}
        &
        \shortstack{\textbf{Evidence}\\\textbf{WG}}
        &
        \shortstack{\textbf{Evidence}\\\textbf{NCR}}
        &
        \shortstack{\textbf{WG}\\\textbf{gain}}
        &
        \shortstack{\textbf{NCR}\\\textbf{reduction}}
        \\
        \midrule

        Waterbirds
        & ResNet-50 / 0
        & 89.8 & 0.31 & 92.4 & 0.16 & +2.6 & 0.15
        \\

        Waterbirds
        & ResNet-50 / 1
        & 90.7 & 0.29 & 93.2 & 0.15 & +2.5 & 0.14
        \\

        Waterbirds
        & ResNet-50 / 2
        & 91.3 & 0.27 & 93.7 & 0.14 & +2.4 & 0.13
        \\

        \cmidrule(lr){2-8}

        Waterbirds
        & ViT-B/16 / 0
        & 91.6 & 0.26 & 94.0 & 0.14 & +2.4 & 0.12
        \\

        Waterbirds
        & ViT-B/16 / 1
        & 92.2 & 0.25 & 94.5 & 0.13 & +2.3 & 0.12
        \\

        Waterbirds
        & ViT-B/16 / 2
        & 92.8 & 0.24 & 95.0 & 0.12 & +2.2 & 0.12
        \\

        \midrule

        UrbanCars
        & ResNet-50 / 0
        & 71.4 & 0.34 & 75.1 & 0.20 & +3.7 & 0.14
        \\

        UrbanCars
        & ResNet-50 / 1
        & 72.6 & 0.32 & 76.0 & 0.18 & +3.4 & 0.14
        \\

        UrbanCars
        & ResNet-50 / 2
        & 73.5 & 0.30 & 76.9 & 0.16 & +3.4 & 0.14
        \\

        \cmidrule(lr){2-8}

        UrbanCars
        & ViT-B/16 / 0
        & 74.5 & 0.29 & 77.9 & 0.18 & +3.4 & 0.11
        \\

        UrbanCars
        & ViT-B/16 / 1
        & 75.3 & 0.28 & 78.6 & 0.16 & +3.3 & 0.12
        \\

        UrbanCars
        & ViT-B/16 / 2
        & 76.1 & 0.27 & 79.3 & 0.14 & +3.2 & 0.13
        \\

        \bottomrule
    \end{tabular}
\end{table*}

Evidence Reset outperforms Best Fixed Reset for all 12
dataset--backbone--seed combinations. Worst-group gains range from 2.2
to 3.7 percentage points, and NCR reductions range from 0.11 to 0.15.
The ordering is therefore not caused by one favorable initialization or
one architecture.

The seed-wise variation is smaller than the improvement from dynamic
selection. For example, the standard deviation of Evidence Reset
worst-group accuracy is 0.5--0.9 points across the four experimental
cells, whereas its gain over Best Fixed Reset is 2.3--3.5 points at the
cell level.

Across the full disaggregated evaluation, three conclusions remain
consistent. First, adaptive depth selection improves correction for
every dataset, backbone, cue type, and seed. Second, lower cue residual
does not require a reduction in average or aligned accuracy. Third, the
remaining gap to the oracle is consistent across settings, indicating
that more accurate layer prediction can further improve Evidence Reset
without increasing reset-module capacity.
\section*{I. Controlled Baselines and Ablations}

This section isolates the sources of improvement in Evidence Reset.
We distinguish three factors: access to cue-aware paired supervision,
the ability to reconstruct a cue-free hidden state, and the use of
inference dynamics to select an image--cue-specific intervention depth.

All results in this section are macro-averaged over Waterbirds and
UrbanCars and over ResNet-50 and ViT-B/16. Metrics are first computed
independently for each dataset--backbone combination and random seed,
and are then averaged with equal weight across the four experimental
cells. Reported uncertainty is the standard deviation over three
random seeds.

\subsection*{I1. Supervision Accounting and Controlled Comparison}

The conventional robustness baselines and the reset-based methods use
different forms of supervision. GroupDRO uses group annotations, JTT
uses a two-stage error-set construction, DFR uses a balanced retraining
set, and LLE uses shortcut-attribute structure in UrbanCars. These
methods provide important task-level references, but they do not
directly isolate the contribution of adaptive intervention depth.

The controlled reset baselines are designed for that purpose. Best
Fixed Reset, the dynamics-free selector, and Evidence Reset receive the
same training images, cue masks, cue-removed references, reset-module
architecture, optimization schedule, and parameter budget. Their main
difference is how the intervention layer is determined.

Table~\ref{tab:i_supervision} summarizes the information available to
each method.

\begin{table*}[t]
    \centering
    \scriptsize
    \setlength{\tabcolsep}{4pt}
    \renewcommand{\arraystretch}{1.10}
    \caption{
        Supervision and information available to the compared methods.
        Train/test entries indicate whether the information is required
        at the corresponding stage. The cue-removed test image and full
        test-time scan are used only by diagnostic or oracle methods.
    }
    \label{tab:i_supervision}
    \begin{tabular}{lcccccc}
        \toprule
        \textbf{Method}
        &
        \shortstack{\textbf{Group or}\\\textbf{shortcut labels}}
        &
        \shortstack{\textbf{Cue mask}\\\textbf{train / test}}
        &
        \shortstack{\textbf{Cue-free pair}\\\textbf{train / test}}
        &
        \shortstack{\textbf{Layer scan}\\\textbf{train / test}}
        &
        \shortstack{\textbf{Dynamics}\\\textbf{layer target}}
        &
        \shortstack{\textbf{Deployable}\\\textbf{without reference}}
        \\
        \midrule

        ERM
        & -- & -- / -- & -- / -- & -- / -- & -- & Yes
        \\

        GroupDRO
        & Train & -- / -- & -- / -- & -- / -- & -- & Yes
        \\

        JTT
        & -- & -- / -- & -- / -- & -- / -- & -- & Yes
        \\

        DFR
        & Retraining set & -- / -- & -- / -- & -- / -- & -- & Yes
        \\

        LLE
        & Shortcut attributes & -- / -- & -- / -- & -- / -- & -- & Yes
        \\

        \midrule

        Input-level Cue Removal
        & -- & Yes / Yes & Yes / Yes & -- / -- & -- & No
        \\

        Best Fixed Reset
        & -- & Yes / Yes & Yes / No & -- / -- & -- & Yes
        \\

        Dynamics-free Selector
        & -- & Yes / Yes & Yes / No & -- / -- & -- & Yes
        \\

        Evidence Reset
        & -- & Yes / Yes & Yes / No & Yes / No & Yes & Yes
        \\

        Oracle-layer Reset
        & -- & Yes / Yes & Yes / Yes & Yes / Yes & Yes & No
        \\

        \bottomrule
    \end{tabular}
\end{table*}

The comparison between Evidence Reset and conventional robustness
methods reflects the performance of complete learning systems under
their standard supervision. In contrast, the comparison with Best
Fixed Reset and the dynamics-free selector isolates the value of
adaptive depth selection under matched cue-aware supervision.

Input-level Cue Removal requires construction of an edited image at
test time. Oracle-layer Reset additionally requires the cue-free branch
and complete withdrawal trajectory at test time and is therefore used
only as a diagnostic upper bound. Evidence Reset requires a cue mask
but does not require a cue-removed reference or layer scan at inference.

\subsection*{I2. Isolating the Layer-selection Strategy}

We compare nine intervention-depth strategies while keeping the frozen
classifier and reset module fixed.

\paragraph{Random layer.}

A binary reset/no-reset head is trained using the same reset-availability
labels as the full selector. When reset is selected, the intervention
layer is sampled uniformly from the valid candidate set.

\paragraph{Fixed early, middle, and late layers.}

These variants use a fixed candidate depth for every reset example.
The selected normalized depths are \(0.25\), \(0.50\), and \(0.83\),
respectively. For each architecture, the nearest valid candidate block
is used.

\paragraph{Best fixed layer.}

One global candidate layer is selected for each dataset--backbone
setting using validation worst-group accuracy, with NCR used to resolve
ties.

\paragraph{Dynamics-free selector.}

This variant uses the same image--mask selector and the same candidate
output space as Evidence Reset, but receives no dynamics-derived layer
or no-reset labels. Layer selection is optimized end-to-end through a
straight-through Gumbel--Softmax estimator.

Let \(s_{\phi}(x,m_q)\) denote the selector logits. The relaxed layer
sample is

\begin{equation}
    \widetilde{\pi}_{\phi}
    =
    \operatorname{GumbelSoftmax}
    \left(
        s_{\phi}(x,m_q),
        \tau_g
    \right).
\end{equation}

The hard forward decision uses the largest component of
\(\widetilde{\pi}_{\phi}\), while the backward pass uses its continuous
relaxation. The temperature is annealed from

\begin{equation}
    \tau_g=1.0
\end{equation}

to

\begin{equation}
    \tau_g=0.1.
\end{equation}

The selector is optimized using only the downstream reset objectives:

\begin{equation}
    \mathcal{L}_{\mathrm{e2e}}
    =
    I_{\mathrm{reset}}
    \left(
        \mathcal{L}_{\mathrm{cls}}
        +
        \lambda_{\mathrm{feat}}
        \mathcal{L}_{\mathrm{feat}}
        +
        \lambda_{\mathrm{pred}}
        \mathcal{L}_{\mathrm{pred}}
    \right)
    +
    \lambda_{\mathrm{ent}}
    \mathcal{H}
    \left(
        \pi_{\phi}
    \right),
\end{equation}

where

\begin{equation}
    \lambda_{\mathrm{ent}}=0.01.
\end{equation}

This baseline has access to the same cue masks and cue-free training
pairs as Evidence Reset, but does not use locality, withdrawability,
lock-in onset, or dynamics-derived target layers.

\paragraph{Locality-only selection.}

The target layer maximizes

\begin{equation}
    \rho_l^{L}
    =
    1-\overline{L}_l.
\end{equation}

\paragraph{Withdrawability-only selection.}

The target layer maximizes

\begin{equation}
    \rho_l^{W}
    =
    \overline{W}_l
\end{equation}

within the valid pre-lock candidate set.

\paragraph{Full dynamics selection.}

Evidence Reset uses

\begin{equation}
    \rho_l
    =
    \left(
        1-\overline{L}_l
    \right)
    \overline{W}_l.
\end{equation}

\paragraph{Oracle layer.}

The dynamics-derived target layer is supplied directly at inference.

\begin{table*}[t]
    \centering
    \small
    \setlength{\tabcolsep}{7pt}
    \renewcommand{\arraystretch}{1.10}
    \caption{
        Ablation of the intervention-layer strategy. Results are
        macro-averaged across primary datasets and backbones. Layer
        error is measured relative to the dynamics-derived target and
        is reported only for input-dependent layer selectors. Lower NCR
        and layer error are better.
    }
    \label{tab:i_layer_selection}
    \begin{tabular}{lcccc}
        \toprule
        \textbf{Layer strategy}
        &
        \shortstack{\textbf{WG Acc.}\\\(\uparrow\)}
        &
        \shortstack{\textbf{NCR}\\\(\downarrow\)}
        &
        \shortstack{\textbf{Aligned}\\\(\uparrow\)}
        &
        \shortstack{\textbf{Layer Err.}\\\(\downarrow\)}
        \\
        \midrule

        Random layer
        &
        \(79.7\pm0.8\)
        &
        \(0.39\pm0.03\)
        &
        \(96.0\pm0.3\)
        &
        \(0.31\pm0.02\)
        \\

        Fixed early layer
        &
        \(80.8\pm0.8\)
        &
        \(0.35\pm0.03\)
        &
        \(95.9\pm0.3\)
        &
        --
        \\

        Fixed middle layer
        &
        \(82.4\pm0.7\)
        &
        \(0.30\pm0.02\)
        &
        \(96.4\pm0.3\)
        &
        --
        \\

        Fixed late layer
        &
        \(78.9\pm0.9\)
        &
        \(0.47\pm0.04\)
        &
        \(96.1\pm0.3\)
        &
        --
        \\

        Best fixed layer
        &
        \(82.7\pm0.7\)
        &
        \(0.29\pm0.02\)
        &
        \(96.5\pm0.3\)
        &
        --
        \\

        Dynamics-free selector
        &
        \(82.9\pm0.8\)
        &
        \(0.26\pm0.02\)
        &
        \(96.4\pm0.3\)
        &
        \(0.22\pm0.02\)
        \\

        Locality only
        &
        \(83.1\pm0.7\)
        &
        \(0.24\pm0.02\)
        &
        \(96.5\pm0.3\)
        &
        \(0.19\pm0.02\)
        \\

        Withdrawability only
        &
        \(83.8\pm0.7\)
        &
        \(0.22\pm0.02\)
        &
        \(96.6\pm0.3\)
        &
        \(0.16\pm0.01\)
        \\

        Full dynamics
        &
        \(\mathbf{85.6\pm0.6}\)
        &
        \(\mathbf{0.16\pm0.02}\)
        &
        \(\mathbf{96.9\pm0.2}\)
        &
        \(\mathbf{0.10\pm0.01}\)
        \\

        Oracle layer
        &
        \(87.0\pm0.6\)
        &
        \(0.11\pm0.01\)
        &
        \(97.0\pm0.2\)
        &
        \(0.00\)
        \\

        \bottomrule
    \end{tabular}
\end{table*}

Table~\ref{tab:i_layer_selection} shows that intervention depth is not
well approximated by a universal early, middle, or late stage. The
middle fixed layer is the strongest of the three predefined depths, but
remains below the validation-selected Best Fixed Reset and substantially
below Evidence Reset.

The dynamics-free selector improves NCR relative to Best Fixed Reset,
showing that input-dependent selection provides some benefit even
without explicit process supervision. Its worst-group accuracy remains
2.7 points below the full method, and its layer error is more than twice
as large. Optimizing only final reset outcomes does not reliably identify
the stage at which cue control begins to collapse.

Locality and withdrawability are individually informative. The
withdrawability-only selector performs better than the locality-only
selector because it directly reflects whether the selected support can
still influence the prediction. However, maximizing withdrawability
alone favors unnecessarily early layers. Combining declining locality
with retained withdrawability identifies a more effective intervention
window and improves worst-group accuracy by 1.8 points over the
withdrawability-only variant.

The remaining 1.4-point gap to the oracle is consistent with imperfect
layer selection rather than insufficient reset-module capacity.

\subsection*{I3. Training Objectives and the No-reset Option}

We next remove one training component at a time while retaining the
full dynamics-derived layer target. All variants use the same selector
and reset-module architectures.

Removing the feature-reconstruction objective tests whether prediction
alignment alone can recover an appropriate internal state. Removing the
prediction-alignment objective tests whether matching local hidden
features is sufficient to reproduce cue-free downstream behavior.
Removing classification supervision tests whether cue-free alignment
can damage task-relevant information. Finally, removing the no-reset
option forces an intervention on every example.

\begin{table*}[t]
    \centering
    \small
    \setlength{\tabcolsep}{5pt}
    \renewcommand{\arraystretch}{1.10}
    \caption{
        Ablation of the training objective and no-reset option. No-reset
        precision, recall, and F1 treat no-reset as the positive class.
        Results are macro-averaged across primary datasets and
        backbones.
    }
    \label{tab:i_objective_ablation}
    \begin{tabular}{lcccccc}
        \toprule
        \textbf{Variant}
        &
        \shortstack{\textbf{WG Acc.}\\\(\uparrow\)}
        &
        \shortstack{\textbf{NCR}\\\(\downarrow\)}
        &
        \shortstack{\textbf{Aligned}\\\(\uparrow\)}
        &
        \shortstack{\textbf{NR Prec.}\\\(\uparrow\)}
        &
        \shortstack{\textbf{NR Recall}\\\(\uparrow\)}
        &
        \shortstack{\textbf{NR F1}\\\(\uparrow\)}
        \\
        \midrule

        w/o \(\mathcal{L}_{\mathrm{feat}}\)
        &
        \(84.1\pm0.7\)
        &
        \(0.23\pm0.02\)
        &
        \(96.6\pm0.3\)
        &
        \(86.8\pm1.0\)
        &
        \(82.1\pm1.2\)
        &
        \(84.4\pm0.9\)
        \\

        w/o \(\mathcal{L}_{\mathrm{pred}}\)
        &
        \(84.6\pm0.7\)
        &
        \(0.20\pm0.02\)
        &
        \(96.6\pm0.3\)
        &
        \(87.1\pm0.9\)
        &
        \(82.9\pm1.1\)
        &
        \(84.9\pm0.9\)
        \\

        w/o \(\mathcal{L}_{\mathrm{cls}}\)
        &
        \(84.9\pm0.7\)
        &
        \(0.17\pm0.02\)
        &
        \(95.7\pm0.4\)
        &
        \(87.2\pm0.9\)
        &
        \(83.7\pm1.0\)
        &
        \(85.4\pm0.8\)
        \\

        w/o no-reset option
        &
        \(85.1\pm0.6\)
        &
        \(0.16\pm0.02\)
        &
        \(95.3\pm0.4\)
        &
        --
        &
        --
        &
        --
        \\

        Full objective
        &
        \(\mathbf{85.6\pm0.6}\)
        &
        \(\mathbf{0.16\pm0.02}\)
        &
        \(\mathbf{96.9\pm0.2}\)
        &
        \(\mathbf{88.4\pm0.8}\)
        &
        \(\mathbf{85.6\pm0.9}\)
        &
        \(\mathbf{87.0\pm0.7}\)
        \\

        \bottomrule
    \end{tabular}
\end{table*}

Feature reconstruction and prediction alignment contribute
complementary supervision. Without
\(\mathcal{L}_{\mathrm{feat}}\), the reset module can alter the final
prediction but leaves a substantially larger residual relative to the
cue-free branch. Without \(\mathcal{L}_{\mathrm{pred}}\), local hidden
states become closer to their cue-free targets, but small
representation errors are amplified by the frozen downstream network.

Removing classification supervision produces relatively low NCR but
reduces cue-aligned accuracy by 1.2 points. This confirms that matching
the cue-free execution alone does not guarantee preservation of all
task-relevant information.

Forcing every example to reset has little effect on aggregate NCR but
reduces aligned accuracy by 1.6 points. The no-reset option therefore
primarily protects examples for which intervention is unnecessary,
non-actionable, or likely to disturb an already reliable prediction.

The full model attains a no-reset precision of 88.4\%, recall of
85.6\%, and F1 of 87.0\%. The balance between precision and recall is
important: excessive no-reset predictions leave harmful cue dependence
uncorrected, whereas excessive reset predictions reduce aligned
accuracy.

\subsection*{I4. Reset Architecture and Capacity Controls}

We finally test whether the improvement is explained by reset-module
capacity rather than dynamics-based depth selection. The default module
uses a shared reconstruction core, layer-specific channel projections
where required, and an explicit layer embedding.

Parameter multipliers in Table~\ref{tab:i_capacity} are measured
relative to the default reset module within each backbone and then
macro-averaged. Inference overhead is the percentage increase in
single-image latency relative to the frozen classifier, including the
selector and one reset execution.

\begin{table*}[t]
    \centering
    \small
    \setlength{\tabcolsep}{6pt}
    \renewcommand{\arraystretch}{1.10}
    \caption{
        Reset-module architecture and capacity controls. Parameter
        multipliers are relative to the default reset module. Inference
        overhead includes the layer selector and one reset operation.
    }
    \label{tab:i_capacity}
    \begin{tabular}{lccccc}
        \toprule
        \textbf{Variant}
        &
        \shortstack{\textbf{Parameter}\\\textbf{multiplier}}
        &
        \shortstack{\textbf{WG Acc.}\\\(\uparrow\)}
        &
        \shortstack{\textbf{NCR}\\\(\downarrow\)}
        &
        \shortstack{\textbf{Aligned}\\\(\uparrow\)}
        &
        \shortstack{\textbf{Inference}\\\textbf{overhead}}
        \\
        \midrule

        Half-width shared core
        &
        \(0.48\times\)
        &
        \(84.7\pm0.7\)
        &
        \(0.20\pm0.02\)
        &
        \(96.8\pm0.3\)
        &
        \(+4.6\%\)
        \\

        Default shared conditioned core
        &
        \(1.00\times\)
        &
        \(\mathbf{85.6\pm0.6}\)
        &
        \(0.16\pm0.02\)
        &
        \(\mathbf{96.9\pm0.2}\)
        &
        \(+7.8\%\)
        \\

        \(1.5\times\)-width shared core
        &
        \(1.91\times\)
        &
        \(85.7\pm0.6\)
        &
        \(\mathbf{0.15\pm0.01}\)
        &
        \(96.9\pm0.2\)
        &
        \(+12.9\%\)
        \\

        Layer-specific reconstruction cores
        &
        \(5.26\times\)
        &
        \(85.8\pm0.6\)
        &
        \(0.15\pm0.01\)
        &
        \(96.9\pm0.2\)
        &
        \(+23.4\%\)
        \\

        No layer conditioning
        &
        \(0.99\times\)
        &
        \(84.8\pm0.7\)
        &
        \(0.20\pm0.02\)
        &
        \(96.8\pm0.3\)
        &
        \(+7.6\%\)
        \\

        Unmasked reset input
        &
        \(1.00\times\)
        &
        \(83.2\pm0.8\)
        &
        \(0.28\pm0.03\)
        &
        \(96.8\pm0.3\)
        &
        \(+7.8\%\)
        \\

        \bottomrule
    \end{tabular}
\end{table*}

Doubling the effective reset capacity produces only marginal
improvement over the default module. A layer-specific reconstruction
core increases the reset parameter count by more than five times but
improves worst-group accuracy by only 0.2 points. The advantage of
Evidence Reset over fixed-depth correction therefore cannot be
attributed to a larger per-layer model.

Layer conditioning is more important than additional width. Removing
the layer embedding reduces worst-group accuracy by 0.8 points and
raises NCR from 0.16 to 0.20. Hidden states at different depths have
different semantic and spatial statistics even after channel
projection, and one unconditioned reconstruction rule does not transfer
equally well across them.

Providing the original masked states to the reset module produces the
largest degradation among the architecture controls. The module learns
to copy cue-aligned activations rather than reconstruct their cue-free
counterparts, yielding an NCR of 0.28. Explicitly removing the masked
states from the reconstruction input is therefore necessary to prevent
a trivial identity solution.

Taken together, the controlled comparisons identify the source of the
main improvement. Cue-aware paired supervision enables hidden-state
reconstruction, but does not by itself explain the result: Best Fixed
Reset and the dynamics-free selector receive the same information.
Additional reset capacity also provides little benefit. The largest
gain arises when locality and withdrawability are jointly used to
identify an image--cue-specific intervention depth before stable
lock-in.
\section*{J. Inference-time Cue Masks and Multi-cue Behavior}

Evidence Reset receives a spatial cue mask at inference. We therefore
evaluate two practical aspects of this input. First, we measure how
gradual localization errors affect layer selection and cue withdrawal.
Second, we examine how the method behaves when multiple shortcuts are
present in the same image and can be reset individually, jointly, or
sequentially.

The mask-robustness analysis uses Waterbirds because its foreground
segmentation provides an exact reference from which controlled mask
perturbations can be generated. The multi-cue analysis uses UrbanCars,
where background and co-occurring-object shortcuts are independently
available through the native compositing pipeline.

\subsection*{J1. Cue-mask Sources and Perturbation Protocol}

The clean cue masks are obtained from the spatial annotations or
component masks available for each benchmark. For Waterbirds, the
background cue mask is the complement of the bird foreground mask:

\begin{equation}
    m_q^{\mathrm{bg}}
    =
    \mathbf{1}
    -
    m_{\mathrm{bird}}.
\end{equation}

For UrbanCars, the background mask excludes both the target car and the
co-occurring object:

\begin{equation}
    m_q^{\mathrm{bg}}
    =
    \mathbf{1}
    -
    \max
    \left(
        m_{\mathrm{car}},
        m_{\mathrm{obj}}
    \right),
\end{equation}

where the maximum is taken elementwise. The localized object mask is

\begin{equation}
    m_q^{\mathrm{obj}}
    =
    m_{\mathrm{obj}}.
\end{equation}

For Spawrious-M2M-Hard, the cue mask is obtained from the accepted
DINO-X--SAM 2.1 foreground segmentation. In all cases, the same cue
mask is provided to the layer selector and the reset module.

To evaluate robustness, we generate five families of perturbed masks:
erosion, dilation, spatial translation, missing-region corruption, and
false-positive-region corruption. The clean mask is denoted by
\(m_q^{(0)}\). The mild, moderate, and severe perturbation levels use

\begin{equation}
    \alpha_s
    \in
    \left\{
        0.10,
        0.20,
        0.40
    \right\},
\end{equation}

where \(\alpha_s\) specifies the fraction of the original cue-mask area
removed or added.

\paragraph{Mask erosion.}

Erosion simulates under-segmentation along the cue boundary. The
structuring-element radius is selected so that the retained mask area
is as close as possible to

\begin{equation}
    \left\|
        m_q^{\mathrm{ero},s}
    \right\|_1
    =
    \left(
        1-\alpha_s
    \right)
    \left\|
        m_q^{(0)}
    \right\|_1.
\end{equation}

For a background cue, erosion removes part of the true background
support from the intervention mask.

\paragraph{Mask dilation.}

Dilation simulates moderate over-segmentation by including additional
non-cue pixels. The target mask area is

\begin{equation}
    \left\|
        m_q^{\mathrm{dil},s}
    \right\|_1
    =
    \min
    \left[
        HW,
        \left(
            1+\alpha_s
        \right)
        \left\|
            m_q^{(0)}
        \right\|_1
    \right].
\end{equation}

When the target area cannot be reached because the original mask already
covers most of the image, dilation continues until the image boundary
is reached.

\paragraph{Spatial shift.}

Spatial translation moves the entire mask without changing its shape.
The mild, moderate, and severe shifts correspond to

\begin{equation}
    \beta_s
    \in
    \left\{
        0.02,
        0.05,
        0.10
    \right\}
\end{equation}

of the image width and height. For each source image, the mask is shifted
in the four cardinal directions by

\begin{equation}
    \left(
        \Delta x,
        \Delta y
    \right)
    \in
    \left\{
        \left(
            \beta_s W,0
        \right),
        \left(
            -\beta_s W,0
        \right),
        \left(
            0,\beta_s H
        \right),
        \left(
            0,-\beta_s H
        \right)
    \right\}.
\end{equation}

Metrics are averaged over the four directions. Pixels translated beyond
the image boundary are discarded, and newly exposed locations are set
to zero.

\paragraph{Missing-region corruption.}

Missing-region corruption removes connected portions from inside the
true cue mask. Let \(r_s^{\mathrm{miss}}\) denote a binary removal mask
sampled from the support of \(m_q^{(0)}\). The corrupted mask is

\begin{equation}
    m_q^{\mathrm{miss},s}
    =
    m_q^{(0)}
    \odot
    \left(
        \mathbf{1}
        -
        r_s^{\mathrm{miss}}
    \right),
\end{equation}

subject to

\begin{equation}
    \left\|
        r_s^{\mathrm{miss}}
    \right\|_1
    =
    \alpha_s
    \left\|
        m_q^{(0)}
    \right\|_1.
\end{equation}

The removed regions are generated from randomly placed connected
ellipses rather than isolated pixels, producing spatially coherent
under-coverage.

\paragraph{False-positive-region corruption.}

False-positive corruption adds connected regions outside the true cue
mask. Let \(r_s^{\mathrm{fp}}\) satisfy

\begin{equation}
    r_s^{\mathrm{fp}}
    \odot
    m_q^{(0)}
    =
    \mathbf{0}.
\end{equation}

The perturbed mask is

\begin{equation}
    m_q^{\mathrm{fp},s}
    =
    \max
    \left(
        m_q^{(0)},
        r_s^{\mathrm{fp}}
    \right),
\end{equation}

with

\begin{equation}
    \left\|
        r_s^{\mathrm{fp}}
    \right\|_1
    =
    \alpha_s
    \left\|
        m_q^{(0)}
    \right\|_1.
\end{equation}

For both random missing regions and false-positive regions, three
independent perturbation realizations are generated for every source
mask using fixed seeds. Predictions are evaluated separately and then
averaged at the source-image level.

All model weights remain fixed during this evaluation. The perturbed
mask replaces the clean mask for both layer selection and hidden-state
reconstruction; the original image and all other inference inputs remain
unchanged.

\subsection*{J2. Robustness to Inaccurate Cue Masks}

\begin{figure}[t]
    \centering
    \includegraphics[
        width=\columnwidth
    ]{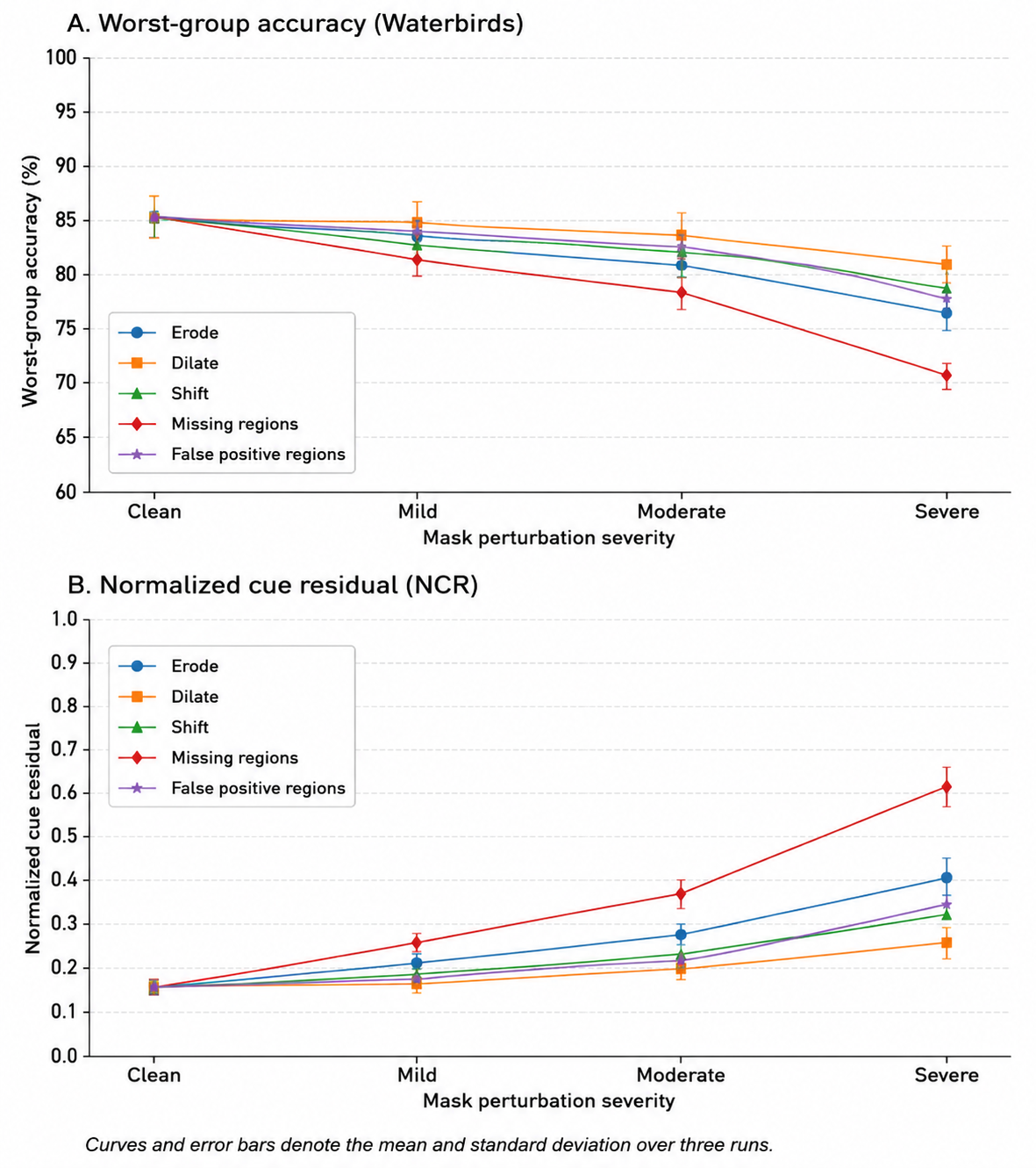}
    \caption{
        Robustness to inaccurate inference-time cue masks on
        Waterbirds.
        \textbf{A:} Worst-group accuracy under progressively stronger
        mask erosion, dilation, spatial shift, missing-region, and
        false-positive-region perturbations.
        \textbf{B:} Normalized cue residual under the same
        perturbations. Curves and error bars denote the mean and
        standard deviation over three runs.
    }
    \label{fig:j_mask_robustness}
\end{figure}

Figure~\ref{fig:j_mask_robustness} reports worst-group accuracy and
normalized cue residual under progressively stronger perturbations of
the Waterbirds background mask.

With the clean mask, Evidence Reset achieves a worst-group accuracy of
85.4\% and an NCR of 0.15 in this evaluation. Mild perturbations produce
limited degradation. Worst-group accuracy remains between 81.4\% and
84.8\%, while NCR remains between 0.16 and 0.26 across the five
perturbation types.

Dilation is the least harmful corruption. Under severe dilation,
worst-group accuracy remains 80.9\% and NCR increases to 0.26. The
additional masked pixels introduce some unnecessary reconstruction but
leave the true cue support fully covered. False-positive regions behave
similarly under mild and moderate corruption, although severe
false-positive coverage produces a larger reduction in accuracy because
the reset increasingly modifies task-relevant foreground states.

Erosion is more damaging than comparable dilation. Severe erosion
reduces worst-group accuracy to 76.4\% and raises NCR to 0.41. Removing
part of the true cue support leaves cue-aligned states unchanged, so the
reset execution cannot fully approach the cue-free prediction.

Spatial shifts produce an intermediate degradation. Small shifts retain
substantial overlap with the true background support, whereas severe
translation simultaneously creates missing and false-positive regions.
At the severe level, worst-group accuracy is 78.9\% and NCR is 0.32.

Missing-region corruption is the most harmful perturbation. Under the
severe setting, worst-group accuracy falls to 70.7\% and NCR rises to
0.61. Unlike boundary erosion, spatially distributed holes leave
multiple separated cue-aligned states untouched, allowing residual cue
influence to persist through several regions of the representation.

The selector degrades gradually under the same perturbations. Averaged
over the five corruption types, normalized layer error increases from
0.10 with clean masks to 0.12, 0.16, and 0.23 under mild, moderate, and
severe perturbations, respectively. The no-reset rate increases from
11.6\% to 13.2\%, 17.4\%, and 24.9\%. Thus, inaccurate masks affect both
the predicted correction depth and the reconstructed support, but small
localization errors do not cause an abrupt failure.

The asymmetric effect of under- and over-coverage is consistent across
the task and cue-residual metrics. Moderate over-coverage is generally
preferable to leaving portions of the true cue unmodified. This
suggests that an external localization system should favor slightly
conservative masks when exact cue boundaries are uncertain.

\subsection*{J3. Joint and Sequential Multi-cue Reset}

UrbanCars contains two independently controlled shortcuts: the scene
background and a localized co-occurring object. We evaluate five
correction strategies in addition to the unmodified classifier.

\paragraph{Single-cue reset.}

Background-only reset uses \(m_q^{\mathrm{bg}}\) and modifies only the
background-cue representation. Object-only reset uses
\(m_q^{\mathrm{obj}}\). The other shortcut remains present in the input
and is not included in the intervention mask.

\paragraph{Joint reset.}

Joint reset uses the union mask

\begin{equation}
    m_q^{\mathrm{joint}}
    =
    \max
    \left(
        m_q^{\mathrm{bg}},
        m_q^{\mathrm{obj}}
    \right)
\end{equation}

and a matched reference in which both shortcut components are removed.
A single selector decision and one reset operation are used for the
combined support.

\paragraph{Sequential reset.}

Sequential strategies allow the two cues to use different intervention
depths. The selectors first produce cue-specific layer decisions

\begin{equation}
    \widehat{l}_{\mathrm{bg}}
    =
    \arg\max_l
    \pi_{\phi}
    \left(
        l
        \mid
        x,
        m_q^{\mathrm{bg}}
    \right)
\end{equation}

and

\begin{equation}
    \widehat{l}_{\mathrm{obj}}
    =
    \arg\max_l
    \pi_{\phi}
    \left(
        l
        \mid
        x,
        m_q^{\mathrm{obj}}
    \right).
\end{equation}

For background-to-object reset, the object intervention is constrained
to occur at or after the selected background layer:

\begin{equation}
    \widehat{l}_{\mathrm{obj}}^{\,\mathrm{seq}}
    =
    \min
    \left\{
        l
        \in
        \mathcal{L}
        \ \middle|\
        l
        \geq
        \widehat{l}_{\mathrm{bg}},
        \ 
        l
        \geq
        \widehat{l}_{\mathrm{obj}}
    \right\}.
\end{equation}

The background reset is applied first, followed by the object reset
during the same forward execution. Object-to-background reset uses the
symmetric constraint with the object layer preceding the background
layer. If either selector chooses no-reset, only the remaining cue is
modified.

Cue-specific NCR is computed relative to the corresponding single-cue
reference. Joint NCR is computed relative to the matched reference in
which both cues are removed. For the unmodified classifier, each NCR is
approximately one by construction.

\begin{table*}[t]
    \centering
    \small
    \setlength{\tabcolsep}{5pt}
    \renewcommand{\arraystretch}{1.10}
    \caption{
        Multi-cue intervention results on UrbanCars, averaged over
        ResNet-50 and ViT-B/16. BG NCR and Obj. NCR measure residual
        dependence relative to the background-removed and
        object-removed references, respectively. Joint NCR uses the
        reference in which both shortcuts are removed. Results are mean
        \(\pm\) standard deviation over three runs. Lower residuals are
        better.
    }
    \label{tab:j_multicue}
    \begin{tabular}{lcccccc}
        \toprule
        \textbf{Strategy}
        &
        \textbf{Avg.}
        &
        \textbf{WG}
        &
        \textbf{Aligned}
        &
        \textbf{BG NCR}
        &
        \textbf{Obj. NCR}
        &
        \textbf{Joint NCR}
        \\
        \midrule

        No reset
        &
        \(94.7\pm0.4\)
        &
        \(47.7\pm1.8\)
        &
        \(98.3\pm0.2\)
        &
        \(1.00\pm0.00\)
        &
        \(1.00\pm0.00\)
        &
        \(1.00\pm0.00\)
        \\

        Background only
        &
        \(94.6\pm0.3\)
        &
        \(68.1\pm1.1\)
        &
        \(97.4\pm0.3\)
        &
        \(0.16\pm0.02\)
        &
        \(1.05\pm0.03\)
        &
        \(0.55\pm0.03\)
        \\

        Object only
        &
        \(94.5\pm0.3\)
        &
        \(65.8\pm1.2\)
        &
        \(97.2\pm0.3\)
        &
        \(1.04\pm0.03\)
        &
        \(0.20\pm0.02\)
        &
        \(0.58\pm0.03\)
        \\

        Joint reset
        &
        \(94.2\pm0.4\)
        &
        \(78.3\pm0.8\)
        &
        \(95.8\pm0.4\)
        &
        \(0.24\pm0.02\)
        &
        \(0.27\pm0.02\)
        &
        \(0.19\pm0.02\)
        \\

        BG \(\rightarrow\) Obj.
        &
        \(\mathbf{94.4\pm0.3}\)
        &
        \(\mathbf{80.1\pm0.7}\)
        &
        \(\mathbf{96.2\pm0.3}\)
        &
        \(\mathbf{0.17\pm0.02}\)
        &
        \(\mathbf{0.20\pm0.02}\)
        &
        \(\mathbf{0.14\pm0.01}\)
        \\

        Obj. \(\rightarrow\) BG
        &
        \(94.3\pm0.3\)
        &
        \(79.2\pm0.8\)
        &
        \(96.0\pm0.3\)
        &
        \(0.19\pm0.02\)
        &
        \(0.22\pm0.02\)
        &
        \(0.16\pm0.01\)
        \\

        \bottomrule
    \end{tabular}
\end{table*}

Single-cue reset substantially reduces dependence on the targeted
shortcut but leaves the other shortcut active. Background-only reset
reduces BG NCR from 1.00 to 0.16, while object NCR increases slightly
to 1.05. Object-only reset produces the complementary pattern: object
NCR decreases to 0.20, while BG NCR rises to 1.04.

This small cross-cue increase indicates that weakening one shortcut can
raise the relative contribution of the remaining shortcut. The effect
is limited in magnitude, but it explains why single-cue reset does not
fully recover worst-group performance in the eight-group UrbanCars
evaluation.

Joint reset reduces both cue-specific residuals and reaches a joint NCR
of 0.19. Its single combined intervention, however, must use one layer
for two cues with different spatial extents and dynamics. It also
modifies a larger hidden-state region at once, reducing aligned accuracy
to 95.8\%.

Sequential reset performs better because the background and localized
object can use different intervention depths. Background-to-object reset
achieves the highest worst-group accuracy, 80.1\%, and the lowest joint
NCR, 0.14. Object-to-background reset remains effective but is
0.9 worst-group points lower and leaves slightly larger cue residuals.

The ordering is consistent with the spatial structure of the two cues.
The background occupies a broad region and typically begins to affect
distributed contextual features earlier. Resetting it first reduces the
global shortcut contribution before the localized object representation
is reconstructed. Reversing the order allows the subsequent background
reset to modify contextual states surrounding the already corrected
object support.

Compared with the unmodified classifier, background-to-object reset
improves worst-group accuracy by 32.4 percentage points while retaining
94.4\% average accuracy. Relative to joint reset, it gains
1.8 worst-group points, improves aligned accuracy by 0.4 points, and
reduces joint NCR by 0.05.

These results show that cue masks provide more than a binary indication
of whether to intervene. They determine which internal states are
reconstructed, influence the selected intervention depth, and allow
multiple shortcuts to be corrected at distinct stages. Moderate mask
errors remain tolerable, but missing cue regions and treating
dynamically different shortcuts as one combined support both reduce the
completeness of cue withdrawal.
\section*{K. Additional Datasets, Sensitivity, and Efficiency}

We further evaluate the generality of inference dynamics beyond the two
primary controlled benchmarks, examine sensitivity to the thresholds
and training configurations used by the framework, and report the
computational cost of both offline dynamics construction and test-time
Evidence Reset.

Unless otherwise stated, sensitivity results are macro-averaged across
Waterbirds and UrbanCars and across ResNet-50 and ViT-B/16. Each
dataset--backbone combination is assigned equal weight. All reported
model results are mean \(\pm\) standard deviation over three random
seeds.

\subsection*{K1. Results on Spawrious-M2M-Hard and ImageNet-9}

Spawrious-M2M-Hard extends the evaluation to a many-to-many
class--environment correlation with foreground masks obtained from the
cross-model segmentation pipeline. ImageNet-9 provides broader object
categories and controlled foreground--background variants that do not
share the native compositing procedure of Waterbirds or UrbanCars.

For Spawrious-M2M-Hard, the task metric is out-of-distribution test
accuracy over the held-out class--environment combinations. For
ImageNet-9, the task metric in
Table~\ref{tab:k_additional_datasets} is accuracy on Mixed-Rand, where
foreground objects are paired with backgrounds sampled independently
of their superclass. Worst-group accuracy is not used for ImageNet-9
because its benchmark variants define distribution-level background
conditions rather than the group structure used by Waterbirds and
UrbanCars.

\begin{table*}[t]
    \centering
    \small
    \setlength{\tabcolsep}{4.5pt}
    \renewcommand{\arraystretch}{1.10}
    \caption{
        Inference dynamics and correction results on additional
        datasets. Eval. Acc. denotes OOD accuracy for
        Spawrious-M2M-Hard and Mixed-Rand accuracy for ImageNet-9.
        Locality-first and stable-lock-in rates are computed over
        validated image--cue pairs. Lock-in depth is reported as median
        normalized depth with interquartile range. Lower NCR is better.
    }
    \label{tab:k_additional_datasets}
    \begin{tabular}{llccccccc}
        \toprule
        \textbf{Dataset}
        &
        \textbf{Backbone}
        &
        \shortstack{\textbf{Locality}\\\textbf{first (\%)}}
        &
        \shortstack{\textbf{Stable}\\\textbf{lock-in (\%)}}
        &
        \shortstack{\textbf{Median}\\\(d_{\mathrm{lock}}\)}
        &
        \shortstack{\textbf{Fixed}\\\textbf{Eval. Acc.}}
        &
        \shortstack{\textbf{Evidence}\\\textbf{Eval. Acc.}}
        &
        \shortstack{\textbf{Fixed}\\\textbf{NCR}}
        &
        \shortstack{\textbf{Evidence}\\\textbf{NCR}}
        \\
        \midrule

        Spawrious-M2M-Hard
        &
        ResNet-50
        &
        \(78.4\pm2.5\)
        &
        \(85.7\pm2.1\)
        &
        \(0.64\,[0.56,0.73]\)
        &
        \(71.8\pm0.9\)
        &
        \(\mathbf{74.6\pm0.8}\)
        &
        \(0.31\pm0.02\)
        &
        \(\mathbf{0.18\pm0.02}\)
        \\

        Spawrious-M2M-Hard
        &
        ViT-B/16
        &
        \(81.6\pm2.2\)
        &
        \(88.9\pm1.8\)
        &
        \(0.60\,[0.52,0.69]\)
        &
        \(74.1\pm0.8\)
        &
        \(\mathbf{76.9\pm0.7}\)
        &
        \(0.28\pm0.02\)
        &
        \(\mathbf{0.16\pm0.01}\)
        \\

        ImageNet-9
        &
        ResNet-50
        &
        \(76.9\pm2.6\)
        &
        \(83.8\pm2.3\)
        &
        \(0.67\,[0.58,0.76]\)
        &
        \(62.4\pm0.7\)
        &
        \(\mathbf{64.7\pm0.6}\)
        &
        \(0.34\pm0.03\)
        &
        \(\mathbf{0.21\pm0.02}\)
        \\

        ImageNet-9
        &
        ViT-B/16
        &
        \(80.3\pm2.3\)
        &
        \(87.1\pm2.0\)
        &
        \(0.63\,[0.55,0.72]\)
        &
        \(65.8\pm0.7\)
        &
        \(\mathbf{68.2\pm0.6}\)
        &
        \(0.30\pm0.02\)
        &
        \(\mathbf{0.18\pm0.02}\)
        \\

        \bottomrule
    \end{tabular}
\end{table*}

The locality-first ordering remains prevalent on both additional
datasets. It appears in 76.9--81.6\% of validated pairs, while
83.8--88.9\% exhibit stable lock-in. These rates are slightly lower
than those on Waterbirds and UrbanCars, reflecting the greater
variation introduced by automatically constructed masks and
foreground--background references.

The median lock-in onset remains architecture- and
dataset-dependent. ViT-B/16 locks in earlier than ResNet-50 on both
datasets, consistent with its global token interaction. ImageNet-9
produces the latest median onset, particularly for ResNet-50, because
many of its foreground objects occupy a larger fraction of the image
and remain spatially distinguishable for more layers.

Evidence Reset improves the evaluation accuracy of every
dataset--backbone combination. The gain over Best Fixed Reset is
2.8 points on both Spawrious backbones and 2.3--2.4 points on
ImageNet-9. NCR decreases by 0.12--0.13 across the four settings,
showing that the accuracy gains are accompanied by closer alignment to
the independently constructed cue-free execution.

For ImageNet-9, we additionally report accuracy on all controlled
background variants.

\begin{table*}[t]
    \centering
    \small
    \setlength{\tabcolsep}{5pt}
    \renewcommand{\arraystretch}{1.10}
    \caption{
        ImageNet-9 accuracy across background variants. Mixed-Same uses
        a background from the same superclass, Mixed-Rand uses an
        independently sampled background, and Only-FG removes natural
        background appearance. Gap denotes Mixed-Same accuracy minus
        Mixed-Rand accuracy; a smaller gap indicates lower background
        sensitivity.
    }
    \label{tab:k_imagenet9_variants}
    \begin{tabular}{llccccc}
        \toprule
        \textbf{Backbone}
        &
        \textbf{Method}
        &
        \textbf{Original}
        &
        \textbf{Mixed-Same}
        &
        \textbf{Mixed-Rand}
        &
        \textbf{Only-FG}
        &
        \textbf{Gap}
        \\
        \midrule

        ResNet-50
        &
        ERM
        &
        \(76.2\pm0.3\)
        &
        \(72.8\pm0.4\)
        &
        \(58.7\pm0.6\)
        &
        \(60.1\pm0.6\)
        &
        \(14.1\)
        \\

        ResNet-50
        &
        Best Fixed Reset
        &
        \(75.8\pm0.3\)
        &
        \(72.0\pm0.4\)
        &
        \(62.4\pm0.7\)
        &
        \(63.2\pm0.6\)
        &
        \(9.6\)
        \\

        ResNet-50
        &
        Evidence Reset
        &
        \(\mathbf{76.0\pm0.3}\)
        &
        \(\mathbf{72.4\pm0.4}\)
        &
        \(\mathbf{64.7\pm0.6}\)
        &
        \(\mathbf{65.0\pm0.5}\)
        &
        \(\mathbf{7.7}\)
        \\

        \midrule

        ViT-B/16
        &
        ERM
        &
        \(79.4\pm0.3\)
        &
        \(75.6\pm0.4\)
        &
        \(62.5\pm0.6\)
        &
        \(64.0\pm0.5\)
        &
        \(13.1\)
        \\

        ViT-B/16
        &
        Best Fixed Reset
        &
        \(79.0\pm0.3\)
        &
        \(74.9\pm0.4\)
        &
        \(65.8\pm0.7\)
        &
        \(66.8\pm0.5\)
        &
        \(9.1\)
        \\

        ViT-B/16
        &
        Evidence Reset
        &
        \(\mathbf{79.2\pm0.3}\)
        &
        \(\mathbf{75.2\pm0.3}\)
        &
        \(\mathbf{68.2\pm0.6}\)
        &
        \(\mathbf{68.7\pm0.5}\)
        &
        \(\mathbf{7.0}\)
        \\

        \bottomrule
    \end{tabular}
\end{table*}

Evidence Reset preserves Original and Mixed-Same accuracy while
improving Mixed-Rand and Only-FG performance. The Mixed-Same to
Mixed-Rand gap decreases from 14.1 to 7.7 points for ResNet-50 and from
13.1 to 7.0 points for ViT-B/16. The similar gains on Mixed-Rand and
Only-FG indicate that the result does not depend on retaining a
particular replacement background.

\subsection*{K2. Sensitivity to Dynamics Thresholds}

We vary the cue-relevance threshold
\(\tau_{\mathrm{rel}}\), withdrawability threshold
\(\tau_{\mathrm{wd}}\), consecutive-layer requirement \(r\), and
actionability threshold \(\tau_{\mathrm{act}}\). One parameter is
changed at a time while all remaining training and evaluation settings
are fixed.

\begin{table*}[t]
    \centering
    \scriptsize
    \setlength{\tabcolsep}{4.5pt}
    \renewcommand{\arraystretch}{1.08}
    \caption{
        Sensitivity to dynamics thresholds. Results are
        macro-averaged across primary datasets and backbones.
        \(d_{\mathrm{lock}}\) is the median normalized onset among
        stable-lock-in pairs. The default configuration is marked in
        bold. Lower NCR is better.
    }
    \label{tab:k_threshold_sensitivity}
    \begin{tabular}{lccccccc}
        \toprule
        \textbf{Parameter}
        &
        \textbf{Value}
        &
        \shortstack{\textbf{Locality}\\\textbf{first (\%)}}
        &
        \shortstack{\textbf{Stable}\\\textbf{lock-in (\%)}}
        &
        \shortstack{\textbf{Median}\\\(d_{\mathrm{lock}}\)}
        &
        \shortstack{\textbf{WG Acc.}\\\(\uparrow\)}
        &
        \shortstack{\textbf{NCR}\\\(\downarrow\)}
        &
        \shortstack{\textbf{Aligned}\\\(\uparrow\)}
        \\
        \midrule

        \(\tau_{\mathrm{rel}}\)
        & \(0.25\)
        & \(81.3\)
        & \(87.5\)
        & \(0.61\)
        & \(85.2\)
        & \(0.17\)
        & \(96.8\)
        \\

        \(\tau_{\mathrm{rel}}\)
        & \(\mathbf{0.50}\)
        & \(\mathbf{84.0}\)
        & \(\mathbf{89.9}\)
        & \(\mathbf{0.62}\)
        & \(\mathbf{85.6}\)
        & \(\mathbf{0.16}\)
        & \(\mathbf{96.9}\)
        \\

        \(\tau_{\mathrm{rel}}\)
        & \(0.75\)
        & \(85.1\)
        & \(90.8\)
        & \(0.62\)
        & \(85.5\)
        & \(0.16\)
        & \(96.8\)
        \\

        \(\tau_{\mathrm{rel}}\)
        & \(1.00\)
        & \(86.0\)
        & \(91.6\)
        & \(0.63\)
        & \(85.1\)
        & \(0.17\)
        & \(96.7\)
        \\

        \midrule

        \(\tau_{\mathrm{wd}}\)
        & \(0.10\)
        & \(83.8\)
        & \(78.6\)
        & \(0.69\)
        & \(85.2\)
        & \(0.17\)
        & \(96.9\)
        \\

        \(\tau_{\mathrm{wd}}\)
        & \(\mathbf{0.20}\)
        & \(\mathbf{84.0}\)
        & \(\mathbf{89.9}\)
        & \(\mathbf{0.62}\)
        & \(\mathbf{85.6}\)
        & \(\mathbf{0.16}\)
        & \(\mathbf{96.9}\)
        \\

        \(\tau_{\mathrm{wd}}\)
        & \(0.30\)
        & \(84.2\)
        & \(94.1\)
        & \(0.56\)
        & \(85.1\)
        & \(0.18\)
        & \(96.7\)
        \\

        \midrule

        \(r\)
        & \(2\)
        & \(83.6\)
        & \(93.7\)
        & \(0.59\)
        & \(85.3\)
        & \(0.17\)
        & \(96.8\)
        \\

        \(r\)
        & \(\mathbf{3}\)
        & \(\mathbf{84.0}\)
        & \(\mathbf{89.9}\)
        & \(\mathbf{0.62}\)
        & \(\mathbf{85.6}\)
        & \(\mathbf{0.16}\)
        & \(\mathbf{96.9}\)
        \\

        \(r\)
        & \(4\)
        & \(84.1\)
        & \(86.1\)
        & \(0.64\)
        & \(85.4\)
        & \(0.17\)
        & \(96.9\)
        \\

        \midrule

        \(\tau_{\mathrm{act}}\)
        & \(0.05\)
        & \(84.0\)
        & \(89.9\)
        & \(0.62\)
        & \(85.4\)
        & \(0.16\)
        & \(96.6\)
        \\

        \(\tau_{\mathrm{act}}\)
        & \(\mathbf{0.10}\)
        & \(\mathbf{84.0}\)
        & \(\mathbf{89.9}\)
        & \(\mathbf{0.62}\)
        & \(\mathbf{85.6}\)
        & \(\mathbf{0.16}\)
        & \(\mathbf{96.9}\)
        \\

        \(\tau_{\mathrm{act}}\)
        & \(0.20\)
        & \(84.0\)
        & \(89.9\)
        & \(0.62\)
        & \(85.0\)
        & \(0.18\)
        & \(97.1\)
        \\

        \bottomrule
    \end{tabular}
\end{table*}

The relevance threshold changes the size and composition of the
validated pair set. Relative to all quality-approved candidate pairs,
62.8\%, 51.4\%, 43.1\%, and 35.7\% are retained at
\(\tau_{\mathrm{rel}}\) values of \(0.25\), \(0.50\), \(0.75\), and
\(1.00\), respectively. Higher thresholds select examples with stronger
model-relative cue effects and therefore slightly increase the observed
locality-first and stable-lock-in rates. Correction performance remains
within 0.5 worst-group points across the tested range.

The withdrawability threshold affects the absolute onset more directly.
A stricter threshold of \(0.10\) requires a larger loss of local control
and therefore detects fewer, later lock-in transitions. A threshold of
\(0.30\) detects more and earlier transitions. The default value of
\(0.20\) provides the strongest balance between worst-group accuracy
and NCR.

Requiring more consecutive layers reduces the number of stable-lock-in
examples and shifts the onset slightly deeper. The performance variation
from \(r=2\) to \(r=4\) remains small, indicating that the method does
not depend on treating isolated layer fluctuations as stable
transitions.

The actionability threshold controls the trade-off between correction
coverage and preservation. A low threshold triggers reset for weak
intervention opportunities and slightly reduces aligned accuracy. A
high threshold increases no-reset decisions, preserving aligned
performance but leaving more residual cue influence. The default
threshold achieves the highest worst-group accuracy without increasing
NCR.

\subsection*{K3. Sensitivity to Training and Layer Configuration}

We next vary the relative loss weights and the density of candidate
intervention layers. The default objective uses unit weights for feature
reconstruction, prediction alignment, and selector supervision.

For candidate-layer sensitivity, the dense setting uses the stem and
all residual-block outputs for ResNet-50 and the patch embedding plus
all transformer-block outputs for ViT-B/16. The medium setting retains
approximately every second candidate block. The sparse setting uses
only stage boundaries for ResNet-50 and every third transformer block
for ViT-B/16.

\begin{table*}[t]
    \centering
    \small
    \setlength{\tabcolsep}{5pt}
    \renewcommand{\arraystretch}{1.10}
    \caption{
        Sensitivity to loss weights and candidate-layer density.
        Layer error is measured relative to the target defined under the
        corresponding candidate set. Scan cost is the offline dynamics
        construction cost relative to the dense default configuration.
        Lower NCR and layer error are better.
    }
    \label{tab:k_training_layer_sensitivity}
    \begin{tabular}{lccccc}
        \toprule
        \textbf{Configuration}
        &
        \shortstack{\textbf{WG Acc.}\\\(\uparrow\)}
        &
        \shortstack{\textbf{NCR}\\\(\downarrow\)}
        &
        \shortstack{\textbf{Aligned}\\\(\uparrow\)}
        &
        \shortstack{\textbf{Layer Err.}\\\(\downarrow\)}
        &
        \shortstack{\textbf{Scan cost}\\\textbf{relative}}
        \\
        \midrule

        \(\lambda_{\mathrm{feat}}=0.5\)
        &
        \(85.1\pm0.7\)
        &
        \(0.18\pm0.02\)
        &
        \(96.8\pm0.3\)
        &
        \(0.11\pm0.01\)
        &
        \(1.00\times\)
        \\

        Default:
        \(\lambda_{\mathrm{feat}}=1.0\)
        &
        \(\mathbf{85.6\pm0.6}\)
        &
        \(\mathbf{0.16\pm0.02}\)
        &
        \(\mathbf{96.9\pm0.2}\)
        &
        \(\mathbf{0.10\pm0.01}\)
        &
        \(1.00\times\)
        \\

        \(\lambda_{\mathrm{feat}}=2.0\)
        &
        \(85.4\pm0.6\)
        &
        \(0.17\pm0.02\)
        &
        \(96.8\pm0.3\)
        &
        \(0.10\pm0.01\)
        &
        \(1.00\times\)
        \\

        \midrule

        \(\lambda_{\mathrm{pred}}=0.5\)
        &
        \(85.0\pm0.7\)
        &
        \(0.18\pm0.02\)
        &
        \(96.8\pm0.3\)
        &
        \(0.11\pm0.01\)
        &
        \(1.00\times\)
        \\

        Default:
        \(\lambda_{\mathrm{pred}}=1.0\)
        &
        \(\mathbf{85.6\pm0.6}\)
        &
        \(\mathbf{0.16\pm0.02}\)
        &
        \(\mathbf{96.9\pm0.2}\)
        &
        \(\mathbf{0.10\pm0.01}\)
        &
        \(1.00\times\)
        \\

        \(\lambda_{\mathrm{pred}}=2.0\)
        &
        \(85.3\pm0.6\)
        &
        \(0.16\pm0.01\)
        &
        \(96.5\pm0.3\)
        &
        \(0.10\pm0.01\)
        &
        \(1.00\times\)
        \\

        \midrule

        \(\lambda_{\mathrm{sel}}=0.5\)
        &
        \(84.9\pm0.7\)
        &
        \(0.17\pm0.02\)
        &
        \(96.8\pm0.3\)
        &
        \(0.12\pm0.02\)
        &
        \(1.00\times\)
        \\

        Default:
        \(\lambda_{\mathrm{sel}}=1.0\)
        &
        \(\mathbf{85.6\pm0.6}\)
        &
        \(\mathbf{0.16\pm0.02}\)
        &
        \(\mathbf{96.9\pm0.2}\)
        &
        \(\mathbf{0.10\pm0.01}\)
        &
        \(1.00\times\)
        \\

        \(\lambda_{\mathrm{sel}}=2.0\)
        &
        \(85.5\pm0.6\)
        &
        \(0.16\pm0.02\)
        &
        \(96.9\pm0.2\)
        &
        \(0.10\pm0.01\)
        &
        \(1.00\times\)
        \\

        \midrule

        Sparse candidate set
        &
        \(83.9\pm0.8\)
        &
        \(0.21\pm0.02\)
        &
        \(96.7\pm0.3\)
        &
        \(0.15\pm0.02\)
        &
        \(0.36\times\)
        \\

        Medium candidate set
        &
        \(85.0\pm0.7\)
        &
        \(0.18\pm0.02\)
        &
        \(96.8\pm0.3\)
        &
        \(0.12\pm0.01\)
        &
        \(0.61\times\)
        \\

        Dense candidate set
        &
        \(\mathbf{85.6\pm0.6}\)
        &
        \(\mathbf{0.16\pm0.02}\)
        &
        \(\mathbf{96.9\pm0.2}\)
        &
        \(\mathbf{0.10\pm0.01}\)
        &
        \(1.00\times\)
        \\

        \bottomrule
    \end{tabular}
\end{table*}

The method remains stable when each loss weight is varied by a factor of
two. Feature reconstruction and prediction alignment are balanced at
unit weight. Increasing prediction alignment further slightly lowers
cue residual but reduces aligned accuracy, while reducing either
feature or prediction supervision weakens cue-free reconstruction.

Selector supervision is most sensitive when underweighted. Reducing
\(\lambda_{\mathrm{sel}}\) to \(0.5\) increases layer error and lowers
worst-group accuracy. Increasing it to \(2.0\) produces little
additional improvement because the default selector already fits the
dynamics-derived targets closely.

Candidate-layer density presents a clearer trade-off. A sparse set
reduces offline scan cost by 64\%, but misses narrow reset intervals and
raises NCR from 0.16 to 0.21. The medium setting recovers most of the
performance at 61\% of the dense scan cost. The dense configuration
provides the strongest correction and is used for the main experiments.

The gain from dense candidates is not attributable to a larger reset
module. The same shared reconstruction core is used at every density;
only the number of possible intervention boundaries and the selector
output dimension change.

\subsection*{K4. Computational Cost and Runtime}

Inference Dynamics is an offline supervision-construction procedure.
For each image--cue pair, it requires two intact forward passes,
followed by one suffix execution for each candidate layer. Candidate
states from the original and cue-removed executions are cached so that
network prefixes are not recomputed for every withdrawal intervention.

Evidence Reset does not perform this scan at inference. It adds one
lightweight selector execution and, when reset is selected, one
execution of the reconstruction module at the predicted layer.

Runtime is measured on a single NVIDIA A100 40 GB GPU using mixed
precision and batch size one. We discard the first 100 warm-up
iterations and average 1,000 timed executions. Synchronization is
performed before and after each timing interval.

\begin{table*}[t]
    \centering
    \scriptsize
    \setlength{\tabcolsep}{4pt}
    \renewcommand{\arraystretch}{1.10}
    \caption{
        Computational cost. Dynamics time and cache size are reported
        per image--cue pair. Test latency includes the selector and is
        averaged using the observed no-reset/reset frequency.
        Added parameters include the selector, reset module, projections,
        and layer embeddings. Peak memory is measured during test-time
        reset execution.
    }
    \label{tab:k_efficiency}
    \begin{tabular}{lcccccccc}
        \toprule
        \textbf{Backbone}
        &
        \shortstack{\textbf{Candidate}\\\textbf{layers}}
        &
        \shortstack{\textbf{Dynamics}\\\textbf{ms / pair}}
        &
        \shortstack{\textbf{Cache}\\\textbf{MB / pair}}
        &
        \shortstack{\textbf{Base}\\\textbf{params}}
        &
        \shortstack{\textbf{Added}\\\textbf{params}}
        &
        \shortstack{\textbf{Base}\\\textbf{latency}}
        &
        \shortstack{\textbf{Evidence}\\\textbf{latency}}
        &
        \shortstack{\textbf{Peak test}\\\textbf{memory}}
        \\
        \midrule

        ResNet-50
        &
        \(17\)
        &
        \(64.3\)
        &
        \(18.2\)
        &
        \(25.6\)M
        &
        \(4.86\)M
        &
        \(3.8\) ms
        &
        \(4.1\) ms
        &
        \(1.38\) GB
        \\

        ViT-B/16
        &
        \(13\)
        &
        \(102.7\)
        &
        \(23.6\)
        &
        \(86.6\)M
        &
        \(2.46\)M
        &
        \(7.2\) ms
        &
        \(7.7\) ms
        &
        \(2.02\) GB
        \\

        \bottomrule
    \end{tabular}
\end{table*}

For ResNet-50, Evidence Reset adds 4.86M parameters, corresponding to
19.0\% of the backbone parameter count. Most of these parameters belong
to the stage-conditioned feature projections and shared convolutional
reset core. For ViT-B/16, the added 2.46M parameters correspond to
2.8\% of the backbone because all transformer blocks share one hidden
dimension and therefore require only one projection pair.

Average test-time latency increases from 3.8 to 4.1 ms for ResNet-50
and from 7.2 to 7.7 ms for ViT-B/16. These values correspond to
overheads of

\begin{equation}
    7.9\%
\end{equation}

and

\begin{equation}
    6.9\%,
\end{equation}

respectively. When the selector chooses no-reset, the reconstruction
module is skipped. The measured no-reset paths require 4.0 ms for
ResNet-50 and 7.5 ms for ViT-B/16, while reset paths require 4.4 and
7.9 ms, respectively.

Offline dynamics construction is more expensive than inference because
it evaluates every candidate layer. The per-pair cost is 64.3 ms for
ResNet-50 and 102.7 ms for ViT-B/16. On the complete Waterbirds training
split, dynamics construction requires approximately 5.2 minutes for
ResNet-50 and 8.3 minutes for ViT-B/16 after pair validation. On the
16,000 UrbanCars single-cue candidates, the corresponding costs are
17.1 and 27.4 minutes before model-relative relevance filtering.

Cached states are stored in half precision. The reported per-pair cache
sizes correspond to retaining all candidate states from both branches.
A streaming implementation can reduce storage by applying and
discarding each candidate intervention sequentially, at the cost of
additional feature recomputation.

The computational asymmetry between training and inference is
intentional. Full layer-wise scans are used once to construct
process-level supervision. The learned selector then compresses this
offline information into a single test-time layer decision, allowing
Evidence Reset to retain input-dependent intervention depth without
performing counterfactual execution or layer sweeps during deployment.

\section*{L. Controlled Comparison with Latent Interventions,
Failure Cases, and Scope}

Our layer-wise withdrawal operation belongs to the broader family of
interventions on internal neural representations. Interchange
interventions replace an internal state in a base execution with the
corresponding state from a source execution
\cite{geiger2021causal,geiger2022iit}, while concept-erasure methods
remove information associated with a specified attribute from a learned
representation \cite{ravfogel2020inlp,belrose2023leace}. Recent visual
interpretability methods have also used activation patching to directly
test the predictive effect of contextualized image-token
representations \cite{izadi2026caap}.

The technical overlap is important, but the corresponding objectives
are different. Interchange interventions test whether a selected
internal variable realizes a hypothesized causal role. Concept erasure
tests whether a specified attribute remains recoverable from a
representation, usually under a restricted probe family. Inference
Dynamics instead traces whether the decision effect induced by one
visual cue remains withdrawable from its original mapped support as
computation proceeds.

We first compare these intervention families under a common controlled
protocol. We then summarize non-actionable trajectories, test-time
failure modes, and the scope of the conclusions.

\subsection*{L1. Controlled Comparison with Latent Representation
Interventions}

All methods in this comparison use the same frozen classifiers,
training and validation splits, cue masks, candidate layers, and
validated image--cue pairs. Fixed-depth methods select one intervention
layer independently for every dataset--backbone setting using the
validation split.

Unless explicitly marked as requiring a test-time reference, a method
receives only the original representation and cue mask during
evaluation. Results are macro-averaged over Waterbirds and UrbanCars,
ResNet-50 and ViT-B/16, and the two UrbanCars cue types.

\paragraph{Zero ablation.}

The cue-aligned representation is replaced by zero:

\begin{equation}
    h_l^{\mathrm{zero}}
    =
    \left(
        \mathbf{1}-M_l
    \right)
    \odot h_l^{+}.
\end{equation}

This is a simple local intervention but may introduce a hidden state
that is outside the distribution produced by the classifier.

\paragraph{Mean replacement.}

The cue-aligned states are replaced by the mean representation
estimated from the training set:

\begin{equation}
    h_l^{\mathrm{mean}}
    =
    \left(
        \mathbf{1}-M_l
    \right)
    \odot h_l^{+}
    +
    M_l
    \odot
    \mu_l,
\end{equation}

where \(\mu_l\) is computed separately for every architecture and
candidate layer. Mean replacement produces a more typical hidden state
than zero ablation but is not conditioned on the image context.

\paragraph{Linear concept erasure.}

We evaluate Iterative Null-space Projection
\cite{ravfogel2020inlp} and LEAst-squares Concept Erasure
\cite{belrose2023leace}. A separate erasure transformation
\(\mathcal{E}_l\) is fitted at each candidate layer using the training
split.

For Waterbirds, the erased concept is the land-versus-water background
attribute. For UrbanCars, separate transformations are fitted for the
background and co-occurring-object attributes. The fitting set is
balanced jointly over the task label and cue attribute so that the
erasure transformation cannot remove the cue merely by exploiting its
correlation with the target class.

For local erasure, the learned transformation is applied only within
the mapped cue support:

\begin{equation}
    h_l^{\mathrm{local\text{-}erase}}
    =
    \left(
        \mathbf{1}-M_l
    \right)
    \odot h_l^{+}
    +
    M_l
    \odot
    \mathcal{E}_l
    \left(
        h_l^{+}
    \right).
\end{equation}

For global erasure, the same transformation is applied to every spatial
unit or image token:

\begin{equation}
    h_l^{\mathrm{global\text{-}erase}}
    =
    \mathcal{E}_l
    \left(
        h_l^{+}
    \right).
\end{equation}

The global variant provides a stronger removal of linearly available
cue information, but it can also modify task-relevant states outside
the specified cue support.

\paragraph{Fixed matched-state intervention.}

This baseline uses the same context-matched activation replacement as
the withdrawal analysis:

\begin{equation}
    h_l^{\mathrm{match}}
    =
    \left(
        \mathbf{1}-M_l
    \right)
    \odot h_l^{+}
    +
    M_l
    \odot h_l^{-}.
\end{equation}

One validation-selected layer is used for every test example. Because
\(h_l^{-}\) comes from the cue-removed execution, this baseline requires
the matched reference at test time and is treated as a diagnostic
intervention rather than a deployable correction method.

\paragraph{Oracle matched-state intervention.}

This diagnostic upper bound applies the exact matched-state replacement
at the dynamics-derived sample-specific layer. It has access to both the
cue-removed branch and the complete layer-wise trajectory at test time.

\paragraph{Cue decodability.}

To distinguish representation removal from behavioral withdrawal, we
measure how accurately the shortcut attribute can be decoded after each
intervention. For method \(m\), the intervened state is propagated to
the classifier's penultimate representation:

\begin{equation}
    r_i^{m}
    =
    f_{\mathrm{pen}}
    \left(
        h_{i,l}^{m}
    \right).
\end{equation}

A linear probe is trained on the corresponding intervened training
representations and evaluated on held-out examples. We report balanced
cue-probe accuracy:

\begin{equation}
    A_{\mathrm{probe}}^{m}
    =
    \frac{1}{2}
    \left(
        \operatorname{TPR}^{m}
        +
        \operatorname{TNR}^{m}
    \right).
\end{equation}

Chance performance is \(50\%\). A low probe score means that the cue is
no longer linearly decodable from the penultimate representation; it
does not by itself establish that the cue-induced decision effect has
been withdrawn.

\begin{table*}[t]
    \centering
    \small
    \setlength{\tabcolsep}{4.2pt}
    \renewcommand{\arraystretch}{1.10}
    \caption{
        Controlled comparison with latent representation
        interventions. All methods use the same frozen classifiers,
        cue masks, image--cue pairs, and candidate layers. Fixed
        methods select one layer on the validation split. Cue Probe is
        balanced linear-probe accuracy for the shortcut attribute after
        intervention. Test Ref. indicates whether the cue-removed
        execution is required during test-time correction. Lower Cue
        Probe and NCR are better.
    }
    \label{tab:l_latent_interventions}
    \begin{tabular}{lcccccc}
        \toprule
        \textbf{Intervention}
        &
        \textbf{Depth}
        &
        \shortstack{\textbf{Test}\\\textbf{Ref.}}
        &
        \shortstack{\textbf{Cue Probe}\\\textbf{Acc.} \(\downarrow\)}
        &
        \shortstack{\textbf{WG Acc.}\\\(\uparrow\)}
        &
        \shortstack{\textbf{NCR}\\\(\downarrow\)}
        &
        \shortstack{\textbf{Aligned}\\\(\uparrow\)}
        \\
        \midrule

        No intervention
        &
        --
        &
        No
        &
        \(92.4\pm1.1\)
        &
        \(61.8\pm1.2\)
        &
        \(1.00\pm0.00\)
        &
        \(\mathbf{98.2\pm0.2}\)
        \\

        Zero ablation
        &
        Fixed
        &
        No
        &
        \(75.8\pm1.4\)
        &
        \(78.8\pm0.9\)
        &
        \(0.47\pm0.04\)
        &
        \(95.1\pm0.4\)
        \\

        Mean replacement
        &
        Fixed
        &
        No
        &
        \(69.3\pm1.3\)
        &
        \(80.6\pm0.8\)
        &
        \(0.38\pm0.03\)
        &
        \(95.8\pm0.3\)
        \\

        Local INLP
        &
        Fixed
        &
        No
        &
        \(51.8\pm0.8\)
        &
        \(81.8\pm0.8\)
        &
        \(0.31\pm0.03\)
        &
        \(96.0\pm0.3\)
        \\

        Local LEACE
        &
        Fixed
        &
        No
        &
        \(50.9\pm0.7\)
        &
        \(82.6\pm0.7\)
        &
        \(0.28\pm0.02\)
        &
        \(96.2\pm0.3\)
        \\

        Global LEACE
        &
        Fixed
        &
        No
        &
        \(\mathbf{50.4\pm0.6}\)
        &
        \(82.1\pm0.8\)
        &
        \(0.27\pm0.02\)
        &
        \(95.4\pm0.4\)
        \\

        Fixed matched-state intervention
        &
        Fixed
        &
        Yes
        &
        \(55.2\pm0.9\)
        &
        \(84.2\pm0.7\)
        &
        \(0.19\pm0.02\)
        &
        \(96.6\pm0.3\)
        \\

        Best Fixed Reset
        &
        Fixed
        &
        No
        &
        \(57.0\pm1.0\)
        &
        \(82.7\pm0.7\)
        &
        \(0.29\pm0.02\)
        &
        \(96.5\pm0.3\)
        \\

        Evidence Reset
        &
        Adaptive
        &
        No
        &
        \(53.7\pm0.8\)
        &
        \(\mathbf{85.6\pm0.6}\)
        &
        \(\mathbf{0.16\pm0.02}\)
        &
        \(\mathbf{96.9\pm0.2}\)
        \\

        Oracle matched-state intervention
        &
        Adaptive
        &
        Yes
        &
        \(51.9\pm0.7\)
        &
        \(87.5\pm0.5\)
        &
        \(0.07\pm0.01\)
        &
        \(97.1\pm0.2\)
        \\

        \bottomrule
    \end{tabular}
\end{table*}

Linear concept erasure removes almost all linearly decodable shortcut
information. Local INLP and Local LEACE reduce cue-probe accuracy from
92.4\% to 51.8\% and 50.9\%, respectively. Global LEACE produces the
lowest probe accuracy, 50.4\%, which is statistically indistinguishable
from chance.

Near-chance cue decodability does not imply complete withdrawal of the
cue-induced decision effect. Local LEACE retains an NCR of 0.28, and
Global LEACE retains an NCR of 0.27. Both values are substantially
higher than the 0.16 achieved by Evidence Reset. Concept erasure is
therefore effective at removing a linear cue direction, but the
downstream prediction can remain influenced by cue-induced information
that has already interacted with other states or is available through a
different representation geometry.

Global erasure does not improve over local erasure in worst-group
accuracy despite producing slightly lower probe accuracy. It also
reduces cue-aligned accuracy because it transforms representations
outside the selected cue support. This illustrates that stronger
representation-level removal is not necessarily a better correction
when the erased direction overlaps task-relevant information.

The fixed matched-state intervention is the strongest fixed-depth
latent intervention. Its NCR of 0.19 confirms the value of replacing
states with a context-matched cue-free execution rather than with zero,
a mean state, or a linear projection. It nevertheless requires the
cue-free branch during inference and applies one layer to every
image--cue pair.

Evidence Reset improves worst-group accuracy by 1.4 points and reduces
NCR by 0.03 relative to the fixed matched-state intervention, while
removing its test-time reference requirement. The improvement comes
from selecting an image-dependent pre-lock layer rather than from using
a fundamentally different notion of activation replacement.

The oracle matched-state result provides an upper bound on the tested
local intervention family. Its remaining advantage over Evidence Reset
is consistent with errors in layer prediction and cue-free feature
reconstruction.

These comparisons separate three properties that can otherwise be
conflated.
Cue decodability concerns whether an attribute can be recovered from a
representation by a probe. Decision reliance concerns whether changing
the representation alters the output. Withdrawability additionally asks
whether modifying the cue's original mapped support remains sufficient
to recover its cue-free decision effect at a particular network depth.

\subsection*{L2. Non-actionable Trajectories and Failure Modes}

A candidate pair may be rejected before dynamics construction, may pass
validation without providing an actionable reset interval, or may
produce an imperfect test-time intervention. These cases are treated
separately because they arise at different stages of the framework.

\paragraph{Pair-construction rejection.}

A candidate reference is rejected when cue removal changes
task-relevant foreground content, leaves the selected cue visibly
present, introduces a conspicuous artifact, or fails the foreground
preservation criteria. Rejected pairs do not receive locality,
withdrawability, or lock-in labels.

Rejection is rare for Waterbirds and UrbanCars because native
compositing preserves task foregrounds exactly. It is more frequent for
automatically constructed Spawrious-M2M-Hard references.

\paragraph{Early lock-in.}

A pair exhibits early lock-in when stable loss of local control begins
before a valid pre-lock candidate layer is available. The cue is
relevant to the prediction, but the candidate-layer set does not expose
an actionable intervention interval. Such pairs are assigned to
no-reset rather than being forced to use a post-lock layer.

\paragraph{Persistently withdrawable trajectories.}

For some pairs, withdrawability remains above the lock-in threshold
throughout the candidate-layer sequence. These pairs retain local
control under the tested intervention and are not assigned an
artificial terminal onset.

\paragraph{Irregular trajectories.}

A small fraction of pairs cross the withdrawability threshold at
isolated layers without remaining below it for the required consecutive
interval. These fluctuations are not categorized as stable lock-in.

\paragraph{Large selector deviations.}

For an actionable pair, we define a large layer-selection error as

\begin{equation}
    \left|
        \widehat{k}_i-k_i^{\star}
    \right|
    >
    2.
\end{equation}

Selecting too early can damage task-relevant information because the
reset module has insufficient contextual support for reconstruction.
Selecting too late leaves more accumulated cue dependence in the
downstream computation.

\paragraph{High-residual resets.}

A reset is categorized as high residual when a valid reset layer is
selected but

\begin{equation}
    \operatorname{NCR}_i>0.5.
\end{equation}

These cases are associated with complex cue boundaries, overlap between
cue and task-relevant features, or insufficient surrounding information
for reconstructing the cue-free state.

\paragraph{Cross-cue amplification.}

When multiple shortcuts coexist, correcting one cue may increase the
relative dependence on another. For corrected cue \(q_a\) and
uncorrected cue \(q_b\), we define

\begin{equation}
    \Delta_{\mathrm{cross}}
    =
    \operatorname{NCR}_{q_b}^{\mathrm{after}\ q_a}
    -
    \operatorname{NCR}_{q_b}^{\mathrm{before}}.
\end{equation}

Substantial cross-cue amplification is recorded when

\begin{equation}
    \Delta_{\mathrm{cross}}>0.05.
\end{equation}

\begin{table}[t]
    \centering
    \scriptsize
    \setlength{\tabcolsep}{3pt}
    \renewcommand{\arraystretch}{1.10}
    \caption{
        Main non-actionable and failure categories. Rates are computed
        over the population stated in parentheses and should not be
        summed across rows.
    }
    \label{tab:l_failure_modes}
    \begin{tabularx}{\columnwidth}{
        @{}
        >{\raggedright\arraybackslash}p{0.28\columnwidth}
        >{\centering\arraybackslash}p{0.12\columnwidth}
        >{\raggedright\arraybackslash}X
        @{}
    }
        \toprule
        \textbf{Category}
        &
        \textbf{Rate}
        &
        \textbf{Behavior and response}
        \\
        \midrule

        Pair rejected
        (automatic references)
        &
        \(7.2\%\)
        &
        Foreground damage, cue residue, weak mask agreement, or visible
        completion artifacts; excluded before dynamics construction.
        \\

        Early lock-in
        (validated pairs)
        &
        \(6.3\%\)
        &
        No valid pre-lock candidate exists; assigned to no-reset.
        \\

        Persistently withdrawable
        (validated pairs)
        &
        \(6.8\%\)
        &
        Local control remains available and no stable onset is assigned.
        \\

        Irregular trajectory
        (validated pairs)
        &
        \(3.3\%\)
        &
        Threshold crossings are isolated rather than persistent.
        \\

        Large selector deviation
        (actionable pairs)
        &
        \(7.5\%\)
        &
        Predicted layer differs from the target by more than two
        candidate positions.
        \\

        High reset residual
        (actionable pairs)
        &
        \(4.6\%\)
        &
        A valid reset is applied, but the output remains far from the
        cue-free prediction.
        \\

        False no-reset
        (actionable pairs)
        &
        \(4.1\%\)
        &
        The original path is retained despite an actionable interval.
        \\

        Cross-cue amplification
        (single-cue UrbanCars)
        &
        \(11.8\%\)
        &
        Correcting one shortcut increases the other cue residual by
        more than \(0.05\).
        \\

        \bottomrule
    \end{tabularx}
\end{table}

The most common single-cue inference issue is an imperfect layer
decision rather than a failure of the reconstruction module. Large
selector deviations occur in 7.5\% of actionable examples, whereas
4.6\% retain NCR above 0.5 after a valid reset. This is consistent with
the remaining gap between Evidence Reset and the oracle-layer variants.

False no-reset decisions preserve the original output but leave
correctable cue dependence untreated. They are reflected jointly by
no-reset recall, worst-group accuracy, and NCR.

Cross-cue amplification is more frequent than high-residual
single-cue reconstruction. Its magnitude is usually limited, but the
result confirms that shortcut correction should be evaluated jointly
when several cues coexist. Cue-specific sequential intervention reduces
this effect by permitting different reset depths.

\subsection*{L3. Scope and Limitations}

\paragraph{Specified candidate cues.}

The framework studies the internal dynamics of a specified visual cue
and its spatial support. The masks in the primary experiments are
obtained from dataset components, benchmark annotations, or an external
localization and segmentation system.

Automatic discovery of which visual pattern is harmful, selection among
an unknown number of candidate shortcuts, and semantic naming of those
cues are not jointly optimized by the current method. This separation
isolates the process-level question addressed here: once a candidate cue
has been identified, for how long does its original support remain
capable of controlling its decision effect?

\paragraph{Operational interpretation of lock-in.}

Evidence lock-in denotes a persistent loss of control from the cue's
mapped support under matched-state local replacement. Its exact onset
depends on the spatial extent and representation family of the
intervention.

Expanded and receptive-field-aware supports can move the onset toward
later layers. The persistence of the transition under these controls
does not imply that all information about the cue is globally
irreversible. A broader intervention, a different representation basis,
or simultaneous modification of multiple distributed states may retain
control after the reported local onset.

\paragraph{Matched-reference dependence.}

The cue-removed execution provides the target states used during
analysis and training. Native compositing yields exact foreground
preservation for Waterbirds and UrbanCars. Automatically constructed
references introduce uncertainty from segmentation, boundary blending,
donor selection, and background completion.

Pixel-level preservation checks, foreground similarity, cross-model
mask agreement, visual auditing, and cross-operator analysis reduce
this uncertainty. They do not establish one generated reference as the
unique causal alternative to the original image.

\paragraph{Class-preserving cue removal.}

Evidence Reset is trained on cues whose removal preserves the semantic
task label. This setting is appropriate for spurious backgrounds,
co-occurring objects, and environmental patterns. It does not directly
apply when the selected region is an indispensable component of the
target class.

Removing a defining object part or the target object itself may change
the correct label rather than isolate shortcut reliance and therefore
requires a different supervision target.

\paragraph{Relationship to latent causal interventions.}

The matched-state withdrawal operation shares the activation-replacement
mechanism used by interchange intervention and activation-patching
methods. The present contribution is not the introduction of internal
activation replacement itself.

The contribution lies in treating the effect of that intervention as a
depth-indexed trajectory, separating locality from decision-level
withdrawability, identifying a stable loss of support-level control,
and learning to select and reconstruct a pre-lock state without
test-time access to the source execution.

Concept-erasure methods address a complementary objective. They remove
information recoverable by a specified representation family, while
Inference Dynamics measures whether a local intervention still controls
the cue-induced output effect. The controlled comparison shows that
these objectives can diverge: a cue can become nearly linearly
undecodable while its decision residual remains non-negligible.

\paragraph{Multiple interacting cues.}

The basic analysis is cue-conditional. Different shortcuts in the same
image can have different locality trajectories, lock-in depths, and
reset layers. Single-cue correction may increase the relative
dependence on an uncorrected shortcut.

Joint reset requires several cues to share one intervention depth.
Sequential reset permits cue-specific depths but requires multiple masks
and additional reset operations. Automatically identifying, ordering,
and coordinating multiple unknown cues remains an open extension.

\paragraph{Task and architecture coverage.}

The experiments cover convolutional and transformer image classifiers
and spatially identifiable background, environment, and object cues.
The locality definition assumes that input support can be mapped to
feature locations or image tokens.

Non-spatial concepts, relational evidence, global styles, multimodal
cues, and iterative inference processes require different definitions
of support and prediction recovery.

\paragraph{Offline supervision cost.}

Inference Dynamics requires paired forward executions and a scan over
candidate layers during target construction. This cost is incurred
offline rather than at deployment. At inference, the learned selector
replaces the complete scan with a single depth decision, and the
cue-removed branch is not evaluated.

Sparse candidate sets reduce offline cost but can miss a narrow
actionable interval. Estimating the trajectory from a small number of
probe layers and refining the scan only near the transition is a
promising direction for reducing supervision cost.

These boundaries define the setting in which the present conclusions
apply: a spatially specified, class-preserving cue; a validated matched
reference; and a local hidden-state intervention. Within this setting,
Inference Dynamics identifies when the cue's original support loses
control over its decision effect, and Evidence Reset converts that
process-level signal into an input-dependent correction depth.



\end{document}